\documentclass[11pt]{article}

\usepackage[final]{acl}

\usepackage[T1]{fontenc}
\usepackage[utf8]{inputenc}

\usepackage{times}
\usepackage{latexsym}
\usepackage{microtype}
\usepackage{inconsolata}
\usepackage{amsmath,amssymb}
\usepackage{xurl}

\usepackage{graphicx}
\usepackage{subcaption}

\usepackage{colortbl}
\definecolor{em}{gray}{0.9}
\newcommand{\cem}{\cellcolor{em}}
\usepackage{enumitem}
\usepackage{booktabs}
\usepackage{multirow}
\usepackage[ruled,vlined,linesnumbered]{algorithm2e}

\usepackage{listings}
\usepackage[most]{tcolorbox}
\tcbset{
  colframe=black!65,
  colback=black!1,
  boxrule=0.2mm,
  width=\textwidth,
  arc=1mm,
  fonttitle=\bfseries\normalsize,
  coltitle=white,
  colbacktitle=cyan!45!black
}

\definecolor{mydarkblue}{rgb}{0,0.08,0.45}
\PassOptionsToPackage{colorlinks,citecolor=mydarkblue,urlcolor=mydarkblue,linkcolor=mydarkblue}{hyperref}

\title{AdaptiveK: Complexity-Driven Sparse Autoencoders for Interpretable Language Model Representations}

\author{Yifei Yao\textsuperscript{1}, Hanrong Zhang\textsuperscript{2}, \textbf{Mengnan Du\textsuperscript{3*}}\\
\textsuperscript{1}Zhejiang University \,
\textsuperscript{2}University of Illinois Chicago \,\\
\textsuperscript{3}The Chinese University of Hong Kong, Shenzhen\\
\small\texttt{yifei3.23@intl.zju.edu.cn}, 
\small\texttt{hzhan135@uic.edu}, 
\small\texttt{mengnandu@cuhk.edu.cn}\\[2pt]
\small\textsuperscript{*}Corresponding author
}

\begin{document}
\maketitle
\let\thefootnote\relax\footnotetext{Accepted by ACL 2026.}
\begin{abstract}
Understanding the internal representations of large language models (LLMs) remains a central challenge for interpretability research. Sparse autoencoders (SAEs) offer a promising solution by decomposing activations into interpretable features, but existing approaches rely on fixed sparsity constraints that fail to account for input complexity. We propose \textbf{AdaptiveK SAE} (Adaptive Top K Sparse Autoencoders), a novel framework that dynamically adjusts sparsity levels based on the semantic complexity of each input. Leveraging linear probes, we demonstrate that context complexity is linearly encoded in LLM representations, and we use this signal to guide feature allocation during training. Experiments across ten language models demonstrate that this complexity-driven adaptation  outperforms fixed-sparsity approaches on reconstruction fidelity, explained variance, cosine similarity and interpretability metrics while eliminating the  burden of extensive hyperparameter tuning. Our code is available at: \url{https://github.com/hiyukie/adaptiveK}.
\end{abstract}

\section{Introduction}

As large language models (LLMs) continue to advance, understanding their internal representations becomes increasingly crucial yet challenging. These models operate as ``black boxes'' with activation spaces that resist straightforward analysis \citep{olah2020zoom,  
ferrando2024primer}. Individual components typically respond to multiple unrelated concepts (polysemanticity) \citep{olah2023distributed}, while the models encode more distinct features than their dimensional capacity would suggest (superposition) \citep{arora2018linear, gurnee2023finding, elhage2022toy}. This efficient but complex encoding creates a significant interpretability barrier as traditional approaches cannot untangle the overlapping information patterns. Sparse autoencoders (SAEs) \citep{bricken2023towards, cunningham2023sparse} address this challenge by decomposing model activations into sparse combinations of interpretable features, revealing the underlying structure of the model's representations.

Recent research has rapidly expanded the capabilities of sparse autoencoders, scaling them to extract millions of interpretable features from frontier models \citep{templeton2024scaling} while introducing numerous architectural innovations \citep{rajamanoharan2024improving, gao2024scaling, bussmann2024batchtopk, rajamanoharan2024jumping, ProLUNonlinearity, mudide2024efficient, bussmannlearning, karvonen2024measuring, marks2024enhancing, braun2024identifying}. However, despite these advancements, current SAE architectures rely on uniform sparsity constraints regardless of input complexity. Whether through activation-limiting approaches like TopK \citep{gao2024scaling} and BatchTopK \citep{bussmann2024batchtopk} that enforce a fixed number of active features (k), or penalty-based methods like Gated SAEs  \citep{rajamanoharan2024improving} and P-anneal \citep{karvonen2024measuring} that apply consistent regularization pressure, these designs create fundamental inefficiencies where conceptually simple inputs receive excessive representational capacity while complex inputs face insufficient feature allocation. This limitation becomes increasingly problematic at scale as \citet{gao2024scaling} demonstrate that larger language models require proportionally more features to achieve comparable reconstruction quality. Moreover, finding optimal sparsity settings requires extensive hyperparameter experimentation to navigate the critical reconstruction-sparsity trade-off \citep{karvonen2025saebench}.

To overcome this limitation, we propose that \emph{a sparse autoencoder should adaptively adjust sparsity levels based on input complexity}. When a simple semantic concept can be effectively explained and reconstructed using only a few features, activating additional features becomes unnecessary. This approach not only conserves computational resources but also prevents unnecessary feature activation on simpler texts, reducing both overfitting and representational noise.

How can we define semantic simplicity or complexity within LLM representation spaces? \citet{peters2018dissecting} demonstrated that probes can map LLM intermediate representations to semantic and syntactic information. Similarly, higher-level concepts such as political perspective \citep{kim2025linear}, sentiment \citep{tigges2023linear}, and spatiotemporal information \citep{gurnee2023language} have been shown to be linearly represented in activation spaces. 
% Furthermore, when language models process texts of varying complexity, predicting the next token in simpler texts requires less information than in more complex ones. Driven by autoregressive training objectives, models must learn to differentiate and represent various levels of textual complexity to minimize prediction loss.
Based on these observations, we hypothesize that the internal representations formed by large language models during text processing naturally encode multidimensional properties of text, including its complexity.

Our study first establishes that text complexity is linearly encoded in language model representations. We score contexts using GPT-4.1-mini API~\citep{GPT41Mini} across six semantic dimensions to create aggregate complexity scores, then train linear regression probes on model activations to predict these scores. Experiments with eight different scale LLMs demonstrate high correlation coefficient, confirming our hypothesis that LLMs naturally encode text complexity in their representation spaces. Analysis reveals that texts of varying complexity require proportional representational capacity, with complex inputs necessitating more active features for accurate encoding. This key insight suggests that adaptive sparsity mechanisms could significantly improve autoencoder efficiency.

Based on our complexity prediction capabilities, we develop the \textbf{Adaptive Top K Sparse Autoencoder (AdaptiveK SAE)}, which to our knowledge is the first work to solve computational efficiency bottlenecks in sparse autoencoder training while maintaining feature quality. Our approach quantifies context complexity using a linear probe trained on multi-dimensional complexity annotations. This score determines an appropriate sparsity level, activating more features for complex inputs while maintaining high sparsity for simpler ones. This complexity-driven adaptation better balances reconstruction quality, sparsity, and interpretability. Experiments demonstrate our framework outperforms fixed-sparsity approaches across multiple model scales. Our main contributions include:
\vspace{-5pt}
\begin{itemize}
[leftmargin=*]\setlength\itemsep{-0.1em}
\item We propose AdaptiveK Sparse Autoencoders, a novel framework that dynamically adjusts sparsity levels based on input complexity.
\item We demonstrate that text complexity is linearly encoded in model representations, establishing a direct relationship between semantic complexity and representational capacity needs in LLMs.
\item Experiments across ten LLMs show that our SAE consistently outperforms fixed-sparsity baselines on reconstruction fidelity, explained variance, cosine similarity and other metrics.
\end{itemize}

\begin{figure}[t]
  % \centering
    \includegraphics[width=1\linewidth]{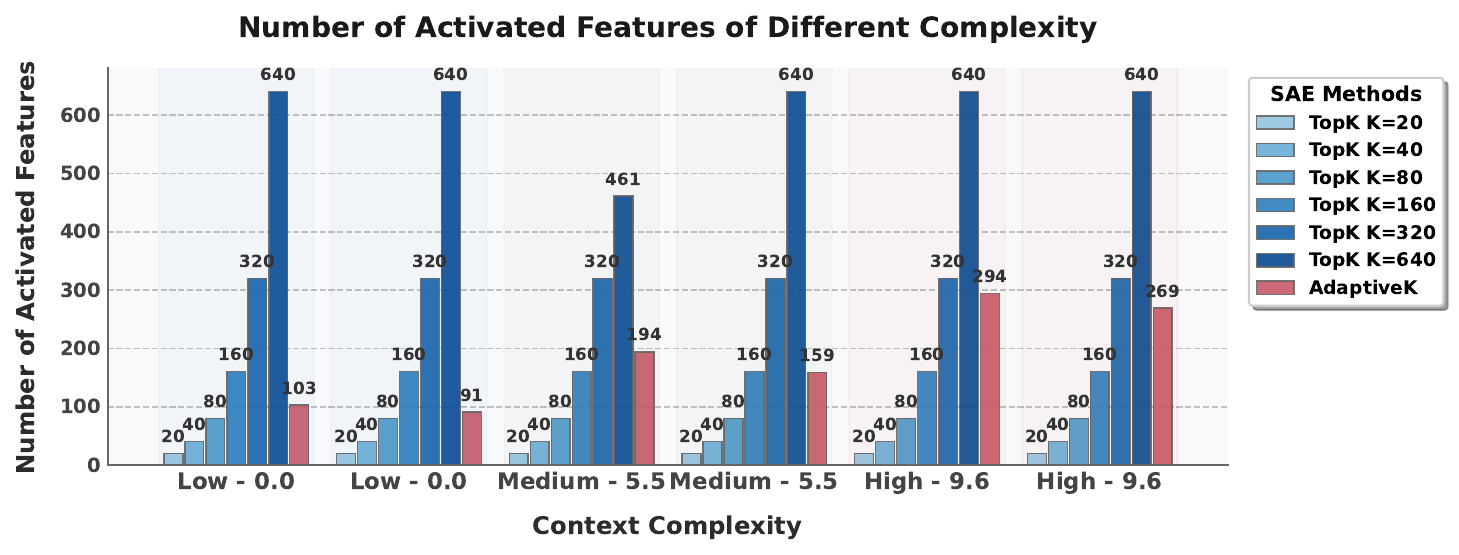}
  \captionsetup{skip=8pt} 
  \caption{Two samples were selected from each complexity level (simplest=0, moderate=5.5, and most complex=9.6) from test set. In TopK SAE, feature activation strictly follows the k value between 20-320, but often falls below the threshold when k=640. For Pythia-160M, fixed TopK (blue) maintain constant activation, while AdaptiveK (red) dynamically scales with complexity.}
  \vspace{-8pt}
  \label{fig:activation_bar}
\end{figure}

\begin{figure*}[t]
  \includegraphics[width=0.8\linewidth]{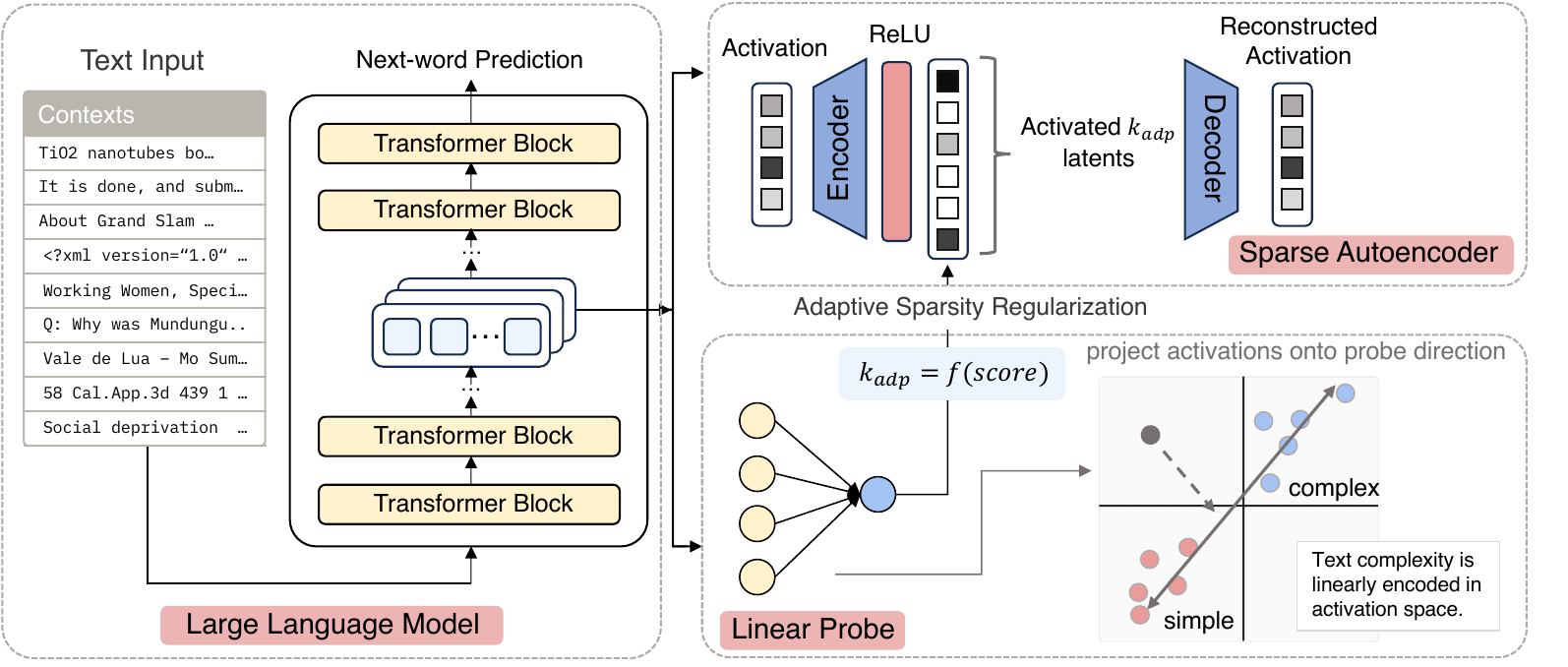}
  \centering
  \caption{Overall pipeline of the AdaptiveK SAE. Input text is fed into a LLM to extract internal activations, which are then passed through both a linear probe that predicts text complexity and a SAE for decomposition. During training, the linear probe's complexity score dynamically determines the number of features to activate, allowing more features for complex inputs and fewer for simple ones.}
  \vspace{-8pt}
  \label{fig:overall_pipeline}
\end{figure*}

\section{Related Work}

\subsection{Sparse Autoencoders and Improvements}
A significant challenge in neural network interpretability is polysemanticity, where units (\emph{e.g.}, neurons) respond to diverse, semantically distinct inputs, complicating functional analysis \citep{elhage2022toy, olah2020zoom}. The superposition hypothesis \citep{park2023linear} suggests that networks represent more features than available neurons by encoding them as directions in activation space, rather than solely via individual neuron activities. SAEs offer an unsupervised dictionary learning approach to tackle this, decomposing internal network activations (particularly in LLMs) to reveal latent interpretable units \citep{ferrando2024primer, shu2025surveysparseautoencodersinterpreting}. Core to SAEs is an encoder mapping an activation $x$ to a higher-dimensional sparse representation $z$, and a decoder reconstructing $\hat{x}$ from $z$. Goal for SAEs are to isolate monosemantic and composable features, thus offering a more faithful representation of the model's internal computational state \citep{huben2023sparse}.

Following initial SAE proposals \citep{cunningham2023sparse}, research rapidly advanced SAE design \citep{lee2025evaluatingdesigningsparseautoencoders}. Efforts addressed limitations like L1 penalty-induced shrinkage, leading to Gated SAEs \citep{rajamanoharan2024improving}. Alternative sparsity mechanisms emerged, including TopK \citep{gao2024scaling}, BatchTopK \citep{bussmann2024batchtopk}, JumpReLU \citep{rajamanoharan2024jumping}, and ProLU \citep{ProLUNonlinearity}. Architectural innovations like Switch SAEs improved computational scaling \citep{mudide2024efficient}, while Matryoshka SAEs targeted feature hierarchy and splitting/absorption issues \citep{bussmannlearning}. Optimization objectives were also refined through techniques like P-annealing \citep{karvonen2024measuring}, feature alignment \citep{marks2024enhancing}, and end-to-end training \citep{braun2024identifying}. While these advancements often optimize proxy metrics like sparsity and fidelity, their alignment with true interpretability remains an active area of evaluation.

\subsection{Linear Probe in Large Language Models}
\label{subsec: linear probe}
Linear probes have emerged as a fundamental method for elucidating how LLMs represent complex information within their activation spaces \citep{li2023inference, von2024language, mikolov2013linguistic}. This approach is grounded in the hypothesis that important high-level concepts are encoded linearly as directions in representational space \citep{kim2025linear, liu2311context}. A substantial body of research supports the finding that linear probes are more effective than nonlinear probes at aligning model representations with specific behaviors. For instance, \citet{gurnee2023language} applied linear probes to Llama-2 models to reveal that LLMs learn linear representations of space and time that are robust to prompting variations, unified across entity types, and encoded by specific neurons within the network. Similarly, \citet{tigges2023linear} demonstrated that sentiment in language models emerges along specific linear directions in activation space, with a single dimension causally controlling sentiment polarity (positive versus negative) in model outputs through direct interventions. Additionally, concepts such as topic direction \citep{turner2023steering}, political ideology \citep{kim2025linear}, game states \citep{nanda2023emergent}, truthfulness \citep{marks2023geometry} and safety \citep{arditi2024refusal} have also been identified as important features that are linearly encoded within the internal activation spaces of LLMs.

\section{Preliminaries and Motivation}

\subsection{Baseline Sparse Autoencoders}
Following the initial works that introduced SAEs for decomposing model representations \citep{cunningham2023sparse}, a variety of architectural refinements have emerged. Our comparative analysis of these baselines was facilitated by the dictionary\_learning library \citep{marks2024dictionarylearning}. Foundational ReLU SAEs \citep{bricken2023towards} and the refined one~\citep{anthropic2024circuits} typically map an input activation $x\in\mathbb{R}^d$ to a sparse latent $z\in\mathbb{R}^M$ (where $M\gg d$) and then to a reconstruction $\hat{x}$. The core operations involve an encoder:
\begingroup
\setlength{\abovedisplayskip}{3pt}
\setlength{\belowdisplayskip}{3pt}
\begin{equation}
    z=\mathrm{ReLU}(W_{enc}(x-b_{pre})+b_{enc}),
\end{equation}
\endgroup and a decoder:
\begingroup
\setlength{\abovedisplayskip}{3pt}
\setlength{\belowdisplayskip}{3pt}
\begin{equation}
\label{equ:decoder}
    \hat{x}=W_{dec}z+b_{pre},
\end{equation}
\endgroup
with a training loss that combines reconstruction error with an $L_{1}$ sparsity penalty on $z$.
\begingroup
\setlength{\abovedisplayskip}{3pt}
\setlength{\belowdisplayskip}{3pt}
\begin{equation}
    L=\|x-\hat{x}\|_2^2+\lambda\|z\|_1.
\end{equation}
\endgroup
To mitigate issues like feature shrinkage from the $L_{1}$ penalty, Gated SAEs \citep{rajamanoharan2024improving} decouple the $L_{1}$-penalized feature selection gate $g(x)$ from magnitude estimation $m(x)$, forming activations as $z=g(x) \odot m(x)$. Other approaches enforce sparsity directly: TopK SAEs \citep{gao2024scaling} select the K highest pre-activations.
\begingroup
\setlength{\abovedisplayskip}{3pt}
\setlength{\belowdisplayskip}{3pt}
\begin{equation}
    z=\mathrm{ReLU}(\mathrm{ReLU}(W_{enc}(x-b_{pre})+b_{enc}),K).
\end{equation}
\endgroup
BatchTopK SAEs \citep{bussmann2024batchtopk} extend this by selecting the top $N \times K$ activations across a batch of $N$ samples. JumpReLU SAEs \citep{rajamanoharan2024jumping} utilize a discontinuous activation $z_i=\operatorname{act}_i \cdot H\left(\operatorname{act}_i-\theta_i\right)$ with a learned threshold $\theta_i$ (where $\operatorname{act}_i$ is preactivation and $H$ is Heaviside), often paired with a $L_{0}$ sparsity term and trained via Straight-Through Estimators. For refining loss-based sparsity, P-anneal ReLU SAEs \citep{karvonen2024measuring} employ an $L_{p}$ norm penalty, $\lambda\sum_i|z_i|^p$, where $p$ anneals from 1 towards 0. Lastly, to address feature hierarchy and issues like splitting or absorption, Matryoshka BatchTopK SAEs \citep{bussmannlearning} train nested dictionaries of increasing capacity, using BatchTopK for sparsity within each level.

\subsection{Our Motivation}

Despite advances in SAE scalability, current approaches face a limitation: they apply uniform sparsity constraints regardless of input complexity. This ``one-size-fits-all'' approach creates inefficiencies across the representation space. Whether employing fixed activation methods such as TopK \citep{gao2024scaling} or regularization techniques like Gated SAEs \citep{rajamanoharan2024improving} existing architectures cannot adapt to varying complexity of different inputs. As shown in Fig. \ref{fig:activation_bar} conventional SAE methods with fixed K maintain constant activation (\emph{e.g.}, 80 features across all complexities), while our AdaptiveK dynamically scales from 103 to 394 features based on input complexity. This inefficiency becomes more pronounced at scale, determining optimal sparsity parameters necessitates extensive hyperparameter optimization to balance reconstruction fidelity against sparsity constraints.

Our approach is motivated by the observation that text complexity is linearly encoded in language model representations, suggesting a more efficient solution: adaptively adjusting sparsity levels based on input complexity. Simple inputs require fewer features for reconstruction, while complex ones need more representational capacity. This adaptive allocation improves computational efficiency, reduces overfitting on simple inputs, and enhances interpretability, all achieved within a single training run without extensive hyperparameter tuning.

\section{The Proposed AdaptiveK SAE}
\label{SAE}
In this work, we propose AdaptiveK Sparse Autoencoder that dynamically adjusts sparsity to input complexity. By allocating representational capacity in proportion to content complexity, AdaptiveK addresses the core inefficiency of uniform sparsity (Fig.~\ref{fig:overall_pipeline}). The architecture comprises two components: (1) a linear probe that predicts input complexity (Sec.~\ref{Linear Probe}); and (2) an SAE with AdaptiveK activation (Sec.~\ref{sec:adaptive-k-architecture}). We also present a three-phase training procedure that stabilizes and improves learning (Sec.~\ref{sec:adaptive-k-training}).

\subsection{Linear Probe for Complexity Prediction}\label{Linear Probe}

\noindent
\textbf{Linear Probe Training.}
\label{Training Settings}
Our dataset comprises contexts from pile-uncopyrighted \citep{gao2020pile}, each containing 1024 tokens. Complexity is quantified on a scale from 0 to 10, based on a six-dimensional evaluation (lexical complexity, syntactic complexity, conceptual density, domain specificity, logical structure and logical structure) by GPT-4.1-mini (scoring prompts detailed in Appendix~\ref{sec:linear-probes-details}), yielding target labels $y_i$. These complexity scores are floating-point numbers with one decimal place. For each context, we extract the hidden state activations of the last token from selected layers of the auto-regressive transformer language models Pythia (70M, 160M) \citep{biderman2023pythia}, Gemma-2 (2B, 9B) \citep{team2024gemma}, Llama-3.1 (8B) \citep{grattafiori2024llama3herdmodels}, Qwen-3 (8B, 14B) \citep{yang2025qwen3technicalreport} and Phi-4 (14B) \citep{abdin2024phi4technicalreport}, which encapsulate contextual information. The input data for each context $i$ is represented as a pair $[x_i, y_i]$, where ${x_i} \in \mathbb{R}^{d_{\text{model}}}$ is the vector of hidden state activations and ${y_i} \in \mathbb{R}$ is the corresponding complexity score.

Following \citet{gurnee2023language}, we employ ridge regression to mitigate overfitting and multicollinearity issues common with high-dimensional activation vectors \citep{kim2025linear}. The objective is to find the weight vector $w \in \mathbb{R}^{d_{\text{model}}}$ and bias term $b \in \mathbb{R}$ that minimize the $L_2$-regularized squared loss:
\vspace{-8pt}
\begin{equation}
L(w, b) = \frac{1}{n} \sum_{i=1}^{n} (y_i - (w^T x_i + b))^2 + \frac{\lambda}{2} ||w||_2^2,
\end{equation}
where $n$ is the number of training contexts, and $\lambda$ is the regularization hyperparameter. The activation matrix $A \in \mathbb{R}^{n \times d_{\text{model}}}$ is constructed by concatenating these feature vectors, with the target defined as $y \in \mathbb{R}^{n}$. With implicit bias handling (\emph{e.g.}, by centering data or adding a feature column of ones to $A$), the optimal weight vector $\hat{w}$ is given by the closed-form solution \citep{belinkov2022probing}:
\begingroup
\setlength{\abovedisplayskip}{3pt}
\setlength{\belowdisplayskip}{3pt}
\begin{equation}
    \hat{w} = \left(A^{T}A + \lambda I\right)^{-1} A^{T}y.
\end{equation}
\endgroup
To determine the optimal $\lambda$, we perform 5-fold cross-validation. For each $\lambda$ in the set $\{0.001,\,0.01,\,0.1,\,1.0,\,10.0,\,100.0,\,1000.0\}$ , the probe is trained on four folds and evaluated on the held-out fold using the root mean squared error (RMSE). We select $\lambda = 100.0$, which yields the lowest average RMSE across the folds. The final probe, with parameters $(\hat{w}, \hat{b})$, is then trained on the entire dataset using this optimal $\lambda$. Subsequently, this trained probe is used to predict complexity scores for new contexts based on their last token activation $x$ via the linear function $\hat{y} = \hat{w}^{T}x + \hat{b}$. More details are given in Appendix~\ref{sec:linear-probes-details}.

\begin{figure*}[t]
  \centering
  \begin{subfigure}[b]{0.24\linewidth}
    \centering
    \includegraphics[width=\linewidth]{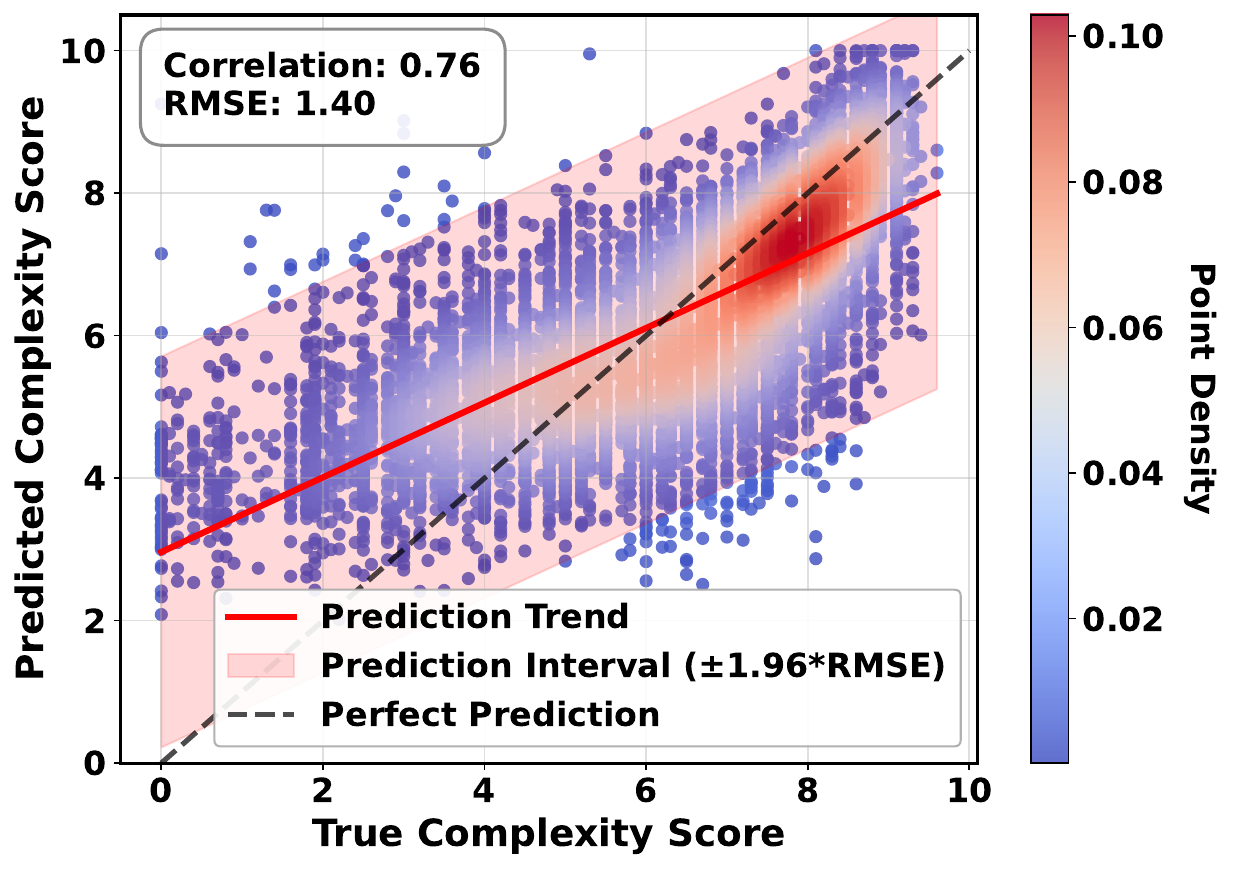}
    \caption{Pythia-70M}
    \label{fig:linear-70}
  \end{subfigure}
  \hfill
  \begin{subfigure}[b]{0.24\linewidth}
    \centering
    \includegraphics[width=\linewidth]{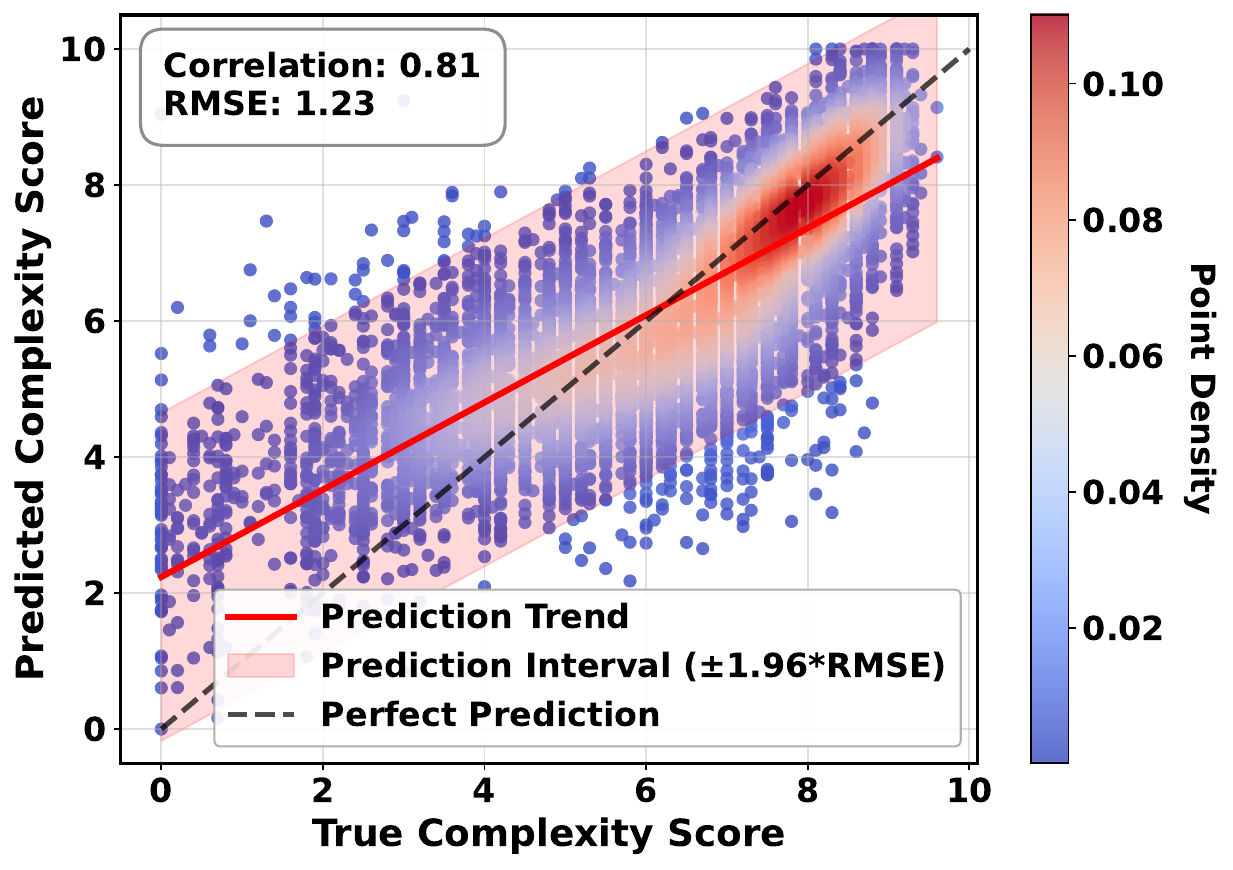}
    \caption{Gemma-2-2B}
    \label{fig:linear-gemma}
  \end{subfigure}
  \hfill
  \begin{subfigure}[b]{0.24\linewidth}
    \centering
    \includegraphics[width=\linewidth]{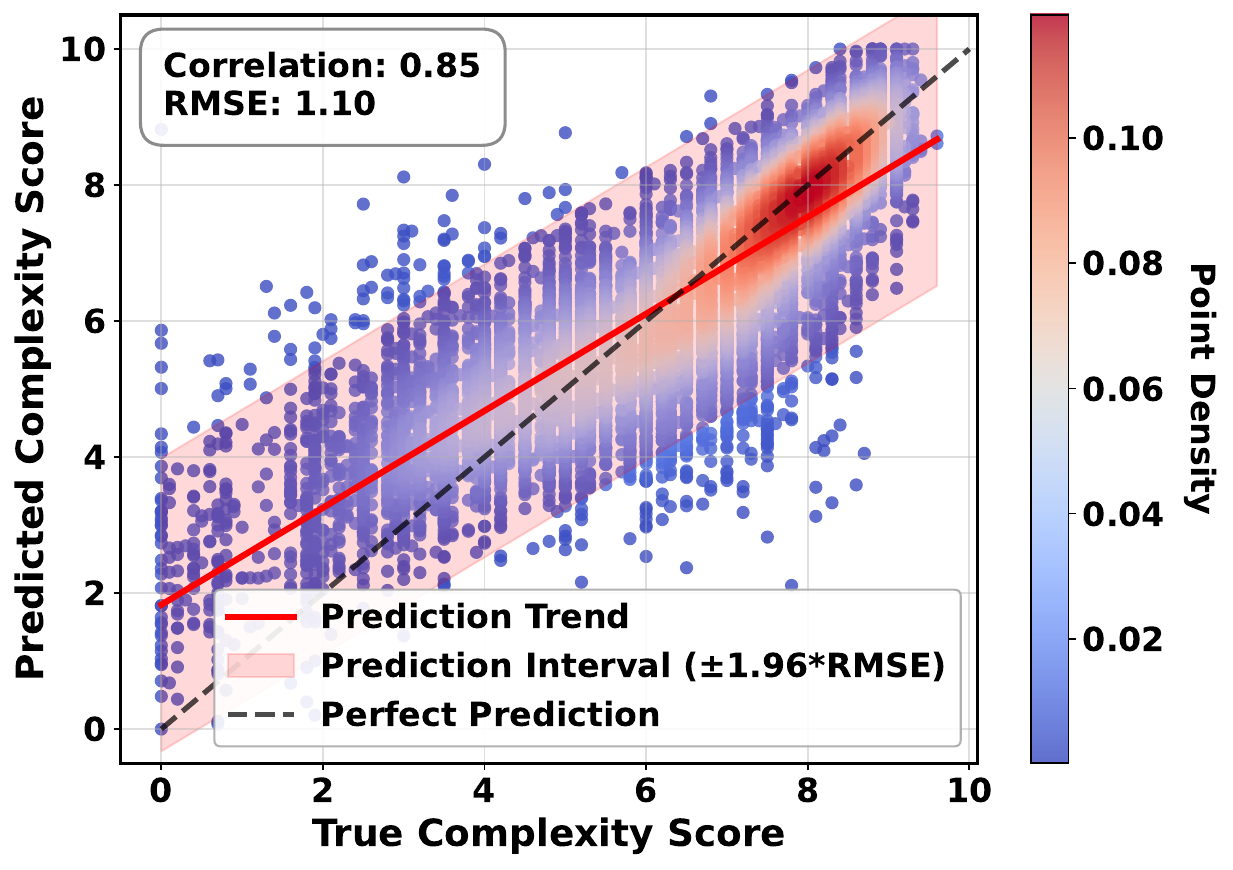}
    \caption{Qwen-3-8B}
    \label{fig:linear-qwen-8b}
  \end{subfigure}
  \hfill
  \begin{subfigure}[b]{0.24\linewidth}
    \centering
    \includegraphics[width=\linewidth]{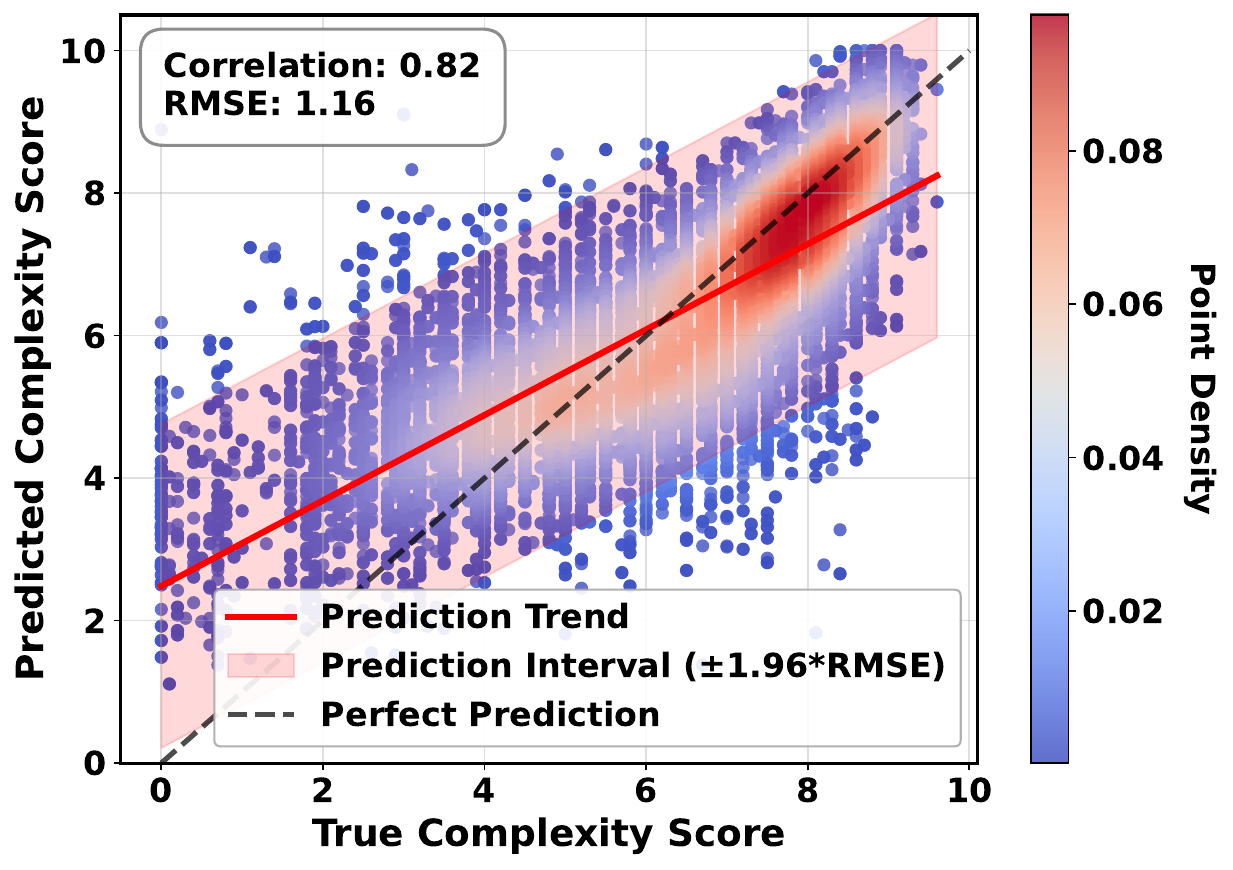}
    \caption{Phi-14B}
    \label{fig:linear-phi-14b}
  \end{subfigure}
  
  % \vspace{-5pt}
  \caption{Visualization of linear probe performance across different LLM scales. Points represent test contexts, with redder areas indicating higher sample density. The red line depicts predicted complexity trends. Most samples fall within prediction intervals, confirming the linear probe's effectiveness. Spearman Correlation and RMSE values (upper left) demonstrate improved prediction accuracy with increasing model scale. More results are in Fig. \ref{fig:more linear probe}.}
  % \vspace{-6pt}
  \label{fig:linear probe}
\end{figure*}

\begin{figure*}[t]
  \centering
  \begin{subfigure}[b]{0.24\linewidth}
    \centering
    \includegraphics[width=\linewidth]{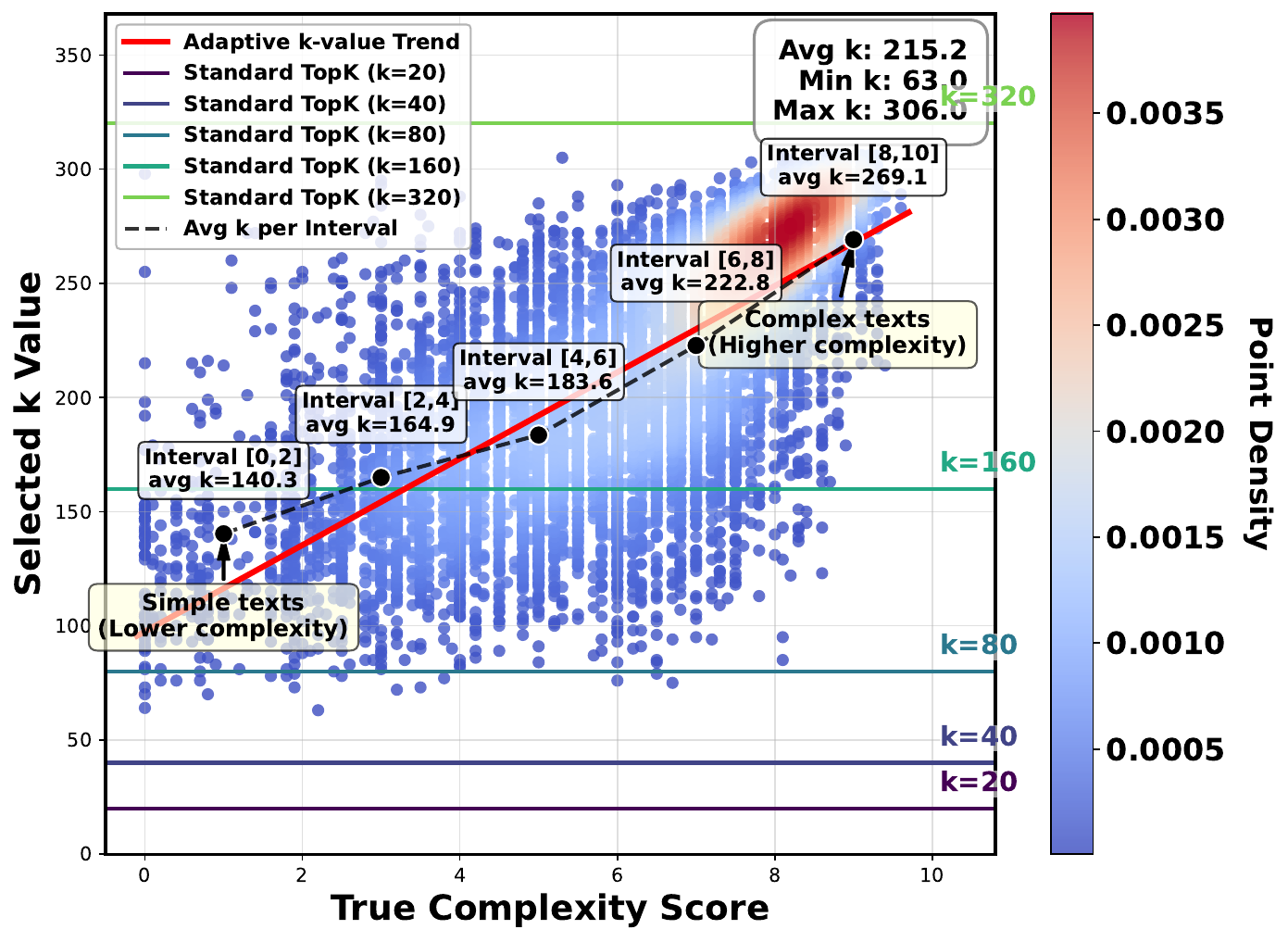}
    \caption{Pythia-70M}
    \label{fig:k-70}
  \end{subfigure}
  \hfill
  \begin{subfigure}[b]{0.24\linewidth}
    \centering
    \includegraphics[width=\linewidth]{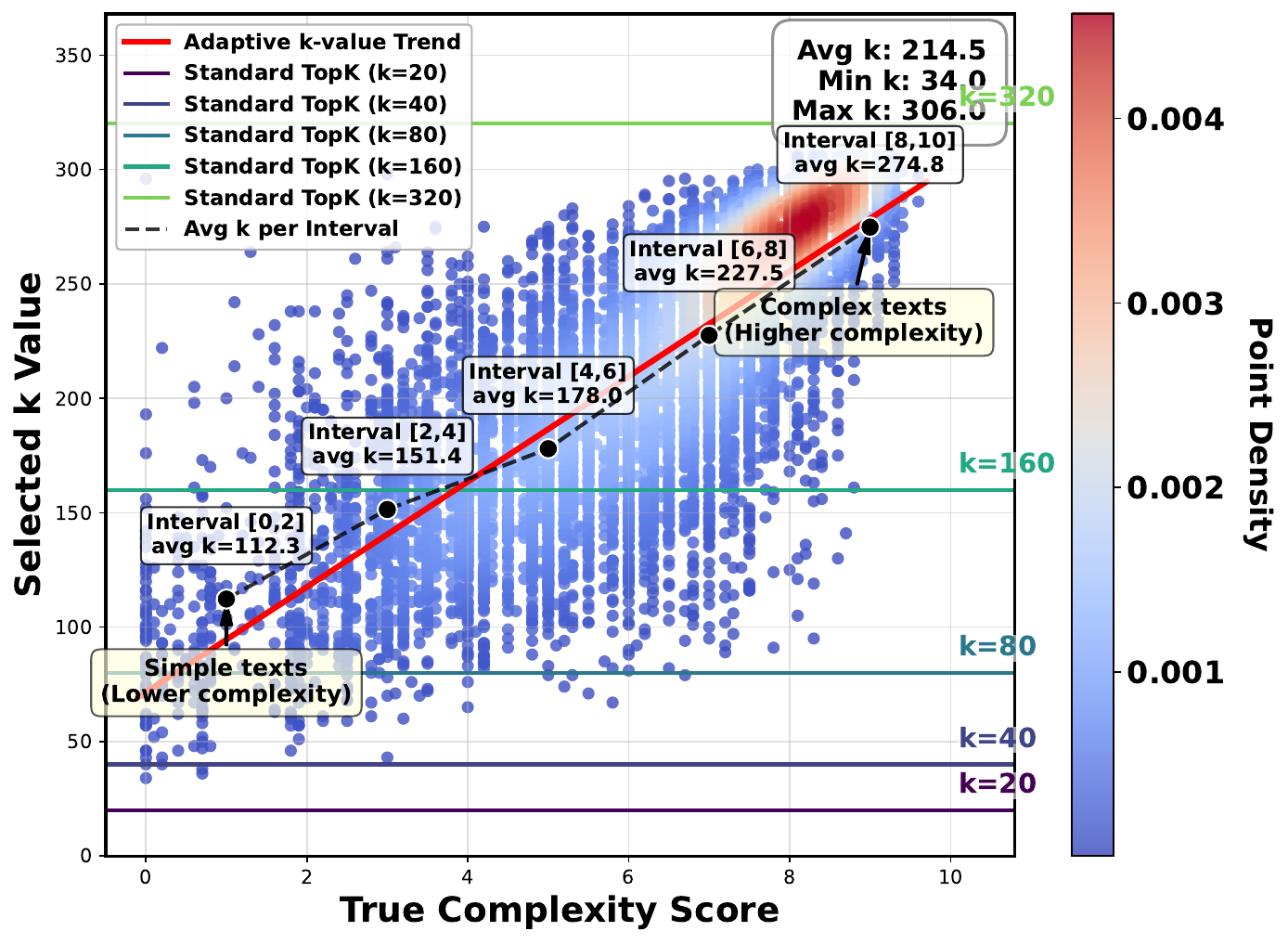}
    \caption{Gemma-2-2B}
    \label{fig:k-gemma}
  \end{subfigure}
  \hfill
  \begin{subfigure}[b]{0.24\linewidth}
    \centering
    \includegraphics[width=\linewidth]{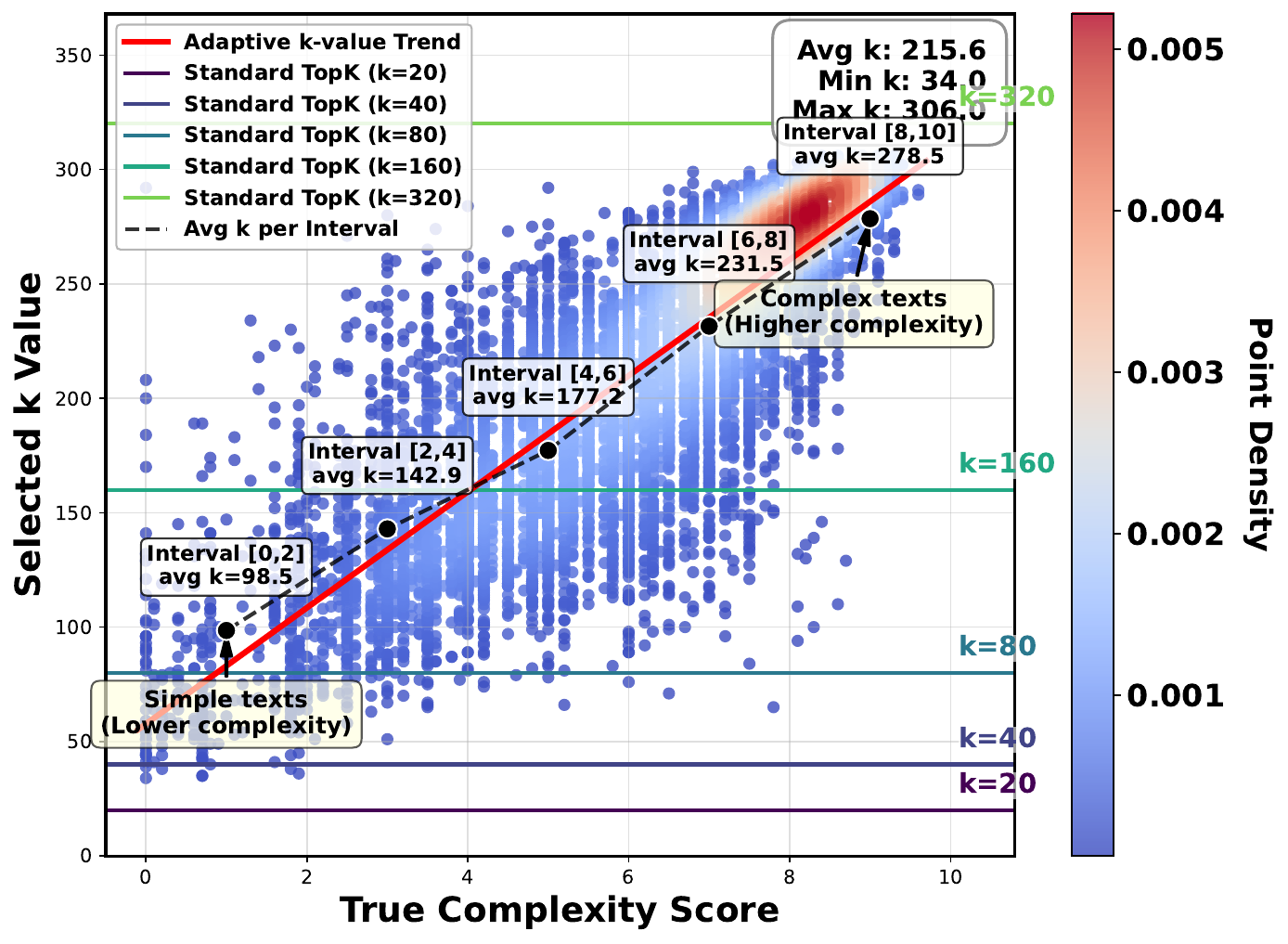}
    \caption{Qwen-3-8B}
    \label{fig:k-qwen-8b}
  \end{subfigure}
  \hfill
  \begin{subfigure}[b]{0.24\linewidth}
    \centering
    \includegraphics[width=\linewidth]{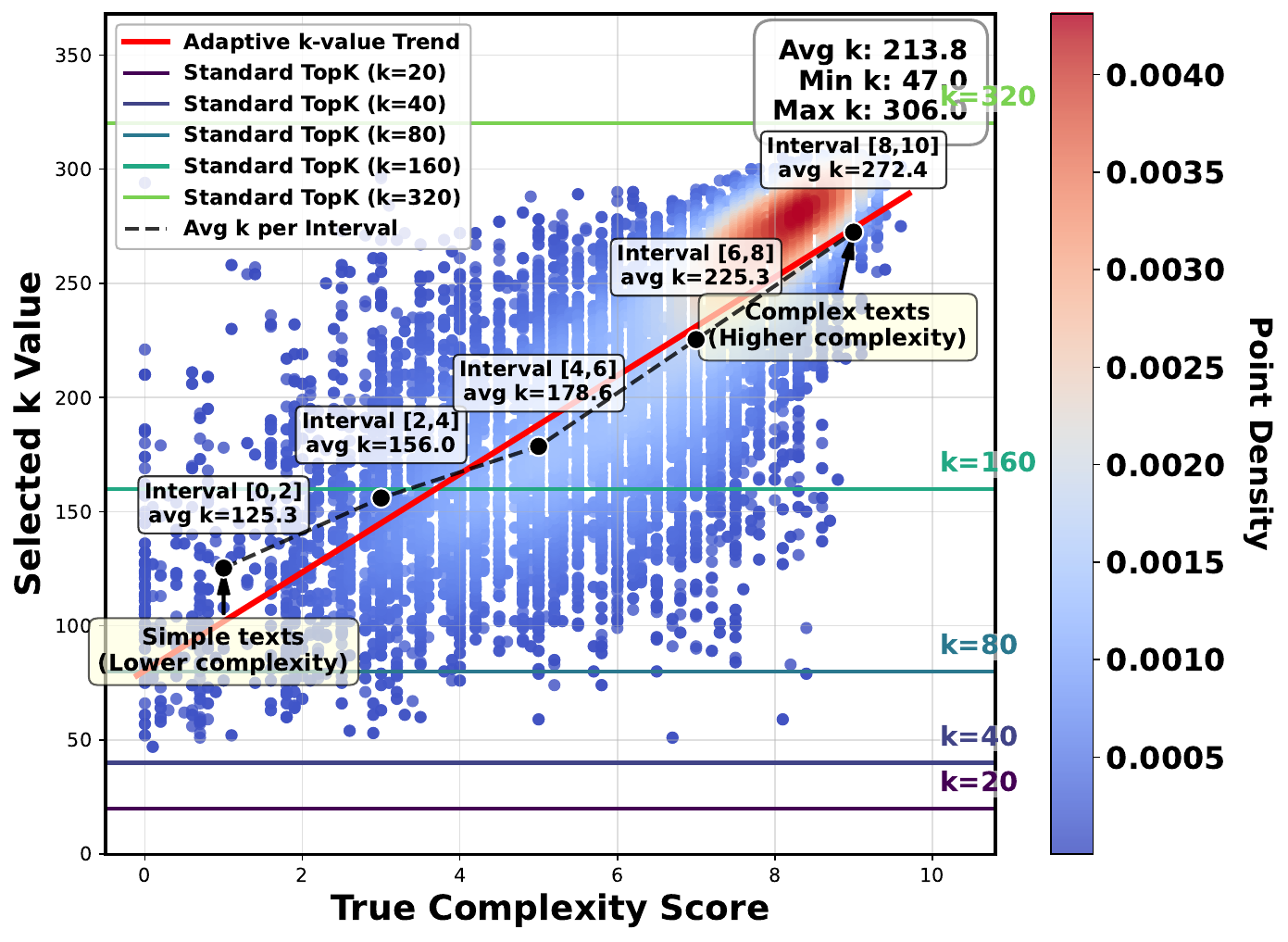}
    \caption{Phi-14B}
    \label{fig:k-phi-14b}
  \end{subfigure}
  
  % \vspace{-5pt}
  \caption{Visualization of Dynamic Feature Allocation by Text Complexity showing the relationship between complexity scores and allocated feature counts (K values). Average K values per complexity interval (connected by red lines) demonstrate that complex texts receive higher K allocations, with this relationship becoming increasingly linear as LLM scale grows. Horizontal lines indicate fixed Standard TopK baselines with K values on the right. More LLM results are in Fig. \ref{fig:more k}.}
  % \vspace{-15pt}
  \label{fig:k}
\end{figure*}

\begin{figure*}[t]
  \centering
  \begin{subfigure}[b]{0.24\linewidth}
    \centering
    \includegraphics[width=\linewidth]{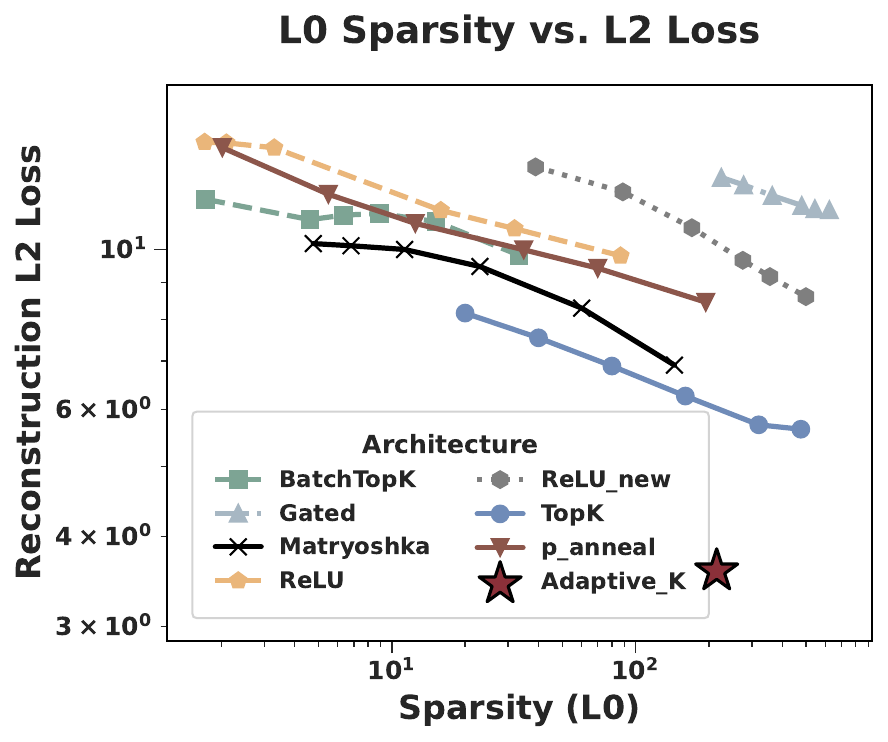}
    \caption{Pythia-70M}
  \end{subfigure}
  \hfill 
  \begin{subfigure}[b]{0.24\linewidth}
    \centering
    \includegraphics[width=\linewidth]{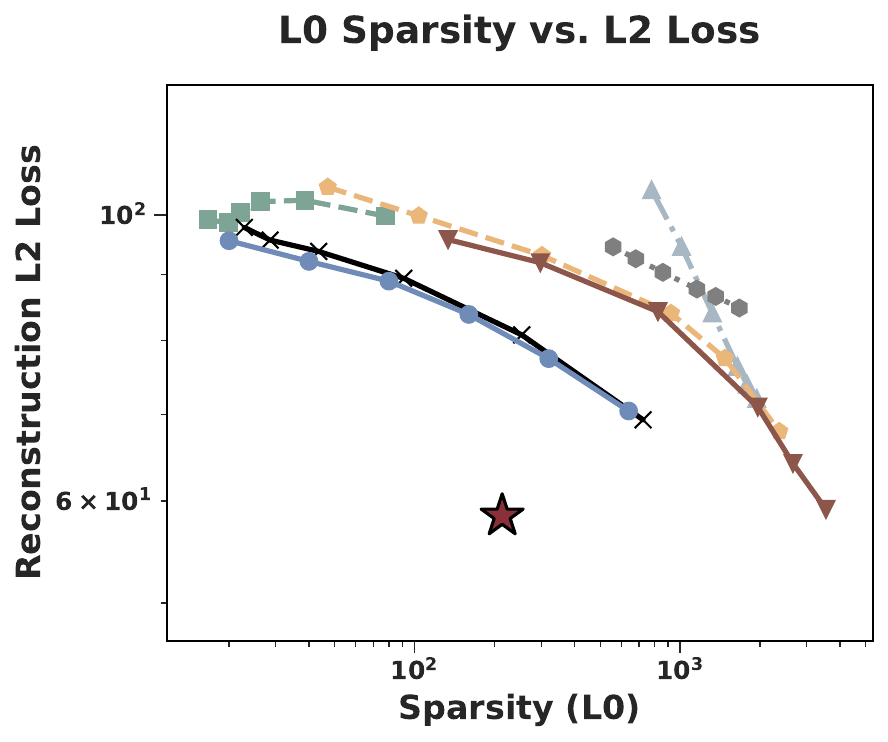}
    \caption{Gemma-2-2B}
  \end{subfigure}
  \hfill
  \begin{subfigure}[b]{0.24\linewidth}
    \centering
    \includegraphics[width=\linewidth]{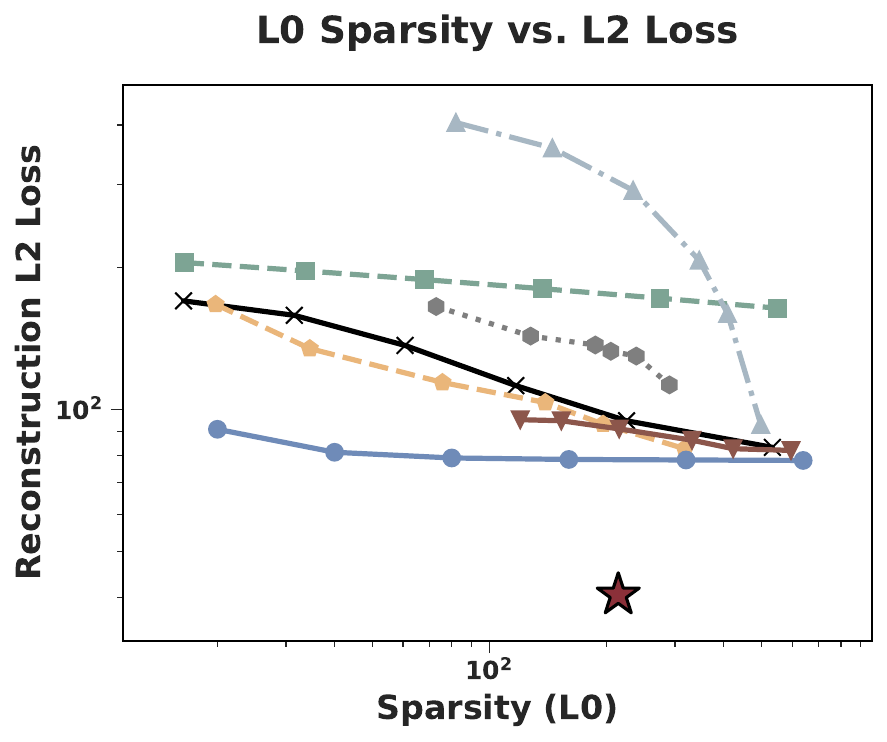}
    \caption{Qwen-3-8B}
  \end{subfigure}
  \hfill
  \begin{subfigure}[b]{0.24\linewidth}
    \centering
    \includegraphics[width=\linewidth]{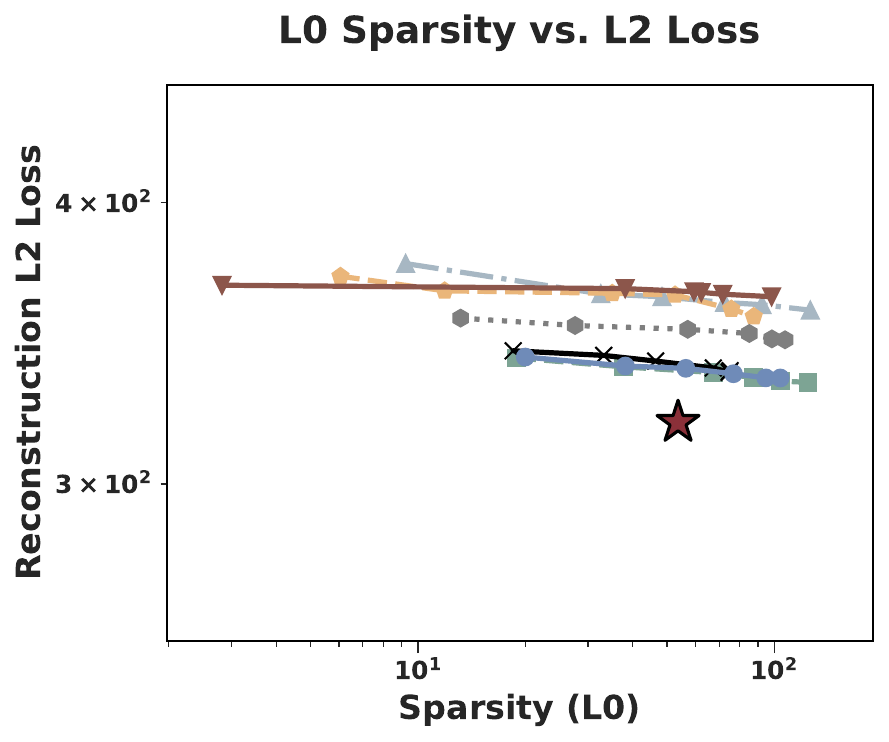}
    \caption{Phi-14B}
  \end{subfigure}
  
  % \vspace{-5pt}
  \caption{L2 Loss pareto frontier results. More LLM results are in Fig. \ref{fig:more pareto_l2}.}
  % \vspace{-10pt}
  \label{fig:pareto_l2}
\end{figure*}

\begin{figure*}[t]
  \centering
  \begin{subfigure}[b]{0.24\linewidth}
    \centering
    \includegraphics[width=\linewidth]{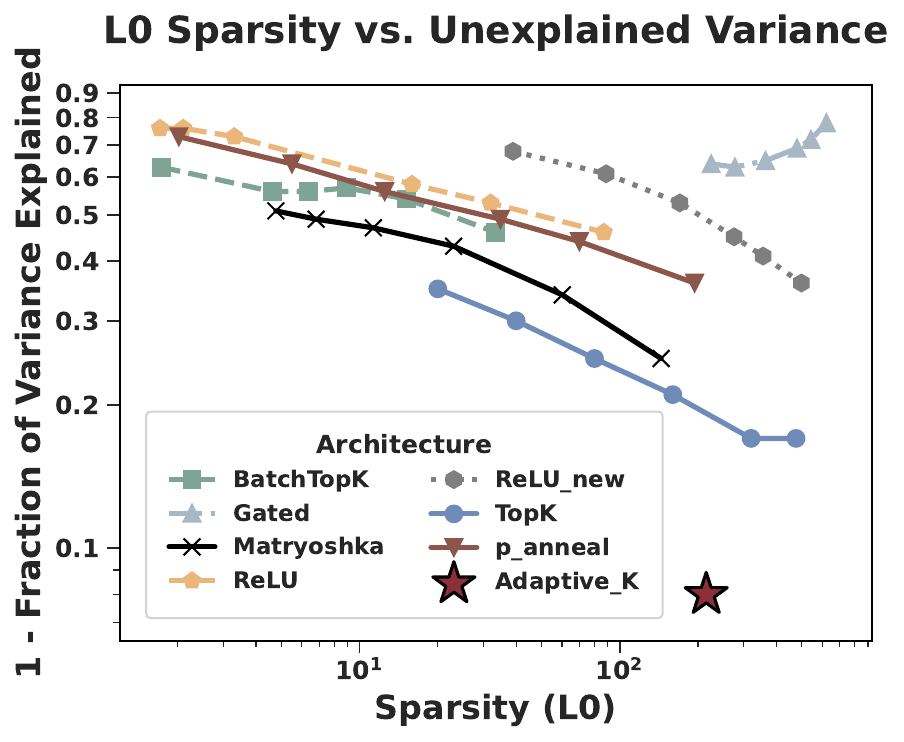}
    \caption{Pythia-70M}
  \end{subfigure}
  \hfill
  \begin{subfigure}[b]{0.24\linewidth}
    \centering
    \includegraphics[width=\linewidth]{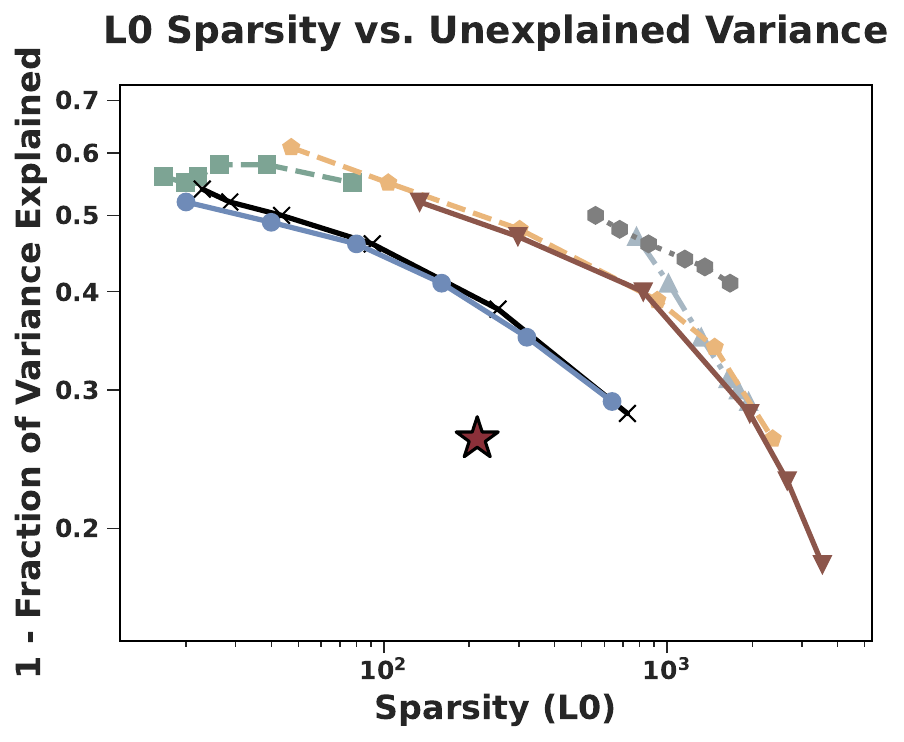}
    \caption{Gemma-2-2B}
  \end{subfigure}
  \hfill
  \begin{subfigure}[b]{0.24\linewidth}
    \centering
    \includegraphics[width=\linewidth]{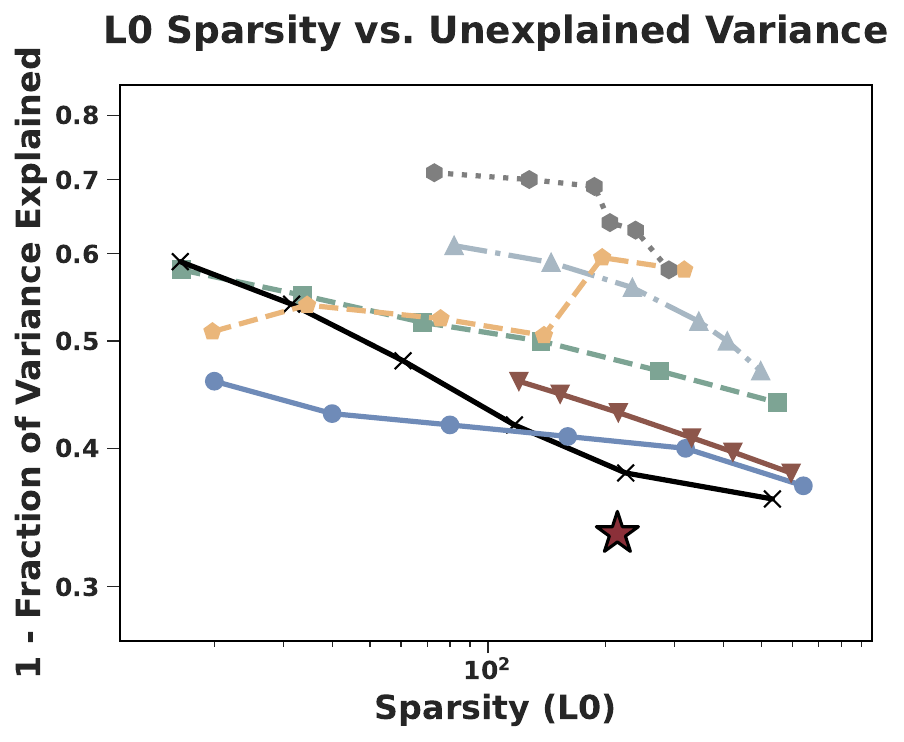}
    \caption{Qwen-3-8B}
  \end{subfigure}
  \hfill
  \begin{subfigure}[b]{0.24\linewidth}
    \centering
    \includegraphics[width=\linewidth]{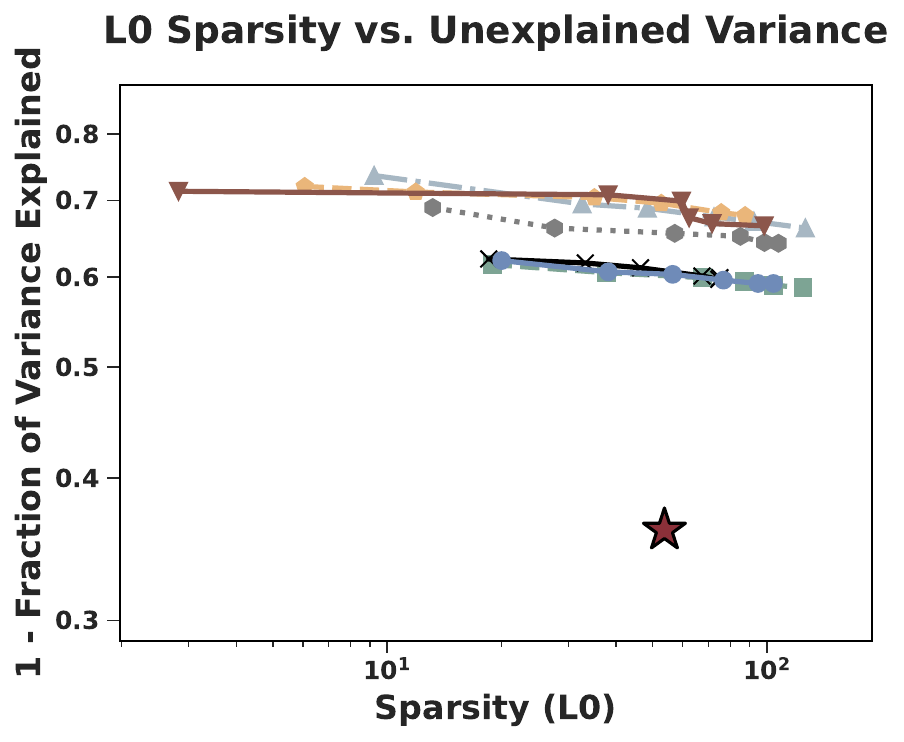}
    \caption{Phi-14B}
  \end{subfigure}
  
  % \vspace{-5pt}
  \caption{Unexplained Variance pareto frontier results. More LLM results are in Fig. \ref{fig:more pareto_unexplained}.}
  % \vspace{-15pt}
  \label{fig:pareto_unexplained}
\end{figure*}

\noindent
\textbf{Linear Probe Evaluation.}
As mentioned in Appendix \ref{subsec: linear probe}, many high-level features have been demonstrated to exist linearly in LLM representation spaces. However, context complexity is indeed multifaceted, spanning lexical, syntactic, and other linguistic dimensions, and thus differs from the single-attribute features studied in prior work.

By comparing the performance of linear probes, MLP, and XGBoost in predicting context complexity, we provide evidence for the linear encoding of context complexity features in language model representation spaces. We contrasted a single-hidden-layer MLP with the structure $f(\mathbf{x}) = \mathbf{W}_2\text{ReLU}(\mathbf{W}_1\mathbf{x} + \mathbf{b}_1) + \mathbf{b}_2$ containing 256 neurons in hidden layer, and an XGBoost model that minimizes error through gradient boosting across multiple decision trees, with the prediction formula $\hat{y}_i = \sum_{k=1}^K f_k(\mathbf{x}_i)$, against our linear probe.

\begin{table}[t]
  \centering
  \caption{Comparison of Linear Probe, MLP, and XGBoost models for context complexity prediction on Pythia-70M, trained on 250,000 contexts and evaluated on 10,000 test samples using three metrics.}
  \label{tab:probe-comparison}
  \scalebox{0.8}{
  \begin{tabular}{lccc}
    \toprule
    Probe & RMSE & Pearson & Spearman \\
    \midrule
    \cem Linear   & \cem {1.41} & \cem {0.72}  & \cem {0.76}  \\
    MLP      & 1.37 & 0.74 & 0.77 \\
    XGBoost  & 1.42 & 0.71  & 0.74  \\
    \bottomrule
  \end{tabular}}
  \vspace{-8pt}
\end{table}
As shown in Tab.~\ref{tab:probe-comparison}, linear model is comparable to theoretically more expressive nonlinear models like XGBoost across all three metrics: Pearson and Spearman correlations and RMSE. These findings extend the applicability of the linear representation hypothesis to multifaceted features such as context complexity. Fig. \ref{fig:linear probe} visualizes our probe's performance on test data across eight models. Each point represents one context from the test set, showing the predicted versus true complexity score. With most points falling within the prediction interval (±1.96 $\times$ RMSE), this performance confirms that context complexity is largely encoded linearly in representation space.

\subsection{AdaptiveK SAE Architecture}
\label{sec:adaptive-k-architecture}

% 为了解决上述劣势，一些motivation

Unlike existing SAEs apply uniform sparsity constraints across all inputs, requiring extensive hyperparameter experimentation, our AdaptiveK Sparse Autoencoder incorporates a complexity estimation component that adaptively determines the appropriate sparsity level for each context. The overall pipeline is shown in Fig. \ref{fig:overall_pipeline}.

For an input activation vector $x \in \mathbb{R}^d$, we compute a complexity score using the linear probe from Sec. \ref{Linear Probe}:
$
    c = \hat{w}^{T}x + \hat{b},
$
where $\hat{w} \in \mathbb{R}^d$ and $\hat{b} \in \mathbb{R}$ are trained from the ridge regression. This score is then mapped to a sparsity level $k_{\mathrm{adp}}$ through a sigmoid-based transformation:
% \begin{equation}
% k_{\mathrm{adp}}=k_{\min }+\frac{1}{1+e^{-s \cdot\left(\frac{c-c_{\min }}{c_{\max }-c_{\min }}-0.5\right)}} \cdot\left(k_{\max }-k_{\min }\right),
% \end{equation}
\begingroup
\small
\setlength{\abovedisplayskip}{5pt}
\setlength{\belowdisplayskip}{5pt}
\begin{equation}
k_{\mathrm{adp}} = k_{\min}
 + \frac{1}{1 + e^{-s\!\left(\frac{c-c_{\min}}{c_{\max}-c_{\min}} - 0.5\right)}}
   \,(k_{\max}-k_{\min}).
\end{equation}
\endgroup
where $k_{\min}$ and $k_{\max}$ define the range of possible $k_{\mathrm{adp}}$ values, $c_{\min}$ and $c_{\max}$ represent the minimum and maximum complexity scores, and $s$ controls the steepness of the sigmoid function.
The sparse autoencoder component then processes the input with an adaptive TopK activation function:
% \begin{equation}
% z=\operatorname{TopK}\left(W_{\mathrm{enc}}\left(x-b_{\mathrm{pre}}\right), k_{\mathrm{adp}}\right),
% \end{equation}
\begingroup
\setlength{\abovedisplayskip}{5pt}
\setlength{\belowdisplayskip}{5pt}
\begin{equation}
z=\operatorname{TopK}\left(W_{\mathrm{enc}}\left(x-b_{\mathrm{pre}}\right), k_{\mathrm{adp}}\right),
\end{equation}
\endgroup
where $\text{TopK}(\cdot, k_{\mathrm{adp}})$ retains only the $k_{\mathrm{adp}}$ largest activations and sets all others to zero. The encoder matrix $W_{\text{enc}} \in \mathbb{R}^{M \times d}$, decoder matrix $W_{\text{dec}} \in \mathbb{R}^{d \times M}$, and bias vector $b_{\text{pre}} \in \mathbb{R}^d$ are trainable parameters. Output $\hat{x}$ follows Equation \ref{equ:decoder}.

Therefore, AdaptiveK SAE eliminates the computational burden of training separate models for each sparsity level to find the optimal trade-off, requiring only a single training run. Also, it addresses the feature suppression problem that occurs with L1 penalties while improving performance on context-level tasks by allocating representational capacity proportional to context complexity. 
%Experiments in Section \ref{Results} demonstrate that this complexity-driven adaptation achieves better reconstruction without requiring extensive hyperparameter tuning.

\subsection{AdaptiveK SAE Training}
\label{sec:adaptive-k-training}
\begin{algorithm}[h]\small
\DontPrintSemicolon
\KwIn{Activation data $D = \{x_i\}_{i=1}^N$, random initialized SAE.}
// Phase 1: Complexity probe pretraining\;
Train linear probe to predict complexity scores from activations\;
// Phase 2: SAE training with frozen probe\;
\While{not converged}{
  Apply adaptive sparsity constraints based on complexity\;
  Update SAE parameters using $L_{SAE} = L_{recon} + \alpha L_{sparsity} + \beta L_{aux}$\;
}
// Phase 3: Joint fine-tuning\;
\While{not converged}{
  Update all parameters using $L_{joint} = L_{SAE} + \gamma(L_{probe} + \delta L_{deviation})$\;
}
\KwOut{Trained AdaptiveK SAE.}
\caption{AdaptiveK SAE Training}
\label{alg:adaptiveK}
\end{algorithm}

We employ a three-phase training for AdaptiveK SAE (see Algorithm~\ref{alg:adaptiveK}). First, we pretrain the complexity probe as described in Sec. \ref{Linear Probe}.

In the second phase, we freeze the probe parameters and train only the SAE components using:
% \begin{equation}
% L_{\text{SAE}} = L_{\text{recon}} + \alpha L_{\text{sparsity}} + \beta L_{\text{aux}},
% \end{equation}
\begingroup
\setlength{\abovedisplayskip}{2pt}
\setlength{\belowdisplayskip}{2pt}
\begin{equation}
L_{\text{SAE}} = L_{\text{recon}} + \alpha L_{\text{sparsity}} + \beta L_{\text{aux}},
\end{equation}
\endgroup
where $L_{\text{recon}} = |x - \hat{x}|^2_2$ is the reconstruction loss, $L_{\text {sparsity }}=\frac{\|z\|_1}{\|x\|_2}$ is the normalized $L_1$ penalty with weight $\alpha$ = 0.005, and $L_{\text{aux}}$ is the auxiliary loss with weight $\beta$ = 1/32 for reactivating dead features. We set $base\_k$ = 80, $min\_k$ = 20, and $max\_k$ = 320 for the sparsity range.

In the final joint fine-tuning phase, we jointly optimize both components with:
% \begin{equation}
% L_{\text{joint}} = L_{\text{SAE}} + \gamma(L_{\text{probe}} + \delta L_{\text{deviation}}),
% \end{equation}
\begingroup
\setlength{\abovedisplayskip}{3pt}
\setlength{\belowdisplayskip}{3pt}
\begin{equation}
L_{\text{joint}} = L_{\text{SAE}} + \gamma(L_{\text{probe}} + \delta L_{\text{deviation}}),
\end{equation}
\endgroup
where $\gamma$ = 0.9 controls the probe loss weight and $L_{\text{deviation}} = |w - w^0|_2 + |b - b^0|$ penalizes deviations from pretrained parameters, initially with $\delta$ = 0.2. This penalty prevents the SAE's reconstruction objective from corrupting the probe's complexity mapping. We adaptively adjust $\delta$ between 0.01 and 0.5, decreasing $\delta$ when probe loss improves and increasing $\delta$ when it stagnates.

Our implementation uses a AdaptiveKBuffer that extracts last-token representations from contexts (see Sec. \ref{Training Settings} for details), reducing memory usage while tracking complexity scores for balanced training. Adam optimizer \citep{kingma2014adam} is applied with learning rate 1e-3, warm-up over 15 steps, and linear decay starting at 70\% of training. More training details are in Appendix \ref{sec:additional-sae-training-appendix}.

\section{Experiments}
\label{Results}
In this section, we evaluate our AdaptiveK to answer the following research questions (RQs):
\vspace{-7pt}
\begin{itemize}[leftmargin=*]\setlength\itemsep{-0.3em}
    \item \textbf{RQ1:} How does the relationship between text complexity and adaptive k-values manifest across different language model scales?
    \item \textbf{RQ2:} To what extent does our adaptive sparsity mechanism improve reconstruction quality metrics (L2 loss, variance explained, and cosine similarity) compared to fixed-sparsity approaches?
    \item \textbf{RQ3:} How does AdaptiveK SAE's performance on the Pareto frontier balance the trade-off between sparsity and reconstruction fidelity compared to baseline methods?
    \item \textbf{RQ4:} How is LLM interpretability defined, and how can the interpretability of AdaptiveK SAE be measured at a human-understandable level?
\end{itemize}

\begin{figure*}[t]
  \centering
  \begin{subfigure}[b]{0.24\linewidth}
    \centering
    \includegraphics[width=\linewidth]{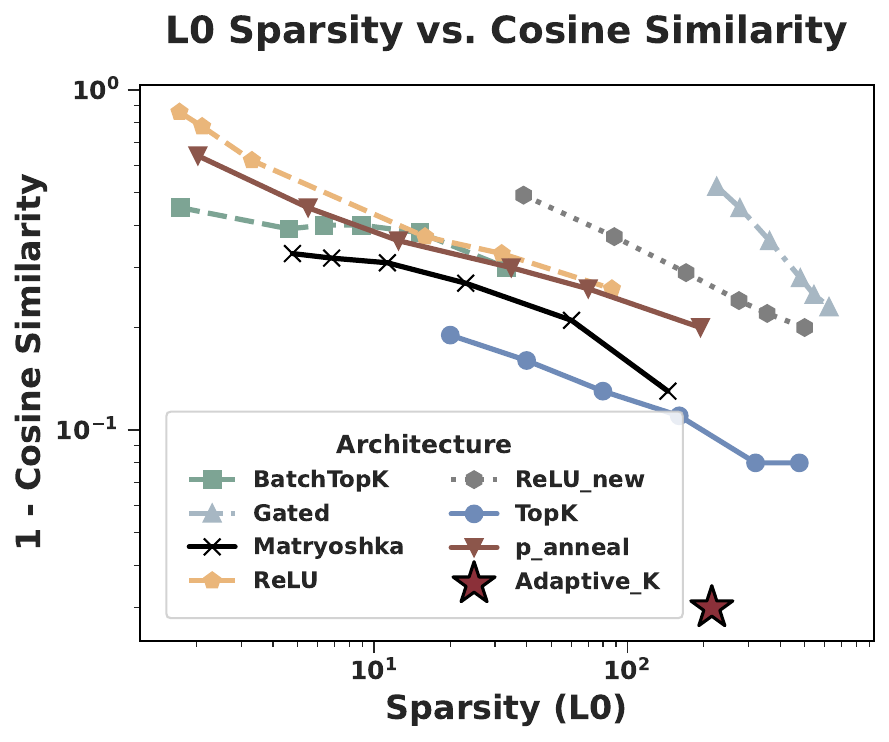}
    \caption{Pythia-70M}
  \end{subfigure}
  \hfill
  \begin{subfigure}[b]{0.24\linewidth}
    \centering
    \includegraphics[width=\linewidth]{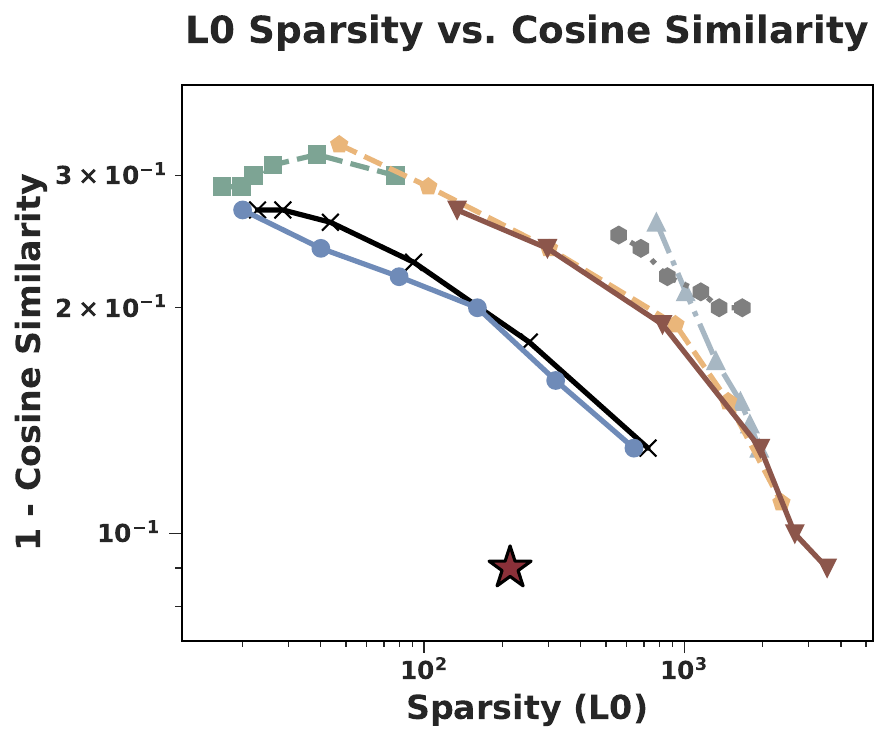}
    \caption{Gemma-2-2B}
  \end{subfigure}
  \hfill
  \begin{subfigure}[b]{0.24\linewidth}
    \centering
    \includegraphics[width=\linewidth]{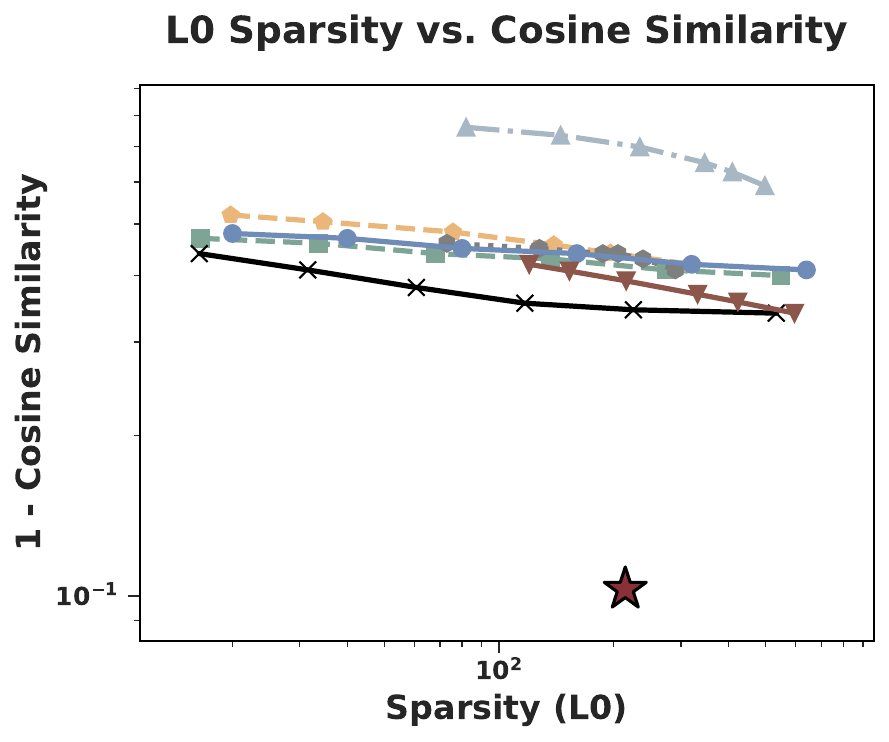}
    \caption{Qwen-3-8B}
  \end{subfigure}
  \hfill
  \begin{subfigure}[b]{0.24\linewidth}
    \centering
    \includegraphics[width=\linewidth]{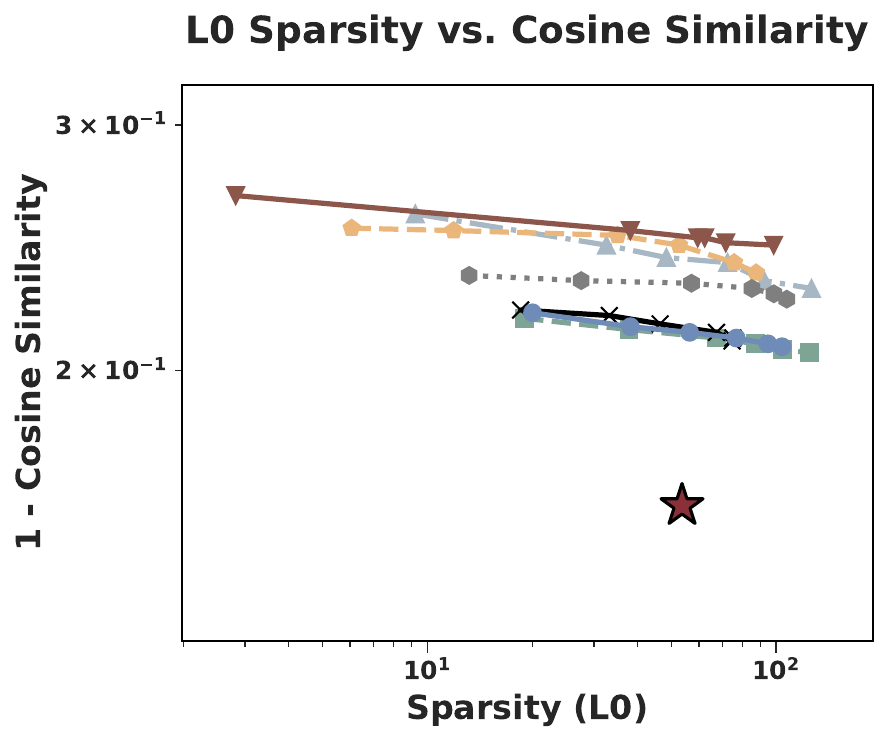}
    \caption{Phi-14B}
  \end{subfigure}
  
  % \vspace{-5pt}
  \caption{Cosine Similarity pareto frontier results. More LLM results are in Fig. \ref{fig:more pareto_cosine}.}
  % \vspace{-5pt}
  \label{fig:pareto_cosine}
\end{figure*}

% \begin{figure*}[h]
%   \centering
%   \includegraphics[width=0.7\textwidth]{figs/maxact.pdf}
%   \vspace{-7pt}
%   \caption{Input-based interpretability analysis using MaxAct method.}
%   \vspace{-15pt}
%   \label{fig:maxact}
% \end{figure*}
% \begin{figure*}[h]
%   \centering
%   \includegraphics[width=0.8\textwidth]{figs/vocabproj.pdf}
%   \vspace{-8pt}
%   \caption{Output-based interpretability analysis using VocabProj method.}
%   \vspace{-10pt}
%   \label{fig:vocabproj}
% \end{figure*}

\subsection{Experimental Settings}
We train SAEs on 8 language models of increasing scale: Pythia (70M, 160M), Gemma-2 (2B, 9B), Llama-3.1 (8B), Qwen-3 (8B, 14B) and Phi-4 (14B), with hidden dimensions from 512 to 5120. All SAEs have a dictionary size of 16384, i.e., 16k latents. Unlike token-level approaches, we operate at context level (though we demonstrate in Appendix \ref{token-level evaluation} that token-level evaluation is also applicable) using pile-uncopyrighted data \citep{gao2020pile}: 250,000 training and 10,000 test contexts, each with 2048 tokens. We train on the last token representation of each context, which captures accumulated contextual information for complexity-driven sparsity adaptation.

For experimental evaluation, we compare our AdaptiveK SAE against 7 baselines: (1) ReLU SAEs \citep{bricken2023towards}, using ReLU activation with L1 penalty; (2) refined ReLU\_new SAEs \citep{anthropic2024circuits}; (3) TopK SAEs \citep{gao2024scaling}, which select the K highest activations; (4) BatchTopK SAEs \citep{bussmann2024batchtopk}, extending TopK across batches; (5) Gated SAEs \citep{rajamanoharan2024improving}, decoupling feature selection from magnitude estimation; (6) P-anneal SAEs \citep{karvonen2024measuring}, using annealing $L_p$ norm penalties; and (7) Matryoshka SAEs \citep{bussmannlearning}, training nested dictionaries with increasing capacity.

Beyond the primary metrics, we evaluated AdaptiveK using SAEBench metrics \citep{karvonen2025saebench} (Appendix \ref{Other Evaluation Metrics}). Additional analyses include layer-wise performance (Appendix \ref{analysis of layers}), extensions to encoder-only and encoder-decoder models (Appendix \ref{encoder-decoder}), hyperparameter (Appendix \ref{hyperparameters}), and training efficiency analysis (Appendix \ref{training efficiency}).

\subsection{Relation of Complexity and k-Values}
\label{relationship}
We plotted the relationship between true complexity and k-value selection, calculated average k-value per complexity interval, and marked fixed k-values that TopK would select. Fig. \ref{fig:k} shows an approximately linear relationship between text complexity and allocated k-values, which becomes increasingly evident as model scale increases. 

% 结合前面，探针有效性
On one hand, Fig. \ref{fig:linear probe} reveals that both predicted complexity and activated feature count increase with sample complexity, validating that complex texts correctly receive larger k values and activate more features. On the other hand, Fig. \ref{fig:pareto_l2} indicate that that AdaptiveK consistently achieves lower reconstruction errors than TopK SAE across all k-values, demonstrating the benefit of adaptive resource allocation by using fewer features for simple texts and more for complex ones. Additional probe evaluations using PCA and layer-wise analysis are in Appendices \ref{PCA} and \ref{layer-wise probe}.

\subsection{Pareto Frontier Results}
\label{pareto}
We evaluated three different metrics vs. sparsity frontier for benchmark SAEs with different sparsity constraints. All SAEs have a dictionary size of 16384. These three metrics are:
\vspace{-0.5em}
\begin{itemize}[leftmargin=*]\setlength\itemsep{-0.5em}
    \item \textbf{Reconstruction L2 Loss:} $\|x-\hat{x}\|_2^2$ measures squared Euclidean distance between reconstruction and input, lower is better.

    \item \textbf{Fraction of Variance Explained:} $\frac{\mathrm{Var}(x-\hat{x})}{\mathrm{Var}(x)}$ captures input variability. Plots show 1 minus this (unexplained variance, lower is better).

    \item \textbf{Cosine Similarity:} $\frac{x\cdot\hat{x}}{\|x\|_2\|\hat{x}\|_2}$ measures directional fidelity. Plots show 1 minus this (lower is better).

\end{itemize}
\vspace{-0.5em}

Fig. \ref{fig:pareto_l2}, \ref{fig:pareto_unexplained}, and \ref{fig:pareto_cosine} show AdaptiveK consistently outperforms all other SAEs across different LLMs in reconstruction error, cosine similarity, and explained variance at equivalent sparsity. While some SAEs match or exceed these metrics at extremely high sparsity (over 10$\times$ greater), such cases violate the sparsity-fidelity tradeoff since infinite-width SAE can theoretically reconstruct perfectly \citep{karvonen2025saebench}. Thus, AdaptiveK transcends traditional Pareto Frontier, achieving results unreachable by other SAEs.

Notably, across all model scales, AdaptiveK's reconstructed activations more accurately match the originals in both distance and direction (Fig. \ref{fig:pareto_l2}, \ref{fig:more pareto_l2}, \ref{fig:pareto_cosine}, \ref{fig:more pareto_cosine}) while explaining a higher proportion of data variance (Fig. \ref{fig:pareto_unexplained}, \ref{fig:more pareto_unexplained}), demonstrating AdaptiveK's robustness and generalizability.

\subsection{Interpretability Analysis}
\noindent\textbf{Definition of LLM Interpretability.}
Following established mechanistic interpretability research \citep{rai2025practicalreviewmechanisticinterpretability, shu2025surveysparseautoencodersinterpreting, zhao2023explainabilitylargelanguagemodels}, we define LLM interpretability as how well learned features correspond to semantically coherent, human-understandable concepts consistently identified across different contexts. A feature is interpretable if it satisfies: (1) input-based semantic coherence: it activates on inputs sharing clear conceptual relationships, and (2) output-based human comprehensibility: it influences semantically consistent vocabulary prediction.

\begin{figure*}[h]
  \centering
  \includegraphics[width=0.98\textwidth]{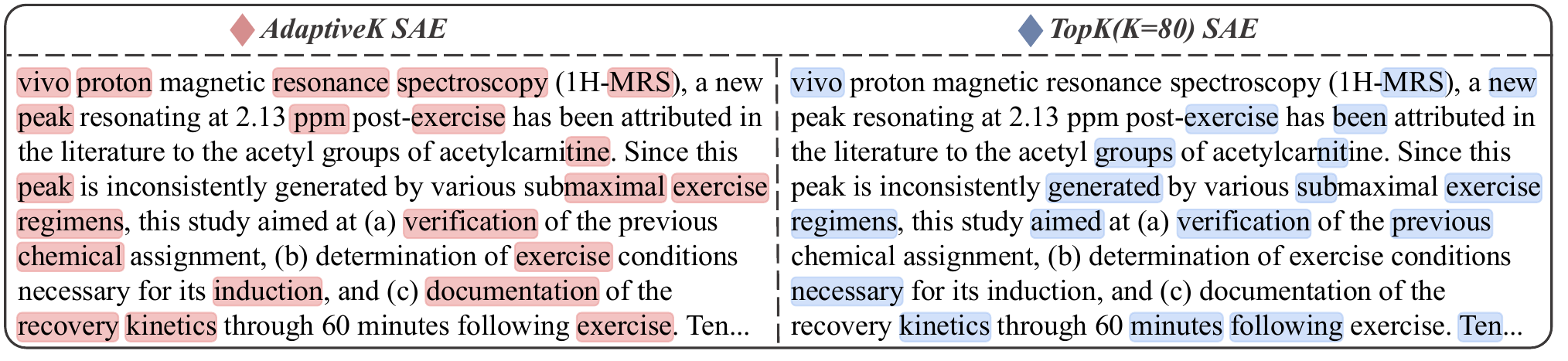}
  % \vspace{-7pt}
  \caption{Input-based interpretability analysis using MaxAct method.}
  % \vspace{-15pt}
  \label{fig:maxact}
\end{figure*}
\begin{figure*}[h]
  \centering
  \includegraphics[width=0.98\textwidth]{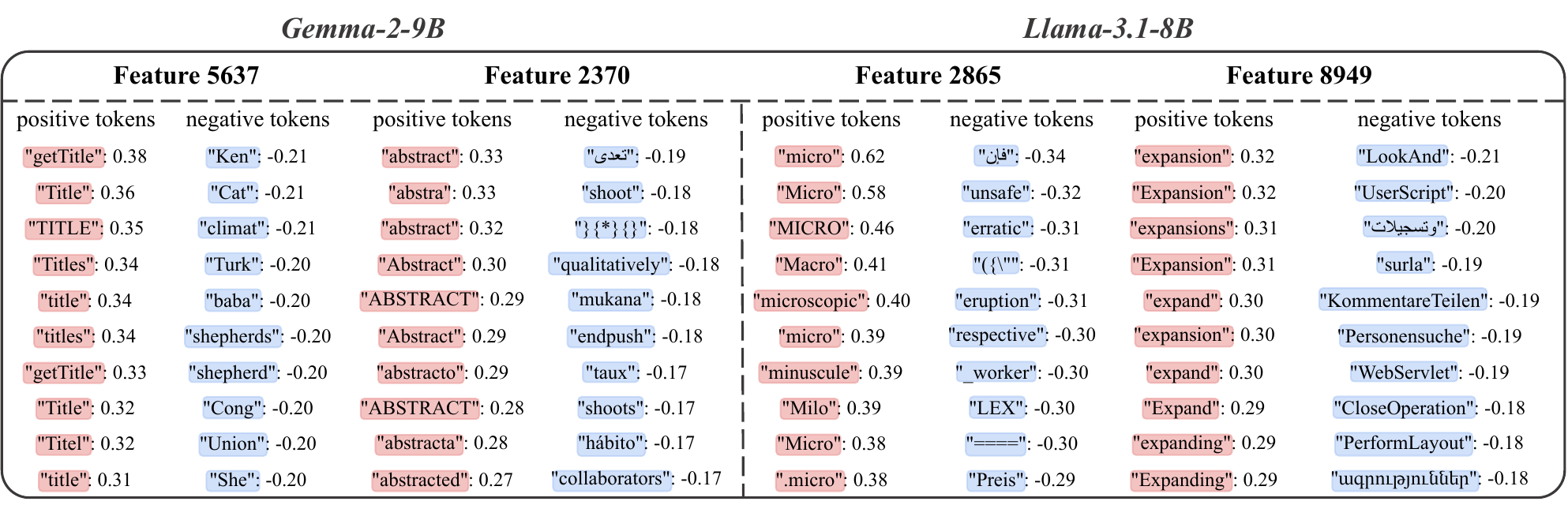}
  % \vspace{-8pt}
  \caption{Output-based interpretability analysis using VocabProj method.}
  % \vspace{-10pt}
  \label{fig:vocabproj}
\end{figure*}

\noindent\textbf{Measures of Interpretability.}
For input-based analysis, we implemented MaxAct \citep{bricken2023towards} to identify text segments maximally activate each SAE feature. Specifically, for each target feature $w_m$, we fed corpus texts $x$ from our test set into the language model to obtain hidden representations $\mathbf{h}_x=f_{<l}(x)$ at layer $l$, computed feature activation strengths $a_m(x) = \operatorname{ReLU}\left(\mathbf{w}_m^{\top} \cdot \mathbf{h}_x+b_m\right)$ through SAE encoder, and selected the top-$K$ text segments with highest activations:
% \begin{equation}
% \mathcal{I}_m=\underset{\left|\mathcal{X}^{\prime}\right|=K}{\arg \max } \sum_{x \in \mathcal{X}^{\prime}} a_m(x)
% \end{equation}
\begingroup
\setlength{\abovedisplayskip}{3pt}
\setlength{\belowdisplayskip}{2pt}
\begin{equation}
\mathcal{I}_m=\underset{\left|\mathcal{X}^{\prime}\right|=K}{\arg \max } \sum_{x \in \mathcal{X}^{\prime}} a_m(x)
\end{equation}
\endgroup
Through experiments on Gemma-2-9B comparing against TopK SAE (k=80), we analyzed thematic consistency of high-activation texts to quantify whether features learned monosemantic, human-interpretable concepts. Fig. \ref{fig:maxact} shows that for biomedical texts, AdaptiveK's activation patterns focus precisely on core professional concepts, including technical terms (``resonance spectroscopy,'' ``MRS,'' ``ppm''), biochemical concepts (``acetylcarnitine,'' ``peak''), and methodological vocabulary (``verification,'' ``chemical,'' ``induction''). In contrast, TopK activates both similar professional terminology and numerous semantically weak function words (``been,'' ``groups,'' ``generated,'' ``aimed''), diluting feature focus and inefficient computational resource utilization with diminishing semantic returns despite employing more features. This demonstrates AdaptiveK's ability to identify semantic complexity and allocate appropriate features while avoiding irrelevant activations.

% \begin{figure*}[t]
%   \centering
%   \includegraphics[width=0.5\textwidth]{figs/maxact.pdf}
%   \vspace{-8pt}
%   \caption{Input-based interpretability analysis using MaxAct method.}
%   \vspace{-10pt}
%   \label{fig:maxact}
% \end{figure*}
% \begin{figure*}[t]
%   \centering
%   \includegraphics[width=0.5\textwidth]{figs/vocabproj.pdf}
%   \vspace{-8pt}
%   \caption{Output-based interpretability analysis using VocabProj method.}
%   \vspace{-15pt}
%   \label{fig:vocabproj}
% \end{figure*}

For output-based analysis, we used VocabProj \citep{shu2025surveysparseautoencodersinterpreting} to analyze feature influence patterns on model vocabulary prediction. By computing inner products between each feature vector $w_m$ in the SAE decoder and the language model's output vocabulary embedding matrix $f_{out}(w)$, we obtained logits quantify which words model tends to generate or suppress when features activate:
% \begin{equation}
% \mathcal{I}_m=\arg \max _{w \in \mathcal{V}} f_{\text {out }}(w) \cdot \mathbf{w}_m^{\top}
% \end{equation}
\begingroup
\setlength{\abovedisplayskip}{2pt}
\setlength{\belowdisplayskip}{2pt}
\begin{equation}
\mathcal{I}_m=\arg \max _{w \in \mathcal{V}} f_{\text {out }}(w) \cdot \mathbf{w}_m^{\top}
\end{equation}
\endgroup
We extracted top 10 positive and negative vocabulary items per feature across Llama-3.1-8B and Gemma-2-9B (Fig. \ref{fig:vocabproj}). In Gemma, Feature 5637 specializes in ``title'' concepts, with positive vocabulary centered on variants including different capitalizations and languages (\emph{e.g.}, German ``Titel''), while negative vocabulary contains unrelated nouns. Feature 2370 focuses on ``abstract'' concepts, demonstrating cross-linguistic unity with variants like Spanish ``abstracto'', and negative vocabulary containing technical symbols. In Llama, Feature 2865 concentrates on ``microscopic'' concepts, with positive vocabulary extending from ``micro'' to related terms like ``microscopic'' and ``minuscule,'' showing semantic expansion from root to concept.
These results reveal AdaptiveK learns semantically coherent features capturing not only literal expressions but also concept variants across contexts, languages, and grammatical forms.
%The negative vocabulary ranging from foreign terms to technical terminology consistently forms clear semantic contrasts with target concepts, indicating robust concept identification.

\section{Conclusion}
In this paper, we introduce AdaptiveK SAE, demonstrating that adaptive sparsity based on input complexity improves representation decomposition in LLMs. By establishing that text complexity is linearly encoded in LLM activations, we developed a framework allocating computational resources proportionally to content complexity, eliminating hyperparameter tuning while outperforming fixed-sparsity baselines. Experiments on eight LLMs confirm that complexity-driven adaptation achieves better performance compared to baseline SAEs.

\newpage
\section*{Limitations}
Due to the cost constraints associated with API annotation, our study utilized a significantly smaller training dataset compared to the 500,000,000 tokens employed in SAEBench. Specifically, we trained on 250,000 contexts, extracting the activation value of the final token from each context as its representational vector. Despite this substantial reduction in training data volume, our AdaptiveK SAE achieved impressive performance across most evaluation metrics, with some results even surpassing the baseline SAEs reported in SAEBench. This remarkable efficiency demonstrates the considerable potential of our approach. Rather than relying exclusively on extensive training data, we have introduced a novel SAE training algorithm that fundamentally rethinks sparsity allocation.

% \section*{Acknowledgments}

% This document has been adapted
% by Steven Bethard, Ryan Cotterell and Rui Yan
% from the instructions for earlier ACL and NAACL proceedings, including those for
% ACL 2019 by Douwe Kiela and Ivan Vuli\'{c},
% NAACL 2019 by Stephanie Lukin and Alla Roskovskaya,
% ACL 2018 by Shay Cohen, Kevin Gimpel, and Wei Lu,
% NAACL 2018 by Margaret Mitchell and Stephanie Lukin,
% Bib\TeX{} suggestions for (NA)ACL 2017/2018 from Jason Eisner,
% ACL 2017 by Dan Gildea and Min-Yen Kan,
% NAACL 2017 by Margaret Mitchell,
% ACL 2012 by Maggie Li and Michael White,
% ACL 2010 by Jing-Shin Chang and Philipp Koehn,
% ACL 2008 by Johanna D. Moore, Simone Teufel, James Allan, and Sadaoki Furui,
% ACL 2005 by Hwee Tou Ng and Kemal Oflazer,
% ACL 2002 by Eugene Charniak and Dekang Lin,
% and earlier ACL and EACL formats written by several people, including
% John Chen, Henry S. Thompson and Donald Walker.
% Additional elements were taken from the formatting instructions of the \emph{International Joint Conference on Artificial Intelligence} and the \emph{Conference on Computer Vision and Pattern Recognition}.

% Bibliography entries for the entire Anthology, followed by custom entries
% \bibliography{anthology,custom}
% Custom bibliography entries only
\bibliography{custom}

@inproceedings{mikolov2013linguistic,
  title={Linguistic regularities in continuous space word representations},
  author={Mikolov, Tom{\'a}{\v{s}} and Yih, Wen-tau and Zweig, Geoffrey},
  booktitle={Proceedings of the 2013 conference of the north american chapter of the association for computational linguistics: Human language technologies},
  pages={746--751},
  year={2013}
}

@misc{shu2025surveysparseautoencodersinterpreting,
      title={A Survey on Sparse Autoencoders: Interpreting the Internal Mechanisms of Large Language Models}, 
      author={Dong Shu and Xuansheng Wu and Haiyan Zhao and Daking Rai and Ziyu Yao and Ninghao Liu and Mengnan Du},
      year={2025},
      eprint={2503.05613},
      archivePrefix={arXiv},
      primaryClass={cs.LG},
      url={https://arxiv.org/abs/2503.05613}, 
}

@online{GPT41Mini,
  title = {{GPT-4.1 Mini - OpenAI API}},
  author = {{OpenAI}},
  year = {2024},
  url = {https://platform.openai.com/docs/models/gpt-4.1-mini},
  organization = {OpenAI}
}

@article{liu2311context,
  title={In-context vectors: Making in context learning more effective and controllable through latent space steering},
  author={Liu, Sheng and Ye, Haotian and Xing, Lei and Zou, James},
  journal={URL https://arxiv. org/abs/2311.06668},
  year={2024}
}

@article{marks2023geometry,
  title={The geometry of truth: Emergent linear structure in large language model representations of true/false datasets},
  author={Marks, Samuel and Tegmark, Max},
  journal={arXiv preprint arXiv:2310.06824},
  year={2023}
}

@article{kim2025linear,
  title={Linear Representations of Political Perspective Emerge in Large Language Models},
  author={Kim, Junsol and Evans, James and Schein, Aaron},
  journal={arXiv preprint arXiv:2503.02080},
  year={2025}
}

@article{tigges2023linear,
  title={Linear representations of sentiment in large language models},
  author={Tigges, Curt and Hollinsworth, Oskar John and Geiger, Atticus and Nanda, Neel},
  journal={arXiv preprint arXiv:2310.15154},
  year={2023}
}

@article{gurnee2023language,
  title={Language models represent space and time},
  author={Gurnee, Wes and Tegmark, Max},
  journal={arXiv preprint arXiv:2310.02207},
  year={2023}
}

@article{li2023inference,
  title={Inference-time intervention: Eliciting truthful answers from a language model},
  author={Li, Kenneth and Patel, Oam and Vi{\'e}gas, Fernanda and Pfister, Hanspeter and Wattenberg, Martin},
  journal={Advances in Neural Information Processing Systems},
  volume={36},
  pages={41451--41530},
  year={2023}
}

@article{von2024language,
  title={A language model's guide through latent space},
  author={Von R{\"u}tte, Dimitri and Anagnostidis, Sotiris and Bachmann, Gregor and Hofmann, Thomas},
  journal={arXiv preprint arXiv:2402.14433},
  year={2024}
}

@article{turner2023steering,
  title={Steering language models with activation engineering},
  author={Turner, Alexander Matt and Thiergart, Lisa and Leech, Gavin and Udell, David and Vazquez, Juan J and Mini, Ulisse and MacDiarmid, Monte},
  journal={arXiv preprint arXiv:2308.10248},
  year={2023}
}

@article{belinkov2022probing,
  title={Probing classifiers: Promises, shortcomings, and advances},
  author={Belinkov, Yonatan},
  journal={Computational Linguistics},
  volume={48},
  number={1},
  pages={207--219},
  year={2022},
  publisher={MIT Press One Broadway, 12th Floor, Cambridge, Massachusetts 02142, USA~…}
}

@article{nanda2023emergent,
  title={Emergent linear representations in world models of self-supervised sequence models},
  author={Nanda, Neel and Lee, Andrew and Wattenberg, Martin},
  journal={arXiv preprint arXiv:2309.00941},
  year={2023}
}

@article{arditi2024refusal,
  title={Refusal in language models is mediated by a single direction},
  author={Arditi, Andy and Obeso, Oscar and Syed, Aaquib and Paleka, Daniel and Panickssery, Nina and Gurnee, Wes and Nanda, Neel},
  journal={arXiv preprint arXiv:2406.11717},
  year={2024}
}

@article{gao2020pile,
  title={The pile: An 800gb dataset of diverse text for language modeling},
  author={Gao, Leo and Biderman, Stella and Black, Sid and Golding, Laurence and Hoppe, Travis and Foster, Charles and Phang, Jason and He, Horace and Thite, Anish and Nabeshima, Noa and others},
  journal={arXiv preprint arXiv:2101.00027},
  year={2020}
}

@inproceedings{biderman2023pythia,
  title={Pythia: A suite for analyzing large language models across training and scaling},
  author={Biderman, Stella and Schoelkopf, Hailey and Anthony, Quentin Gregory and Bradley, Herbie and O’Brien, Kyle and Hallahan, Eric and Khan, Mohammad Aflah and Purohit, Shivanshu and Prashanth, USVSN Sai and Raff, Edward and others},
  booktitle={International Conference on Machine Learning},
  pages={2397--2430},
  year={2023},
  organization={PMLR}
}

@article{elhage2022toy,
  title={Toy models of superposition},
  author={Elhage, Nelson and Hume, Tristan and Olsson, Catherine and Schiefer, Nicholas and Henighan, Tom and Kravec, Shauna and Hatfield-Dodds, Zac and Lasenby, Robert and Drain, Dawn and Chen, Carol and others},
  journal={arXiv preprint arXiv:2209.10652},
  year={2022}
}

@article{olah2020zoom,
  title={Zoom in: An introduction to circuits},
  author={Olah, Chris and Cammarata, Nick and Schubert, Ludwig and Goh, Gabriel and Petrov, Michael and Carter, Shan},
  journal={Distill},
  volume={5},
  number={3},
  pages={e00024--001},
  year={2020}
}

@article{park2023linear,
  title={The linear representation hypothesis and the geometry of large language models},
  author={Park, Kiho and Choe, Yo Joong and Veitch, Victor},
  journal={arXiv preprint arXiv:2311.03658},
  year={2023}
}

@article{ferrando2024primer,
  title={A primer on the inner workings of transformer-based language models},
  author={Ferrando, Javier and Sarti, Gabriele and Bisazza, Arianna and Costa-Juss{\`a}, Marta R},
  journal={arXiv preprint arXiv:2405.00208},
  year={2024}
}

@inproceedings{huben2023sparse,
  title={Sparse autoencoders find highly interpretable features in language models},
  author={Huben, Robert and Cunningham, Hoagy and Smith, Logan Riggs and Ewart, Aidan and Sharkey, Lee},
  booktitle={The Twelfth International Conference on Learning Representations},
  year={2023}
}

@article{cunningham2023sparse,
  title={Sparse autoencoders find highly interpretable features in language models},
  author={Cunningham, Hoagy and Ewart, Aidan and Riggs, Logan and Huben, Robert and Sharkey, Lee},
  journal={arXiv preprint arXiv:2309.08600},
  year={2023}
}

@article{rajamanoharan2024improving,
  title={Improving dictionary learning with gated sparse autoencoders},
  author={Rajamanoharan, Senthooran and Conmy, Arthur and Smith, Lewis and Lieberum, Tom and Varma, Vikrant and Kram{\'a}r, J{\'a}nos and Shah, Rohin and Nanda, Neel},
  journal={arXiv preprint arXiv:2404.16014},
  year={2024}
}

@article{gao2024scaling,
  title={Scaling and evaluating sparse autoencoders},
  author={Gao, Leo and la Tour, Tom Dupr{\'e} and Tillman, Henk and Goh, Gabriel and Troll, Rajan and Radford, Alec and Sutskever, Ilya and Leike, Jan and Wu, Jeffrey},
  journal={arXiv preprint arXiv:2406.04093},
  year={2024}
}

@article{bussmann2024batchtopk,
  title={Batchtopk sparse autoencoders},
  author={Bussmann, Bart and Leask, Patrick and Nanda, Neel},
  journal={arXiv preprint arXiv:2412.06410},
  year={2024}
}

@article{rajamanoharan2024jumping,
  title={Jumping ahead: Improving reconstruction fidelity with jumprelu sparse autoencoders},
  author={Rajamanoharan, Senthooran and Lieberum, Tom and Sonnerat, Nicolas and Conmy, Arthur and Varma, Vikrant and Kram{\'a}r, J{\'a}nos and Nanda, Neel},
  journal={arXiv preprint arXiv:2407.14435},
  year={2024}
}

@misc{ProLUNonlinearity,
   title = {ProLU: A Nonlinearity for Sparse Autoencoders},
   author = {Glen M. Taggart},
   year = {2024},
   howpublished = {\url{https://www.alignmentforum.org/posts/HEpufTdakGTTKgoYF/prolu-a-nonlinearity-for-sparse-autoencoders}},
}

@article{mudide2024efficient,
  title={Efficient dictionary learning with switch sparse autoencoders},
  author={Mudide, Anish and Engels, Joshua and Michaud, Eric J and Tegmark, Max and de Witt, Christian Schroeder},
  journal={arXiv preprint arXiv:2410.08201},
  year={2024}
}

@inproceedings{bussmannlearning,
  title={Learning multi-level features with matryoshka saes, December 19 2024b},
  author={Bussmann, Bart and Leask, Patrick and Nanda, Neel},
  booktitle={URL https://www. alignmentforum. org/posts/rKM9b6B2LqwSB5ToN/learning-multi-level-features-with-matryoshka-saes. Alignment Forum},
  year={2024}
}

@article{karvonen2024measuring,
  title={Measuring progress in dictionary learning for language model interpretability with board game models},
  author={Karvonen, Adam and Wright, Benjamin and Rager, Can and Angell, Rico and Brinkmann, Jannik and Smith, Logan and Mayrink Verdun, Claudio and Bau, David and Marks, Samuel},
  journal={Advances in Neural Information Processing Systems},
  volume={37},
  pages={83091--83118},
  year={2024}
}

@article{marks2024enhancing,
  title={Enhancing Neural Network Interpretability with Feature-Aligned Sparse Autoencoders},
  author={Marks, Luke and Paren, Alasdair and Krueger, David and Barez, Fazl},
  journal={arXiv preprint arXiv:2411.01220},
  year={2024}
}

@article{braun2024identifying,
  title={Identifying functionally important features with end-to-end sparse dictionary learning},
  author={Braun, Dan and Taylor, Jordan and Goldowsky-Dill, Nicholas and Sharkey, Lee},
  journal={Advances in Neural Information Processing Systems},
  volume={37},
  pages={107286--107325},
  year={2024}
}

@misc{marks2024dictionarylearning,
    title = {\texttt{dictionary\_learning}},
    author = {Samuel Marks and Adam Karvonen and Aaron Mueller},    
    year = {2024},
    howpublished = {\url{https://github.com/saprmarks/dictionary_learning}},
}

@article{bricken2023towards,
  title={Towards monosemanticity: Decomposing language models with dictionary learning},
  author={Bricken, Trenton and Templeton, Adly and Batson, Joshua and Chen, Brian and Jermyn, Adam and Conerly, Tom and Turner, Nick and Anil, Cem and Denison, Carson and Askell, Amanda and others},
  journal={Transformer Circuits Thread},
  volume={2},
  year={2023}
}

@misc{anthropic2024circuits,
  title        = {Circuits Updates---April 2024},
  author       = {{Anthropic Interpretability Team}},
  howpublished = {\url{https://transformer-circuits.pub/2024/april-update/index.html}},
  month        = apr,
  year         = 2024,
  urldate      = {2025-05-06}
}

@article{kingma2014adam,
  title={Adam: A method for stochastic optimization},
  author={Kingma, Diederik P and Ba, Jimmy},
  journal={arXiv preprint arXiv:1412.6980},
  year={2014}
}

@article{team2024gemma,
  title={Gemma 2: Improving open language models at a practical size},
  author={Team, Gemma and Riviere, Morgane and Pathak, Shreya and Sessa, Pier Giuseppe and Hardin, Cassidy and Bhupatiraju, Surya and Hussenot, L{\'e}onard and Mesnard, Thomas and Shahriari, Bobak and Ram{\'e}, Alexandre and others},
  journal={arXiv preprint arXiv:2408.00118},
  year={2024}
}

@article{arora2018linear,
  title={Linear algebraic structure of word senses, with applications to polysemy},
  author={Arora, Sanjeev and Li, Yuanzhi and Liang, Yingyu and Ma, Tengyu and Risteski, Andrej},
  journal={Transactions of the Association for Computational Linguistics},
  volume={6},
  pages={483--495},
  year={2018},
  publisher={MIT Press One Rogers Street, Cambridge, MA 02142-1209, USA journals-info~…}
}

@article{gurnee2023finding,
  title={Finding neurons in a haystack: Case studies with sparse probing},
  author={Gurnee, Wes and Nanda, Neel and Pauly, Matthew and Harvey, Katherine and Troitskii, Dmitrii and Bertsimas, Dimitris},
  journal={arXiv preprint arXiv:2305.01610},
  year={2023}
}

@article{olah2023distributed,
  title={Distributed representations: Composition \& superposition},
  author={Olah, Chris},
  journal={Transformer Circuits Thread},
  volume={27},
  year={2023}
}

@misc{templeton2024scaling,
    title        = {Scaling Monosemanticity: Extracting Interpretable Features from Claude 3 Sonnet},
    author       = {Adly Templeton and Tom Conerly and Joshua Marcus and Jack Lindsey
    and Trenton Bricken and Brian Chen and Adam Jermyn and Nicholas L. Turner
    and Cem Anil and Carson Denison and Amanda Askell and Robert Lasenby
    and Yifan Wu and Shauna Kravec and Nicholas Schiefer and Tim Maxwell
    and Nicholas Joseph and Alex Tamkin and Karina Nguyen and Brayden McLean
    and Josiah E. Burke and Tristan Hume and Shan Carter and Tom Henighan
    and Chris Olah},
    howpublished = {\url{https://transformer-circuits.pub/2024/scaling-monosemanticity/index.html}},
    month        = apr,
    year         = 2024,
    note         = {Transformer Circuits Thread. Accessed 2025-05-06}
}

@article{karvonen2025saebench,
  title={SAEBench: A Comprehensive Benchmark for Sparse Autoencoders in Language Model Interpretability},
  author={Karvonen, Adam and Rager, Can and Lin, Johnny and Tigges, Curt and Bloom, Joseph and Chanin, David and Lau, Yeu-Tong and Farrell, Eoin and McDougall, Callum and Ayonrinde, Kola and others},
  journal={arXiv preprint arXiv:2503.09532},
  year={2025}
}

@article{peters2018dissecting,
  title={Dissecting contextual word embeddings: Architecture and representation},
  author={Peters, Matthew E and Neumann, Mark and Zettlemoyer, Luke and Yih, Wen-tau},
  journal={arXiv preprint arXiv:1808.08949},
  year={2018}
}

@misc{grattafiori2024llama3herdmodels,
      title={The Llama 3 Herd of Models}, 
      author={Aaron Grattafiori and Abhimanyu Dubey and Abhinav Jauhri and Abhinav Pandey and Abhishek Kadian and Ahmad Al-Dahle and Aiesha Letman and Akhil Mathur and Alan Schelten and Alex Vaughan and Amy Yang and Angela Fan and Anirudh Goyal and Anthony Hartshorn and Aobo Yang and Archi Mitra and Archie Sravankumar and Artem Korenev and Arthur Hinsvark and Arun Rao and Aston Zhang and Aurelien Rodriguez and Austen Gregerson and Ava Spataru and Baptiste Roziere and Bethany Biron and Binh Tang and Bobbie Chern and Charlotte Caucheteux and Chaya Nayak and Chloe Bi and Chris Marra and Chris McConnell and Christian Keller and Christophe Touret and Chunyang Wu and Corinne Wong and Cristian Canton Ferrer and Cyrus Nikolaidis and Damien Allonsius and Daniel Song and Danielle Pintz and Danny Livshits and Danny Wyatt and David Esiobu and Dhruv Choudhary and Dhruv Mahajan and Diego Garcia-Olano and Diego Perino and Dieuwke Hupkes and Egor Lakomkin and Ehab AlBadawy and Elina Lobanova and Emily Dinan and Eric Michael Smith and Filip Radenovic and Francisco Guzmán and Frank Zhang and Gabriel Synnaeve and Gabrielle Lee and Georgia Lewis Anderson and Govind Thattai and Graeme Nail and Gregoire Mialon and Guan Pang and Guillem Cucurell and Hailey Nguyen and Hannah Korevaar and Hu Xu and Hugo Touvron and Iliyan Zarov and Imanol Arrieta Ibarra and Isabel Kloumann and Ishan Misra and Ivan Evtimov and Jack Zhang and Jade Copet and Jaewon Lee and Jan Geffert and Jana Vranes and Jason Park and Jay Mahadeokar and Jeet Shah and Jelmer van der Linde and Jennifer Billock and Jenny Hong and Jenya Lee and Jeremy Fu and Jianfeng Chi and Jianyu Huang and Jiawen Liu and Jie Wang and Jiecao Yu and Joanna Bitton and Joe Spisak and Jongsoo Park and Joseph Rocca and Joshua Johnstun and Joshua Saxe and Junteng Jia and Kalyan Vasuden Alwala and Karthik Prasad and Kartikeya Upasani and Kate Plawiak and Ke Li and Kenneth Heafield and Kevin Stone and Khalid El-Arini and Krithika Iyer and Kshitiz Malik and Kuenley Chiu and Kunal Bhalla and Kushal Lakhotia and Lauren Rantala-Yeary and Laurens van der Maaten and Lawrence Chen and Liang Tan and Liz Jenkins and Louis Martin and Lovish Madaan and Lubo Malo and Lukas Blecher and Lukas Landzaat and Luke de Oliveira and Madeline Muzzi and Mahesh Pasupuleti and Mannat Singh and Manohar Paluri and Marcin Kardas and Maria Tsimpoukelli and Mathew Oldham and Mathieu Rita and Maya Pavlova and Melanie Kambadur and Mike Lewis and Min Si and Mitesh Kumar Singh and Mona Hassan and Naman Goyal and Narjes Torabi and Nikolay Bashlykov and Nikolay Bogoychev and Niladri Chatterji and Ning Zhang and Olivier Duchenne and Onur Çelebi and Patrick Alrassy and Pengchuan Zhang and Pengwei Li and Petar Vasic and Peter Weng and Prajjwal Bhargava and Pratik Dubal and Praveen Krishnan and Punit Singh Koura and Puxin Xu and Qing He and Qingxiao Dong and Ragavan Srinivasan and Raj Ganapathy and Ramon Calderer and Ricardo Silveira Cabral and Robert Stojnic and Roberta Raileanu and Rohan Maheswari and Rohit Girdhar and Rohit Patel and Romain Sauvestre and Ronnie Polidoro and Roshan Sumbaly and Ross Taylor and Ruan Silva and Rui Hou and Rui Wang and Saghar Hosseini and Sahana Chennabasappa and Sanjay Singh and Sean Bell and Seohyun Sonia Kim and Sergey Edunov and Shaoliang Nie and Sharan Narang and Sharath Raparthy and Sheng Shen and Shengye Wan and Shruti Bhosale and Shun Zhang and Simon Vandenhende and Soumya Batra and Spencer Whitman and Sten Sootla and Stephane Collot and Suchin Gururangan and Sydney Borodinsky and Tamar Herman and Tara Fowler and Tarek Sheasha and Thomas Georgiou and Thomas Scialom and Tobias Speckbacher and Todor Mihaylov and Tong Xiao and Ujjwal Karn and Vedanuj Goswami and Vibhor Gupta and Vignesh Ramanathan and Viktor Kerkez and Vincent Gonguet and Virginie Do and Vish Vogeti and Vítor Albiero and Vladan Petrovic and Weiwei Chu and Wenhan Xiong and Wenyin Fu and Whitney Meers and Xavier Martinet and Xiaodong Wang and Xiaofang Wang and Xiaoqing Ellen Tan and Xide Xia and Xinfeng Xie and Xuchao Jia and Xuewei Wang and Yaelle Goldschlag and Yashesh Gaur and Yasmine Babaei and Yi Wen and Yiwen Song and Yuchen Zhang and Yue Li and Yuning Mao and Zacharie Delpierre Coudert and Zheng Yan and Zhengxing Chen and Zoe Papakipos and Aaditya Singh and Aayushi Srivastava and Abha Jain and Adam Kelsey and Adam Shajnfeld and Adithya Gangidi and Adolfo Victoria and Ahuva Goldstand and Ajay Menon and Ajay Sharma and Alex Boesenberg and Alexei Baevski and Allie Feinstein and Amanda Kallet and Amit Sangani and Amos Teo and Anam Yunus and Andrei Lupu and Andres Alvarado and Andrew Caples and Andrew Gu and Andrew Ho and Andrew Poulton and Andrew Ryan and Ankit Ramchandani and Annie Dong and Annie Franco and Anuj Goyal and Aparajita Saraf and Arkabandhu Chowdhury and Ashley Gabriel and Ashwin Bharambe and Assaf Eisenman and Azadeh Yazdan and Beau James and Ben Maurer and Benjamin Leonhardi and Bernie Huang and Beth Loyd and Beto De Paola and Bhargavi Paranjape and Bing Liu and Bo Wu and Boyu Ni and Braden Hancock and Bram Wasti and Brandon Spence and Brani Stojkovic and Brian Gamido and Britt Montalvo and Carl Parker and Carly Burton and Catalina Mejia and Ce Liu and Changhan Wang and Changkyu Kim and Chao Zhou and Chester Hu and Ching-Hsiang Chu and Chris Cai and Chris Tindal and Christoph Feichtenhofer and Cynthia Gao and Damon Civin and Dana Beaty and Daniel Kreymer and Daniel Li and David Adkins and David Xu and Davide Testuggine and Delia David and Devi Parikh and Diana Liskovich and Didem Foss and Dingkang Wang and Duc Le and Dustin Holland and Edward Dowling and Eissa Jamil and Elaine Montgomery and Eleonora Presani and Emily Hahn and Emily Wood and Eric-Tuan Le and Erik Brinkman and Esteban Arcaute and Evan Dunbar and Evan Smothers and Fei Sun and Felix Kreuk and Feng Tian and Filippos Kokkinos and Firat Ozgenel and Francesco Caggioni and Frank Kanayet and Frank Seide and Gabriela Medina Florez and Gabriella Schwarz and Gada Badeer and Georgia Swee and Gil Halpern and Grant Herman and Grigory Sizov and Guangyi and Zhang and Guna Lakshminarayanan and Hakan Inan and Hamid Shojanazeri and Han Zou and Hannah Wang and Hanwen Zha and Haroun Habeeb and Harrison Rudolph and Helen Suk and Henry Aspegren and Hunter Goldman and Hongyuan Zhan and Ibrahim Damlaj and Igor Molybog and Igor Tufanov and Ilias Leontiadis and Irina-Elena Veliche and Itai Gat and Jake Weissman and James Geboski and James Kohli and Janice Lam and Japhet Asher and Jean-Baptiste Gaya and Jeff Marcus and Jeff Tang and Jennifer Chan and Jenny Zhen and Jeremy Reizenstein and Jeremy Teboul and Jessica Zhong and Jian Jin and Jingyi Yang and Joe Cummings and Jon Carvill and Jon Shepard and Jonathan McPhie and Jonathan Torres and Josh Ginsburg and Junjie Wang and Kai Wu and Kam Hou U and Karan Saxena and Kartikay Khandelwal and Katayoun Zand and Kathy Matosich and Kaushik Veeraraghavan and Kelly Michelena and Keqian Li and Kiran Jagadeesh and Kun Huang and Kunal Chawla and Kyle Huang and Lailin Chen and Lakshya Garg and Lavender A and Leandro Silva and Lee Bell and Lei Zhang and Liangpeng Guo and Licheng Yu and Liron Moshkovich and Luca Wehrstedt and Madian Khabsa and Manav Avalani and Manish Bhatt and Martynas Mankus and Matan Hasson and Matthew Lennie and Matthias Reso and Maxim Groshev and Maxim Naumov and Maya Lathi and Meghan Keneally and Miao Liu and Michael L. Seltzer and Michal Valko and Michelle Restrepo and Mihir Patel and Mik Vyatskov and Mikayel Samvelyan and Mike Clark and Mike Macey and Mike Wang and Miquel Jubert Hermoso and Mo Metanat and Mohammad Rastegari and Munish Bansal and Nandhini Santhanam and Natascha Parks and Natasha White and Navyata Bawa and Nayan Singhal and Nick Egebo and Nicolas Usunier and Nikhil Mehta and Nikolay Pavlovich Laptev and Ning Dong and Norman Cheng and Oleg Chernoguz and Olivia Hart and Omkar Salpekar and Ozlem Kalinli and Parkin Kent and Parth Parekh and Paul Saab and Pavan Balaji and Pedro Rittner and Philip Bontrager and Pierre Roux and Piotr Dollar and Polina Zvyagina and Prashant Ratanchandani and Pritish Yuvraj and Qian Liang and Rachad Alao and Rachel Rodriguez and Rafi Ayub and Raghotham Murthy and Raghu Nayani and Rahul Mitra and Rangaprabhu Parthasarathy and Raymond Li and Rebekkah Hogan and Robin Battey and Rocky Wang and Russ Howes and Ruty Rinott and Sachin Mehta and Sachin Siby and Sai Jayesh Bondu and Samyak Datta and Sara Chugh and Sara Hunt and Sargun Dhillon and Sasha Sidorov and Satadru Pan and Saurabh Mahajan and Saurabh Verma and Seiji Yamamoto and Sharadh Ramaswamy and Shaun Lindsay and Shaun Lindsay and Sheng Feng and Shenghao Lin and Shengxin Cindy Zha and Shishir Patil and Shiva Shankar and Shuqiang Zhang and Shuqiang Zhang and Sinong Wang and Sneha Agarwal and Soji Sajuyigbe and Soumith Chintala and Stephanie Max and Stephen Chen and Steve Kehoe and Steve Satterfield and Sudarshan Govindaprasad and Sumit Gupta and Summer Deng and Sungmin Cho and Sunny Virk and Suraj Subramanian and Sy Choudhury and Sydney Goldman and Tal Remez and Tamar Glaser and Tamara Best and Thilo Koehler and Thomas Robinson and Tianhe Li and Tianjun Zhang and Tim Matthews and Timothy Chou and Tzook Shaked and Varun Vontimitta and Victoria Ajayi and Victoria Montanez and Vijai Mohan and Vinay Satish Kumar and Vishal Mangla and Vlad Ionescu and Vlad Poenaru and Vlad Tiberiu Mihailescu and Vladimir Ivanov and Wei Li and Wenchen Wang and Wenwen Jiang and Wes Bouaziz and Will Constable and Xiaocheng Tang and Xiaojian Wu and Xiaolan Wang and Xilun Wu and Xinbo Gao and Yaniv Kleinman and Yanjun Chen and Ye Hu and Ye Jia and Ye Qi and Yenda Li and Yilin Zhang and Ying Zhang and Yossi Adi and Youngjin Nam and Yu and Wang and Yu Zhao and Yuchen Hao and Yundi Qian and Yunlu Li and Yuzi He and Zach Rait and Zachary DeVito and Zef Rosnbrick and Zhaoduo Wen and Zhenyu Yang and Zhiwei Zhao and Zhiyu Ma},
      year={2024},
      eprint={2407.21783},
      archivePrefix={arXiv},
      primaryClass={cs.AI},
      url={https://arxiv.org/abs/2407.21783}, 
}

@misc{yang2025qwen3technicalreport,
      title={Qwen3 Technical Report}, 
      author={An Yang and Anfeng Li and Baosong Yang and Beichen Zhang and Binyuan Hui and Bo Zheng and Bowen Yu and Chang Gao and Chengen Huang and Chenxu Lv and Chujie Zheng and Dayiheng Liu and Fan Zhou and Fei Huang and Feng Hu and Hao Ge and Haoran Wei and Huan Lin and Jialong Tang and Jian Yang and Jianhong Tu and Jianwei Zhang and Jianxin Yang and Jiaxi Yang and Jing Zhou and Jingren Zhou and Junyang Lin and Kai Dang and Keqin Bao and Kexin Yang and Le Yu and Lianghao Deng and Mei Li and Mingfeng Xue and Mingze Li and Pei Zhang and Peng Wang and Qin Zhu and Rui Men and Ruize Gao and Shixuan Liu and Shuang Luo and Tianhao Li and Tianyi Tang and Wenbiao Yin and Xingzhang Ren and Xinyu Wang and Xinyu Zhang and Xuancheng Ren and Yang Fan and Yang Su and Yichang Zhang and Yinger Zhang and Yu Wan and Yuqiong Liu and Zekun Wang and Zeyu Cui and Zhenru Zhang and Zhipeng Zhou and Zihan Qiu},
      year={2025},
      eprint={2505.09388},
      archivePrefix={arXiv},
      primaryClass={cs.CL},
      url={https://arxiv.org/abs/2505.09388}, 
}

@misc{abdin2024phi4technicalreport,
      title={Phi-4 Technical Report}, 
      author={Marah Abdin and Jyoti Aneja and Harkirat Behl and Sébastien Bubeck and Ronen Eldan and Suriya Gunasekar and Michael Harrison and Russell J. Hewett and Mojan Javaheripi and Piero Kauffmann and James R. Lee and Yin Tat Lee and Yuanzhi Li and Weishung Liu and Caio C. T. Mendes and Anh Nguyen and Eric Price and Gustavo de Rosa and Olli Saarikivi and Adil Salim and Shital Shah and Xin Wang and Rachel Ward and Yue Wu and Dingli Yu and Cyril Zhang and Yi Zhang},
      year={2024},
      eprint={2412.08905},
      archivePrefix={arXiv},
      primaryClass={cs.CL},
      url={https://arxiv.org/abs/2412.08905}, 
}

@misc{rai2025practicalreviewmechanisticinterpretability,
      title={A Practical Review of Mechanistic Interpretability for Transformer-Based Language Models}, 
      author={Daking Rai and Yilun Zhou and Shi Feng and Abulhair Saparov and Ziyu Yao},
      year={2025},
      eprint={2407.02646},
      archivePrefix={arXiv},
      primaryClass={cs.AI},
      url={https://arxiv.org/abs/2407.02646}, 
}

@misc{zhao2023explainabilitylargelanguagemodels,
      title={Explainability for Large Language Models: A Survey}, 
      author={Haiyan Zhao and Hanjie Chen and Fan Yang and Ninghao Liu and Huiqi Deng and Hengyi Cai and Shuaiqiang Wang and Dawei Yin and Mengnan Du},
      year={2023},
      eprint={2309.01029},
      archivePrefix={arXiv},
      primaryClass={cs.CL},
      url={https://arxiv.org/abs/2309.01029}, 
}

@misc{lee2025evaluatingdesigningsparseautoencoders,
      title={Evaluating and Designing Sparse Autoencoders by Approximating Quasi-Orthogonality}, 
      author={Sewoong Lee and Adam Davies and Marc E. Canby and Julia Hockenmaier},
      year={2025},
      eprint={2503.24277},
      archivePrefix={arXiv},
      primaryClass={cs.LG},
      url={https://arxiv.org/abs/2503.24277}, 
}

\clearpage
\appendix

\section{Linear Probes Training Details}
\label{sec:linear-probes-details}

Our training methodology employs texts from the pile-uncopyrighted corpus, which are processed through a tokenizer and aggregated into contexts of 1024 tokens each. Each context undergoes a comprehensive six-dimensional evaluation by GPT-4.1-mini, resulting in a normalized complexity score between 0 and 10 with one decimal place precision. Our dataset comprises 250,000 training contexts and 10,000 test contexts.

For each context, we extract the activation vector of the final token as the representational vector for that context, with dimensionality matching that of the model's hidden layer. During training, batches of unprocessed activation values are sequentially retrieved from the buffer and marked as processed. When all activations have been utilized, the buffer is replenished and shuffled to introduce stochasticity. This iterative cycle continues until a sufficient quantity of activation samples has been accumulated for effective model training.

Here are some examples with various complexity scores, their predicted complexities by the linear probe, corresponding K-values, and the number of activated latent features through AdaptiveK SAE in Fig. \ref{fig:samples}.

\begin{figure*}[htbp]
\begin{scriptsize}
\begin{tcolorbox}[title={Sample 1: True Complexity: 1.2, Pred Complexity: 3.18, K value: 95, Activated Features: 95}, 
fonttitle=\small\bfseries]
\begin{lstlisting}[basicstyle=\ttfamily\scriptsize]
\u00f4m\u00e9\",\n        \"Employ\u00e9 de commerce\",\n        \"Employ\u00e9 de commerce CFC\",\n        \"Employ\u00e9 de remont\u00e9es m\u00e9caniques AFP\",\n        \"Employ\u00e9 d\u2019exploitation AFP\",\n        \"Employ\u00e9 en cuisine AFP\",\n        \"Employ\u00e9 en h\u00f4tellerie AFP\",\n        \"Employ\u00e9 en industrie laiti\u00e8re AFP\",\n        \"Employ\u00e9 en intendance AFP\",\n        \"Employ\u00e9 en intendance AFP\",\n        \"Employ\u00e9 en restauration AFP\",\n        \"Enqu\u00eateur de douane avec dipl\u00f4me f\u00e9d\u00e9ral\",\n        \"Entra\u00eeneur de sport de performance avec brevet f\u00e9d\u00e9ra...
\end{lstlisting}
\end{tcolorbox}
\end{scriptsize}
% \end{figure*}

% \begin{figure*}[t]
\begin{scriptsize}
\begin{tcolorbox}[title={Sample 2: True Complexity: 3.5, Pred Complexity: 4.26, K value: 137, Activated Features: 137}, 
fonttitle=\small\bfseries]
\begin{lstlisting}[basicstyle=\ttfamily\scriptsize]
max-age=31536000; includeSubDomains]\n      x-aspnet-version: [4.0.30319]\n      x-content-type-options: [nosniff]\n      x-ms-ratelimit-remaining-subscription-resource-requests: ['11998']\n      x-powered-by: [ASP.NET]\n    status: {code: 201, message: Created}\n- request:\n    body: null\n    headers:\n      Accept: [application/json]\n      Accept-Encoding: ['gzip, deflate']\n      CommandName: [network dns zone import]\n      Connection: [keep-alive]\n      Content-Type: [application/json; charset=utf-8]...
\end{lstlisting}
\end{tcolorbox}
\end{scriptsize}
% \end{figure*}

% \begin{figure*}[t]
\begin{scriptsize}
\begin{tcolorbox}[title={Sample 3: True Complexity: 5.2, Pred Complexity: 5.39, K value: 187, Activated Features: 187}, 
fonttitle=\small\bfseries]
\begin{lstlisting}[basicstyle=\ttfamily\scriptsize]
 very much interested in the idea of sanctuary. How do the spirits of these two authors and the respective sanctuaries they sought infuse Tom and Nathan\u2019s interactions? What other giants of American literature have an influence, direct or indirect, on the characters in The Brooklyn Follies?\n\n\u201cYou love life,\u201d says Nathan to Tom, \u201cbut you don\u2019t believe in it. And neither do I.\u201d This statement quickly becomes untrue as both men cast off their inertia. To what extent does action create belief for bo...
\end{lstlisting}
\end{tcolorbox}
\end{scriptsize}
% \end{figure*}

% \begin{figure*}[t]
\begin{scriptsize}
\begin{tcolorbox}[title={Sample 4: True Complexity: 6.2, Pred Complexity: 6.05, K value: 216, Activated Features: 216}, 
fonttitle=\small\bfseries]
\begin{lstlisting}[basicstyle=\ttfamily\scriptsize]
 itself implement the\n    // interface because that exposes all the public methods of that interface at the manager level.\n    private static final String INTENT_URL_PREFIX = \"intent:\";\n\n    // The animation duration of a URL being promoted to a tab when triggered by an\n    // intercept navigation. This is faster than the standard tab promotion animation\n    // so that it completes before the navigation.\n    private static final long INTERCEPT_NAVIGATION_PROMOTION_ANIMATION_DURATION_MS = 40;\n\n  ...
\end{lstlisting}
\end{tcolorbox}
\end{scriptsize}
% \end{figure*}

% \begin{figure*}[t]
\begin{scriptsize}
\begin{tcolorbox}[title={Sample 5: True Complexity: 9.1, Pred Complexity: 9.19, K value: 298, Activated Features: 242}, 
fonttitle=\small\bfseries]
\begin{lstlisting}[basicstyle=\ttfamily\scriptsize]
 parameters of the MSSM and to trace back, sector-wise, the sensitivity to initial conditions of the Yukawa couplings and the soft susy breaking parameters. We have established analytically a generic screening of non-universality, in the vicinity of the infrared quasi fixed points. In practice, this property gives the general trend of the behaviour, despite the large number of free parameters, and even when one is not very close to such a quasi fixed point. This shows that non-universality of th...
\end{lstlisting}
\end{tcolorbox}
\end{scriptsize}
% \end{figure*}

% \begin{figure*}[t]
\begin{scriptsize}
\begin{tcolorbox}[title={Sample 6: True Complexity: 9.6, Pred Complexity: 8.91, K value: 294, Activated Features: 294}, 
fonttitle=\small\bfseries]
\begin{lstlisting}[basicstyle=\ttfamily\scriptsize]
 {{\\ensuremath{\\mathbbm{R}}}}^d) \\to [0,\\infty)$ satisfy for all $d\\in {{\\ensuremath{\\mathbbm{N}}}}$, $x=(x_1,\\ldots,x_d)\\in {{\\ensuremath{\\mathbbm{R}}}}^d$ that $\n{\\mathbf{A}_{d}}(x)= \\left(\\max\\{x_1,0\\},\\ldots,\\max\\{x_d,0\\}\\right)\n$ and $\\|x\\|=[\\sum_{i=1}^d(x_i)^2]^{1/2}$, let $\n{\\mathbf{N}}= \\cup_{H\\in  {{\\ensuremath{\\mathbbm{N}}}}}\\cup_{(k_0,k_1,\\ldots,k_{H+1})\\in {{\\ensuremath{\\mathbbm{N}}}}^{H+2}}\n[ \\prod_{n=1}^{H+1} \\left({{\\ensuremath{\\mathbbm{R}}}}^{k_{n}\\times k_{n-1}} \\times{{\\ensurem...
\end{lstlisting}
\end{tcolorbox}
\end{scriptsize}

\caption{Examples showing complexity prediction and feature activation.}
\label{fig:samples}
\end{figure*}

\renewcommand{\arraystretch}{1.3} 
\begin{table*}[h]
\centering
\caption{Sparsity setting of baseline SAEs}
\label{tab:baseline SAE}
\begin{tabular}{cccc}
\toprule
SAE                   & Sparsity & Pythia                    & Gemma/Llama/Qwen/Phi                    \\ \hline
TopK  & \multirow{3}{*}{K Value}             & \multicolumn{2}{c}{20, 40, 80, 160, 320, 640}          \\
Batch TopK            &          & \multicolumn{2}{c}{20, 40, 80, 160, 320, 640}             \\
Matryoshka &          & \multicolumn{2}{c}{20, 40, 80, 160, 320, 640}             \\ \hline
Gated & \multirow{4}{*}{Sparsity Penalities} & 0.6, 0.9, 1.2, 2, 3, 4 & 0.012, 0.015, 0.02, 0.03, 0.04, 0.06 \\
Relu                  &          & 0.6, 0.9, 1.2, 2, 3, 4    & 0.012, 0.015, 0.02, 0.03, 0.04, 0.06 \\
Relu\_new             &          & 0.6, 0.9, 1.2, 2, 3, 4    & 0.012, 0.015, 0.02, 0.03, 0.04, 0.06 \\
P Anneal              &          & 0.3, 0.45, 0.6, 1, 1.5, 2 & 0.006, 0.008, 0.01, 0.015, 0.02, 0.025   \\ \bottomrule
\end{tabular}
\end{table*}

\section{Additional SAE Training Details}
\label{sec:additional-sae-training-appendix}
Our SAEs are trained on the residual stream because researchers typically focus on this component when interpreting or steering model behaviors. This alignment ensures the learned representations directly support the most common SAE applications in model analysis and intervention. 

The training and testing datasets for the AdaptiveK SAE are consistent with those utilized in the linear probe training phase. The overall training process is determined by the current step value, with three distinct phases: step=0 is the pre-training phase, dedicated to probe training; step $<$ total steps$\times$phase ratio involve training the SAE while maintaining frozen probe parameters; step $>$ this threshold initiate the joint fine-tuning phase. The total step count is calculated by dividing the total token count (250,000) by the batch processing capacity (2048 tokens), with phase ratio set at 0.9.

During the third phase, the deviation weight adapts dynamically throughout training. The system maintains records of probe losses from the three most recent steps and calculates the rate of loss change. When rapid loss reduction occurs (change rate exceeding the threshold 0.5), the deviation weight is reduced to 0.8 of its original value; otherwise, it increases to 1.2, with an upper limit of 0.5. A sigmoid function maps predicted complexity scores (0-10) to feature quantity ranges (from min\_k to max\_k, established at 20 to 320), with the sigmoid midpoint corresponding to base\_k (80) and steepness (0.6) controlling the mapping curve's configuration. This enables the AdaptiveK SAE to dynamically allocate sparsity by assigning fewer features to simpler texts (low complexity) while allocating more features to complex texts (high complexity).

As the AdaptiveK SAE was trained on 250,000 token activations, we employed identical training and testing datasets for training and evaluating baseline SAEs, with their sparsity configurations detailed in Tab. \ref{tab:baseline SAE}.

\section{More Results}
\subsection{Linear Probe Performance}
\begin{figure*}[t]
  \centering
  \begin{subfigure}[b]{0.24\linewidth}
    \centering
    \includegraphics[width=\linewidth]{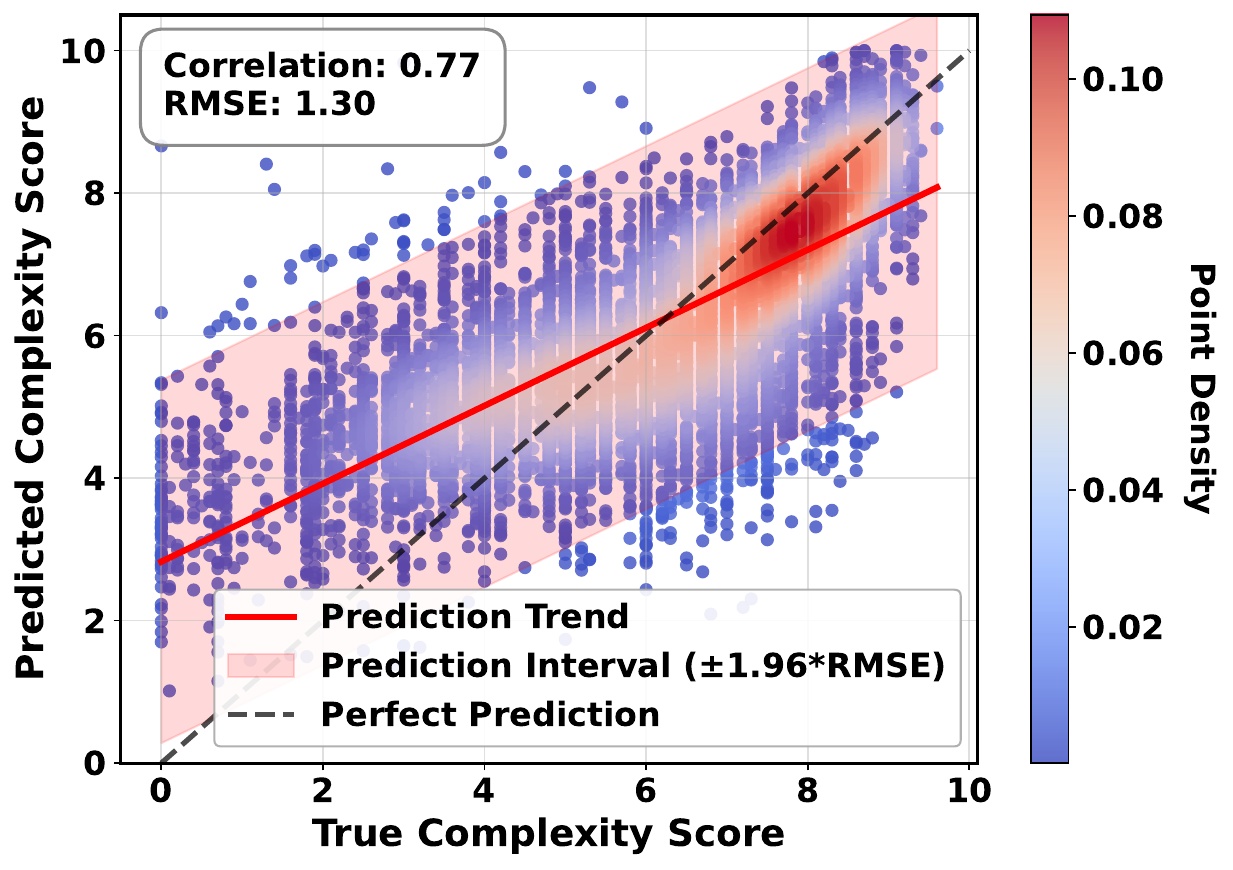}
    \caption{Pythia-160M}
    \label{fig:linear-160}
  \end{subfigure}
  \hfill
  \begin{subfigure}[b]{0.24\linewidth}
    \centering
    \includegraphics[width=\linewidth]{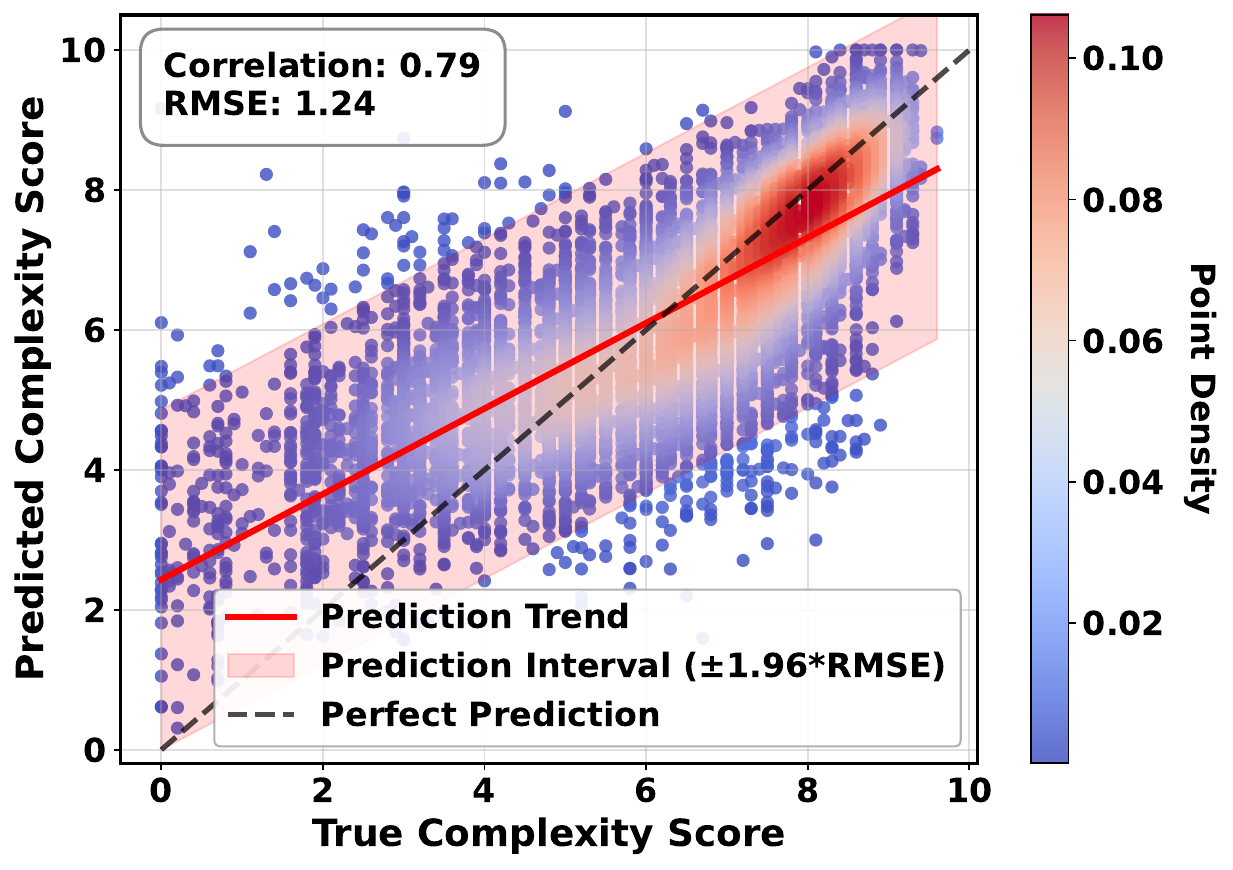}
    \caption{Gemma-2-9B}
    \label{fig:linear-gemma-9b}
  \end{subfigure}
  \hfill
  \begin{subfigure}[b]{0.24\linewidth}
    \centering
    \includegraphics[width=\linewidth]{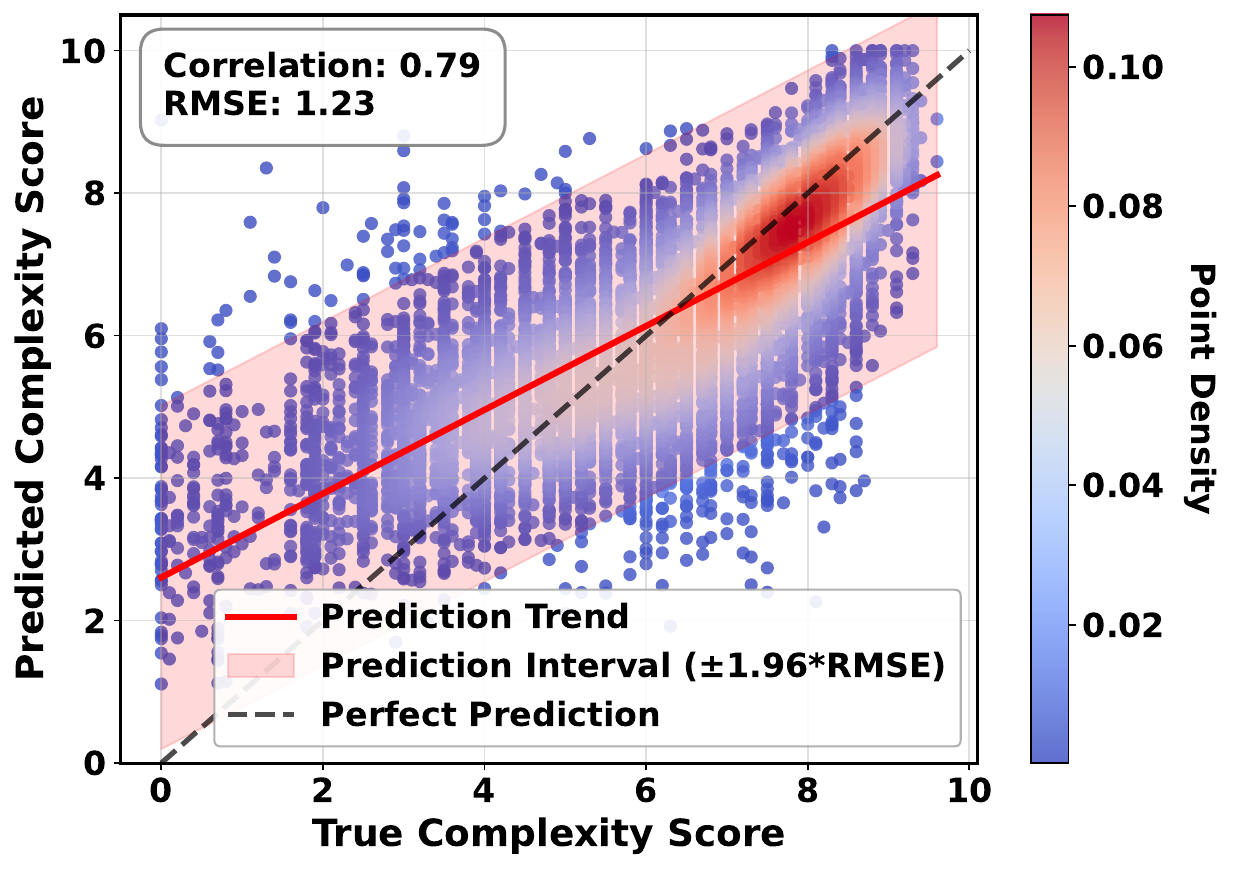}
    \caption{Llama-3.1-8B}
    \label{fig:linear-llama-8b}
  \end{subfigure}
  \hfill
  \begin{subfigure}[b]{0.24\linewidth}
    \centering
    \includegraphics[width=\linewidth]{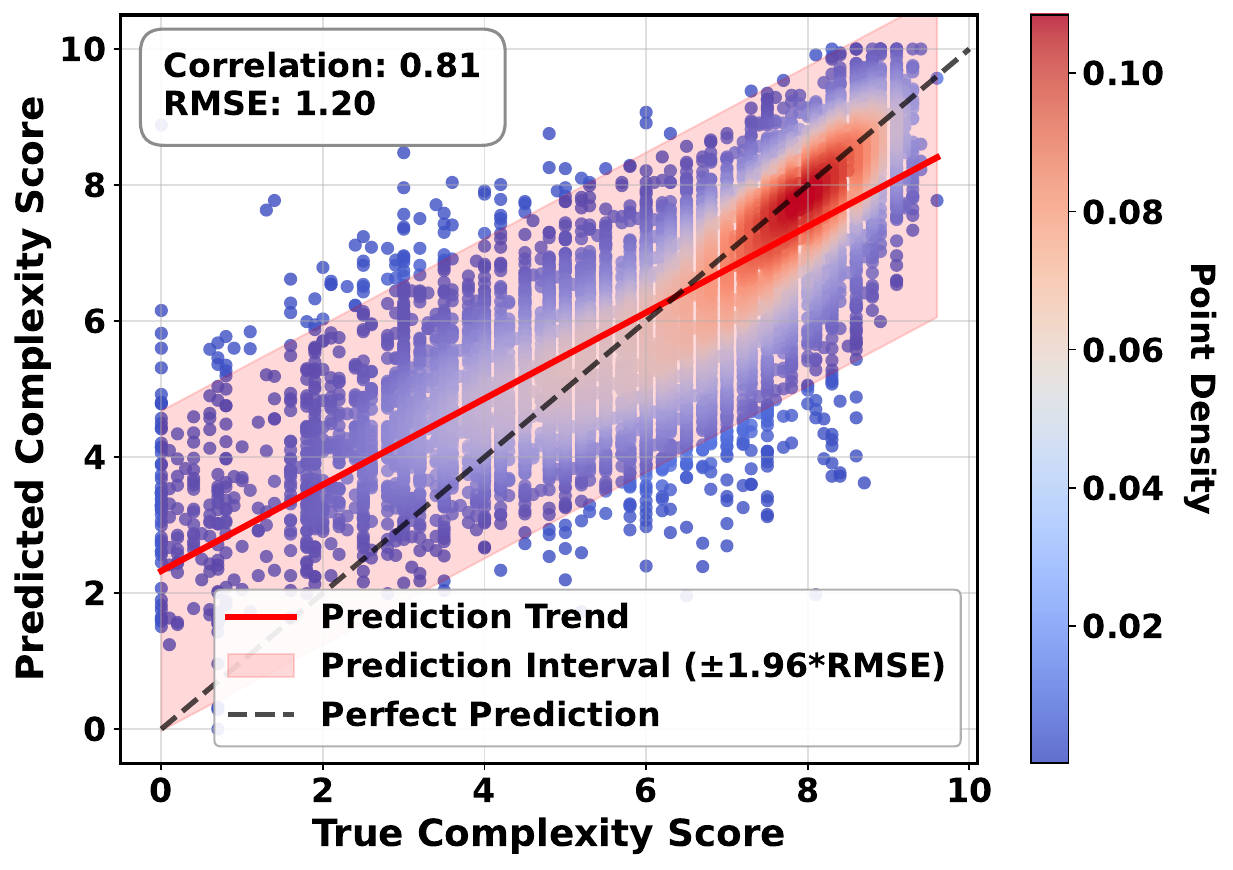}
    \caption{Qwen-3-14B}
    \label{fig:linear-qwen-14b}
  \end{subfigure}
  
  \caption{Supplement to the results of Fig. \ref{fig:linear probe}}
  \label{fig:more linear probe}
\end{figure*}

Fig. \ref{fig:more linear probe} presents linear probe performance across a broader set of LLMs.

\subsection{Relationship between Complexity Scores and Allocated Feature Counts}
\begin{figure*}[t]
  \centering
  \begin{subfigure}[b]{0.24\linewidth}
    \centering
    \includegraphics[width=\linewidth]{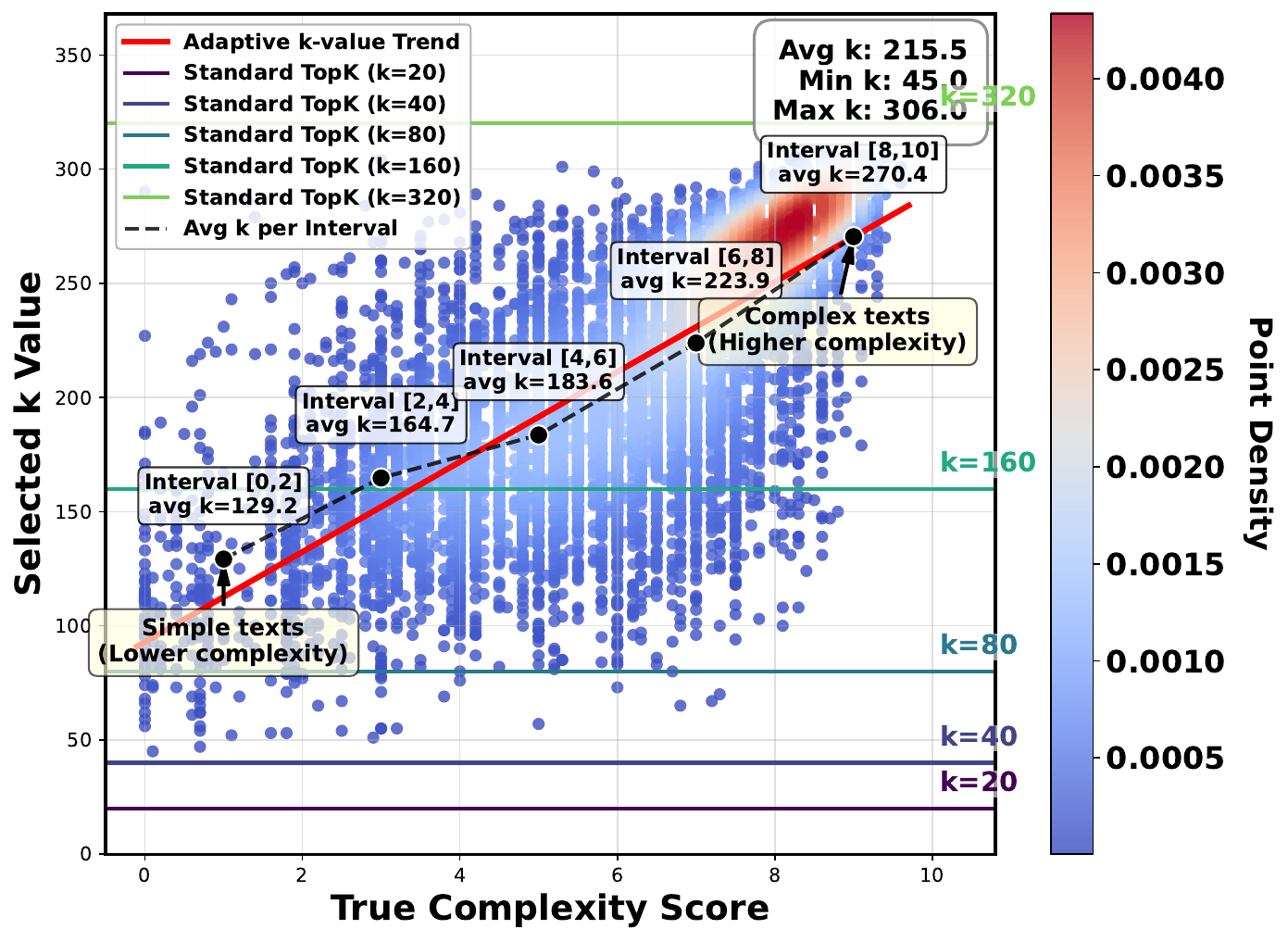}
    \caption{Pythia-160M}
    \label{fig:k-160}
  \end{subfigure}
  \hfill
  \begin{subfigure}[b]{0.24\linewidth}
    \centering
    \includegraphics[width=\linewidth]{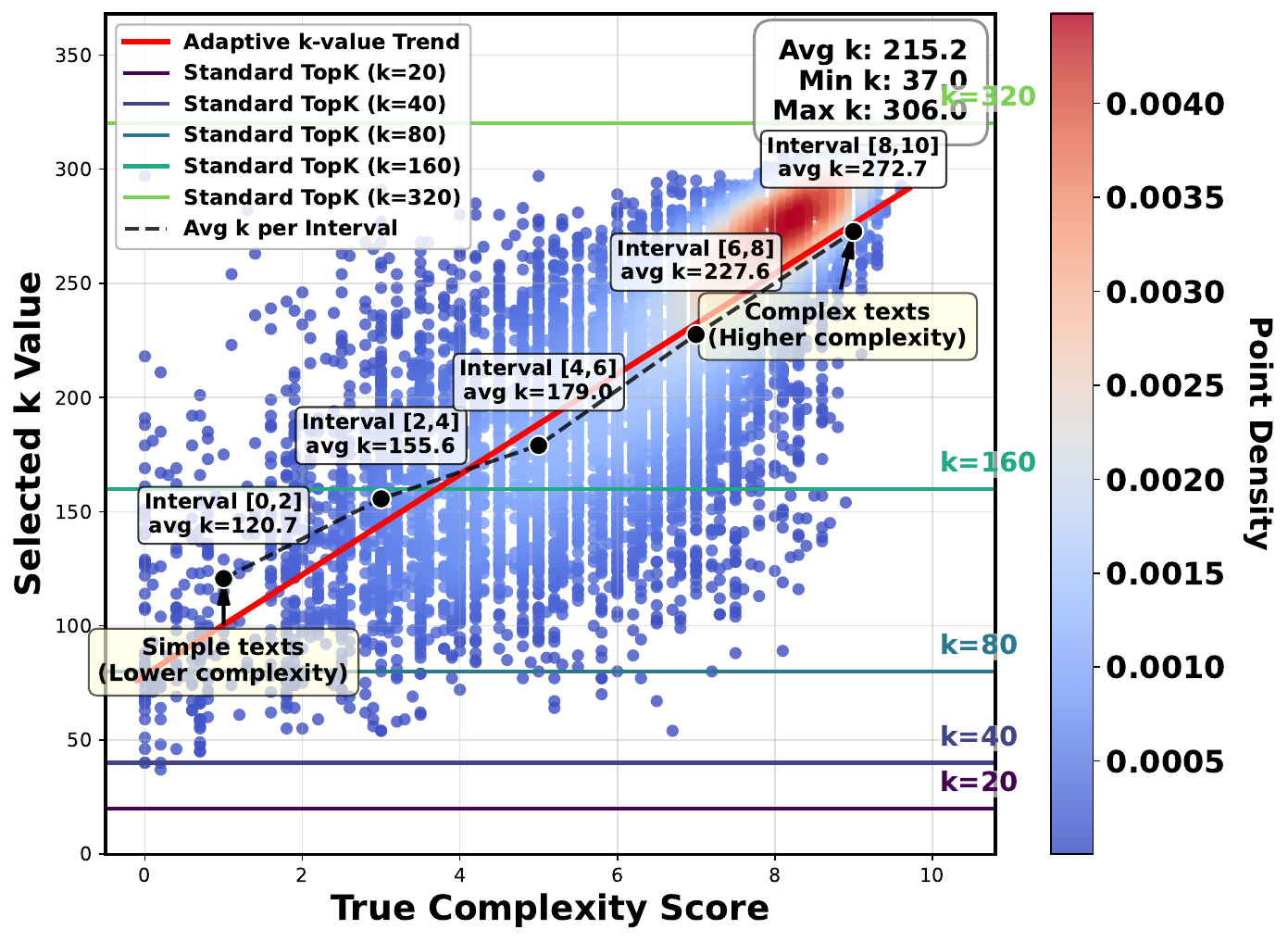}
    \caption{Gemma-2-9B}
    \label{fig:k-gemma-9b}
  \end{subfigure}
  \hfill
  \begin{subfigure}[b]{0.24\linewidth}
    \centering
    \includegraphics[width=\linewidth]{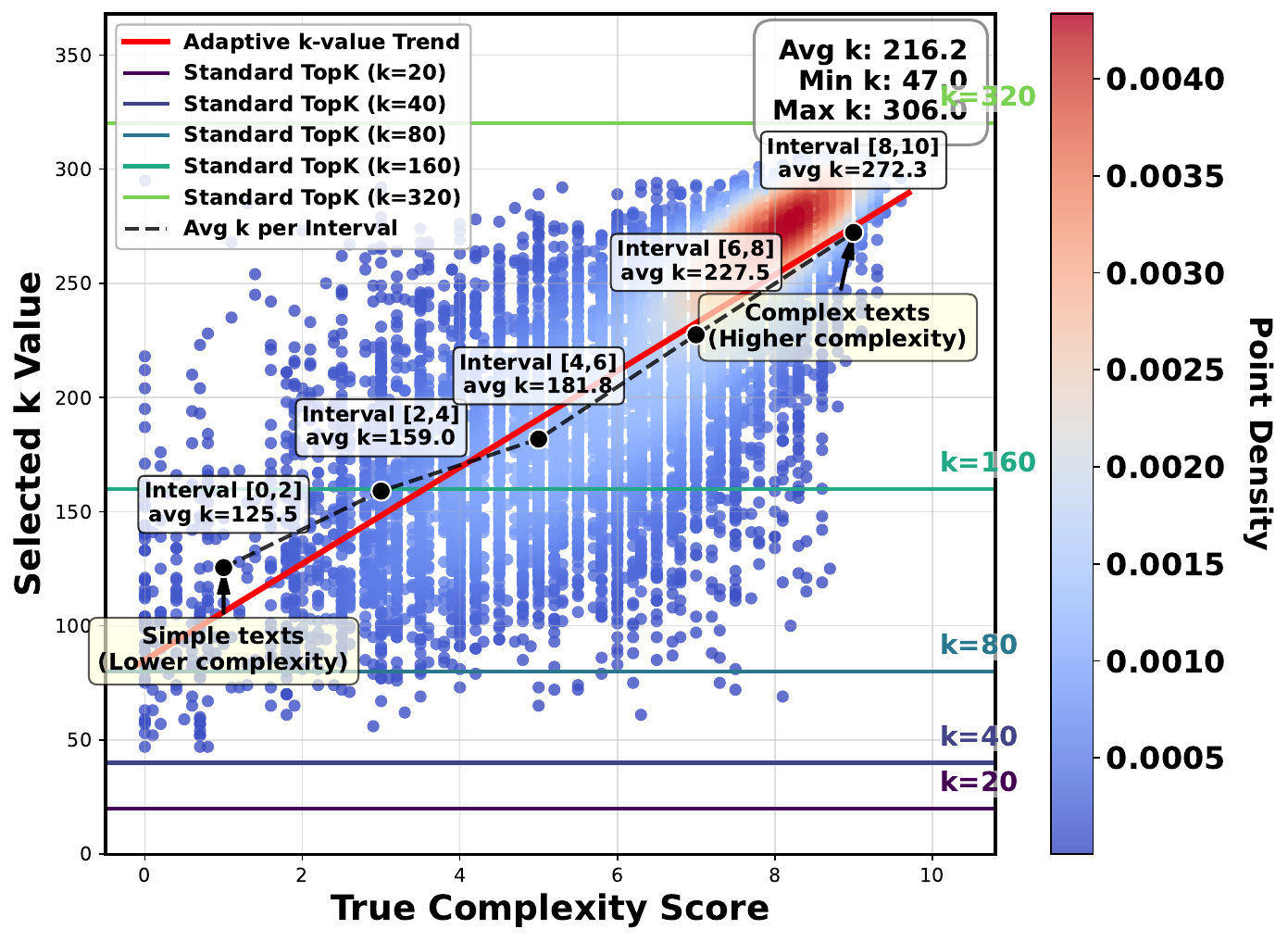}
    \caption{Llama-3.1-8B}
    \label{fig:k-llama-8b}
  \end{subfigure}
  \hfill
  \begin{subfigure}[b]{0.24\linewidth}
    \centering
    \includegraphics[width=\linewidth]{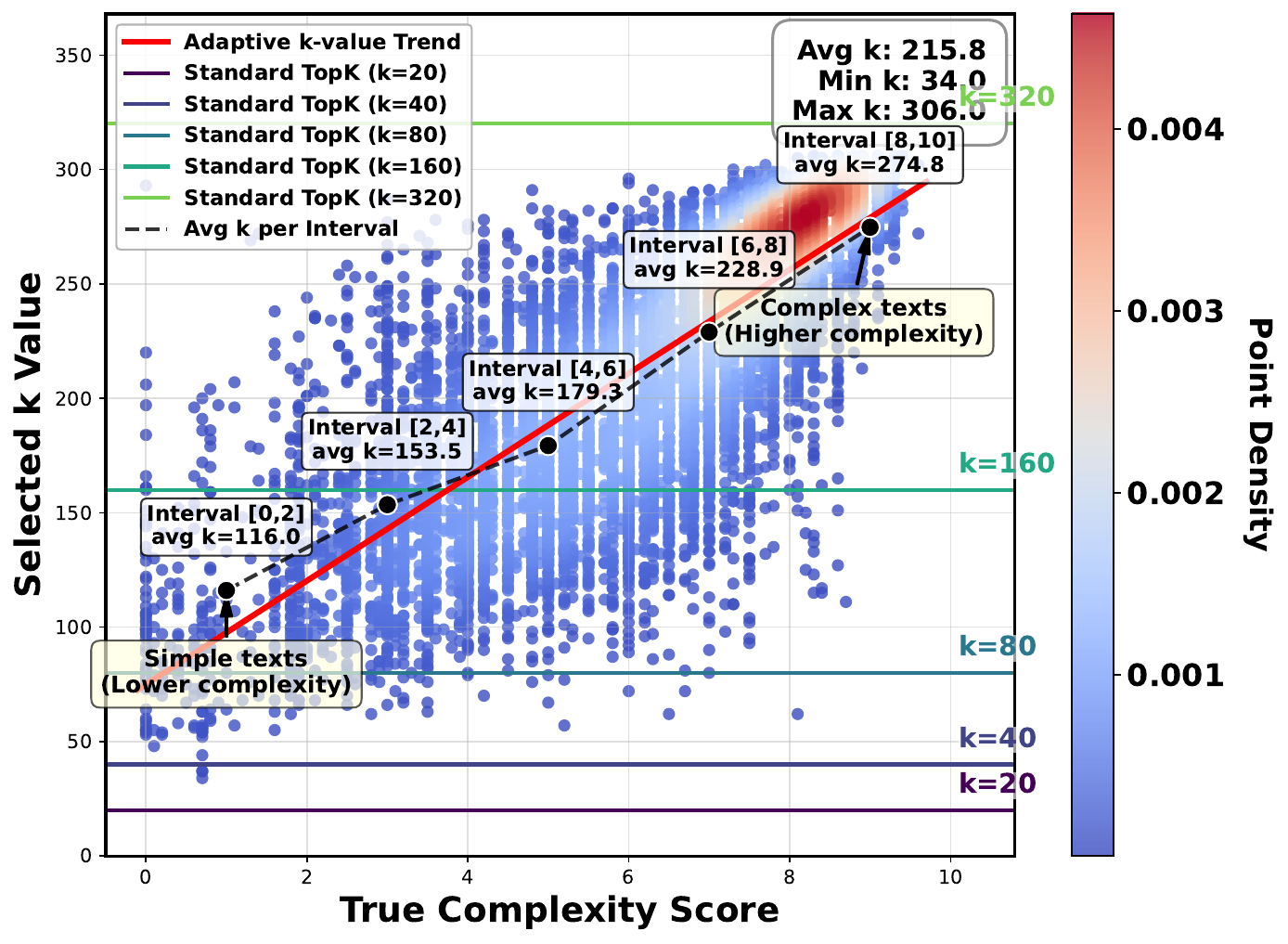}
    \caption{Qwen-3-14B}
    \label{fig:k-qwen-14b}
  \end{subfigure}
  
  \caption{Supplement to the results of Fig. \ref{fig:k}}
  \label{fig:more k}
\end{figure*}

Fig. \ref{fig:more k} presents the relationship between complexity scores and allocated feature counts across a broader set of LLMs.

\subsection{Pareto Frontier Results}
\begin{figure*}[t]
  \centering
  \begin{subfigure}[b]{0.24\linewidth}
    \centering
    \includegraphics[width=\linewidth]{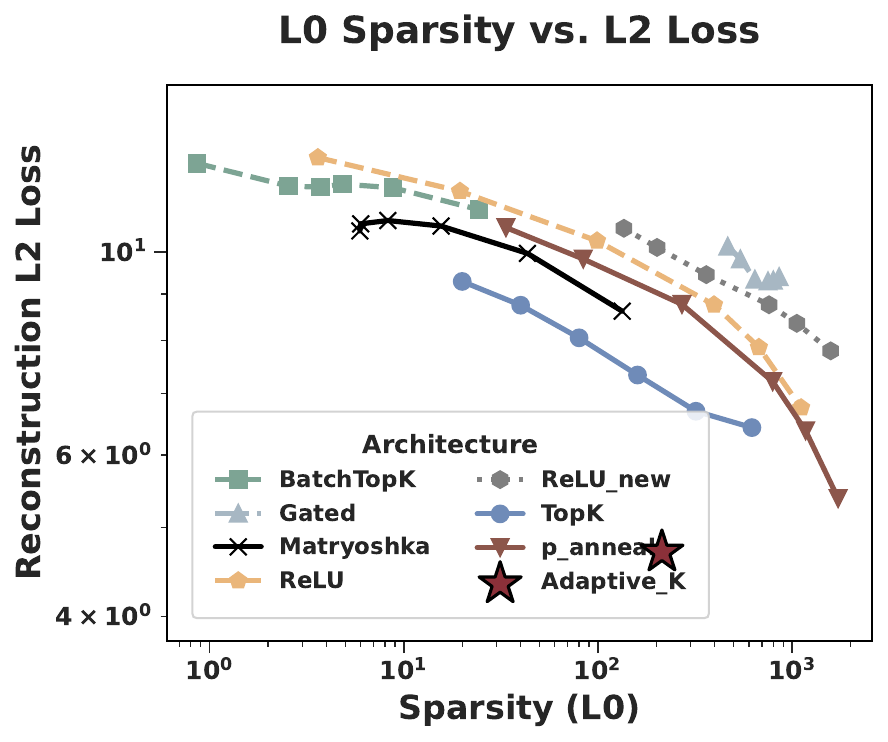}
    \caption{Pythia-160M}
  \end{subfigure}
  \hfill 
  \begin{subfigure}[b]{0.24\linewidth}
    \centering
    \includegraphics[width=\linewidth]{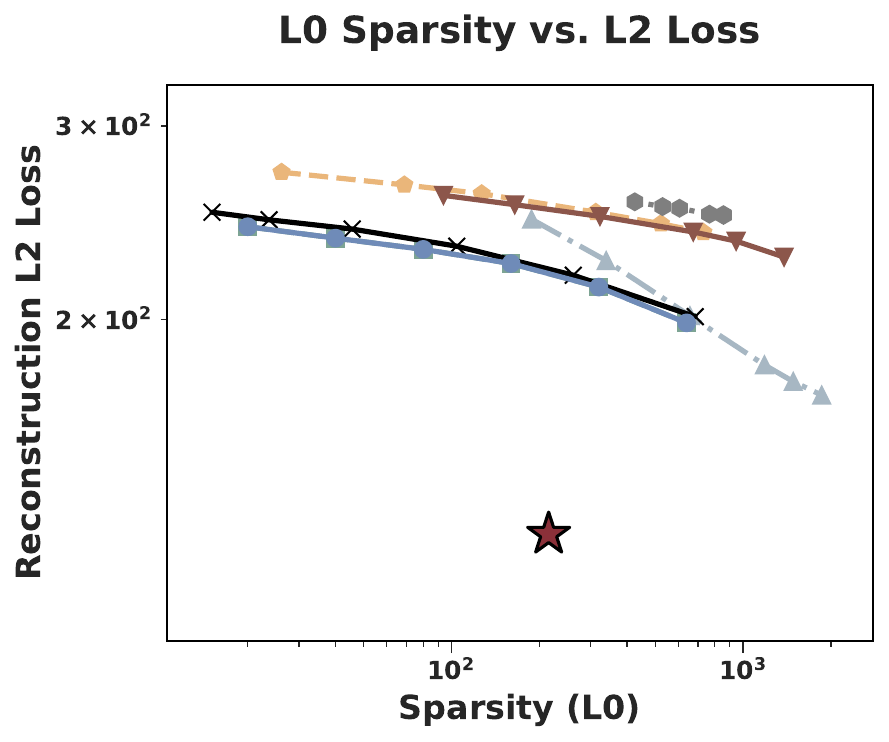}
    \caption{Gemma-2-9B}
  \end{subfigure}
  \hfill
  \begin{subfigure}[b]{0.24\linewidth}
    \centering
    \includegraphics[width=\linewidth]{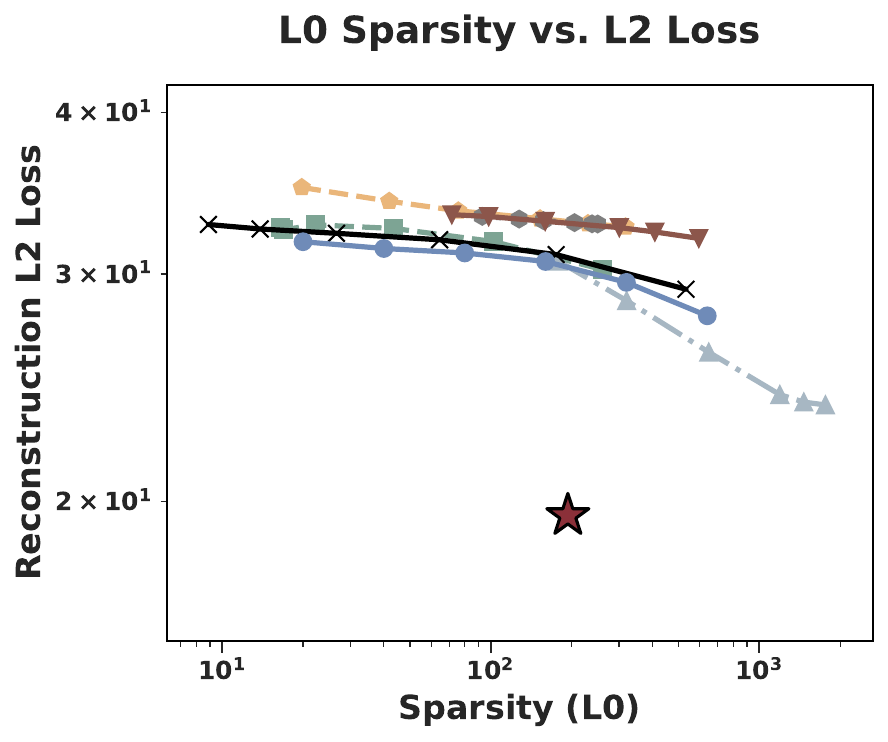}
    \caption{Llama-3.1-8B}
  \end{subfigure}
  \hfill
  \begin{subfigure}[b]{0.24\linewidth}
    \centering
    \includegraphics[width=\linewidth]{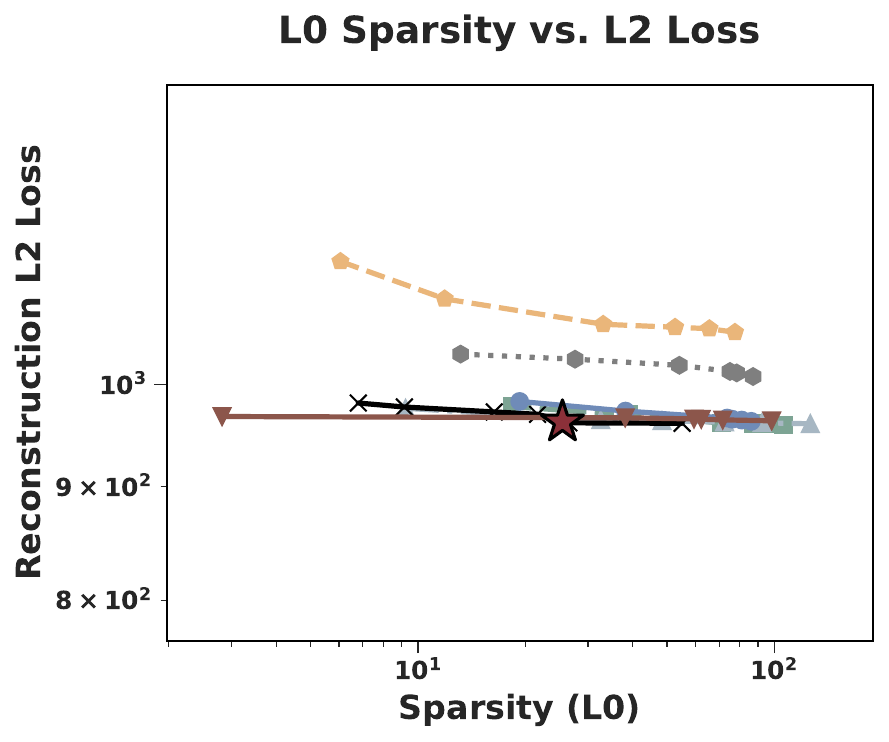}
    \caption{Qwen-3-14B}
  \end{subfigure}
  
  \caption{Supplement to the results of Fig. \ref{fig:pareto_l2}}
  \label{fig:more pareto_l2}
\end{figure*}

\begin{figure*}[t]
  \centering
  \begin{subfigure}[b]{0.24\linewidth}
    \centering
    \includegraphics[width=\linewidth]{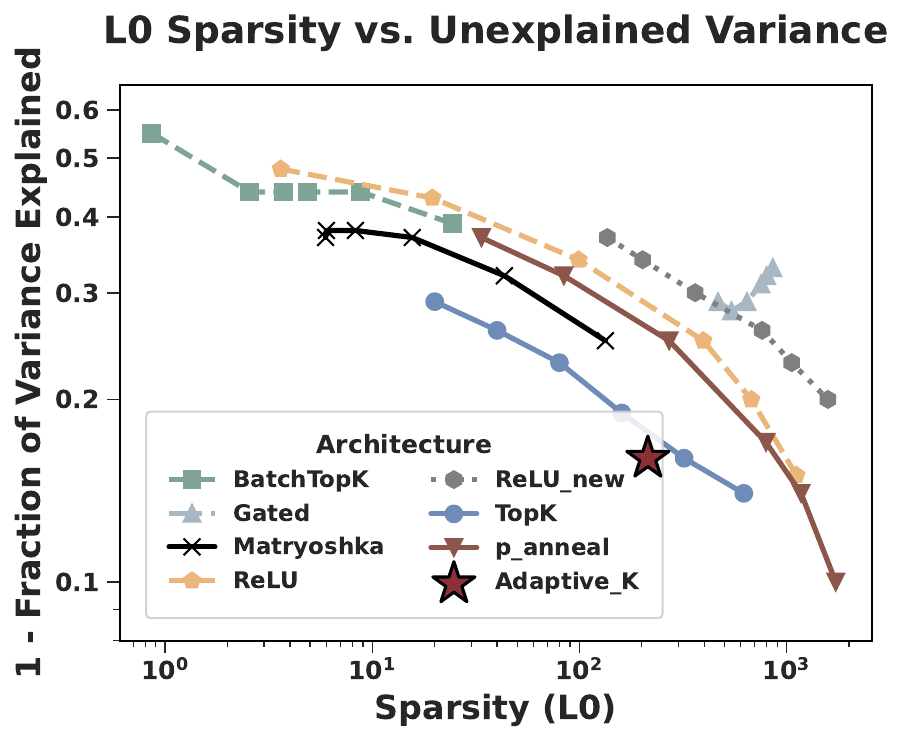}
    \caption{Pythia-160M}
  \end{subfigure}
  \hfill
  \begin{subfigure}[b]{0.24\linewidth}
    \centering
    \includegraphics[width=\linewidth]{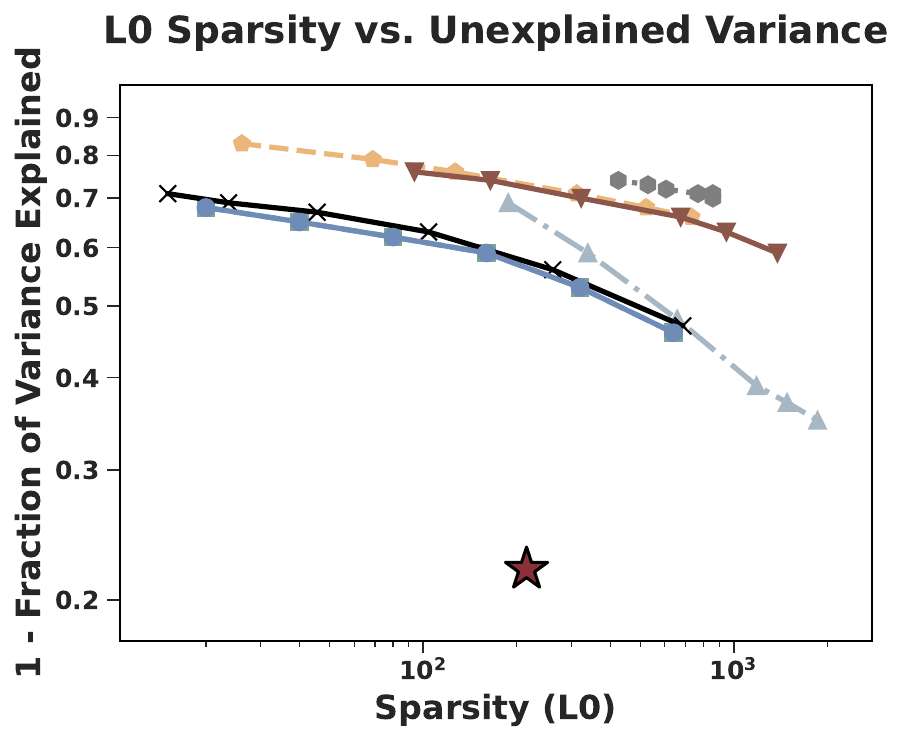}
    \caption{Gemma-2-9B}
  \end{subfigure}
  \hfill
  \begin{subfigure}[b]{0.24\linewidth}
    \centering
    \includegraphics[width=\linewidth]{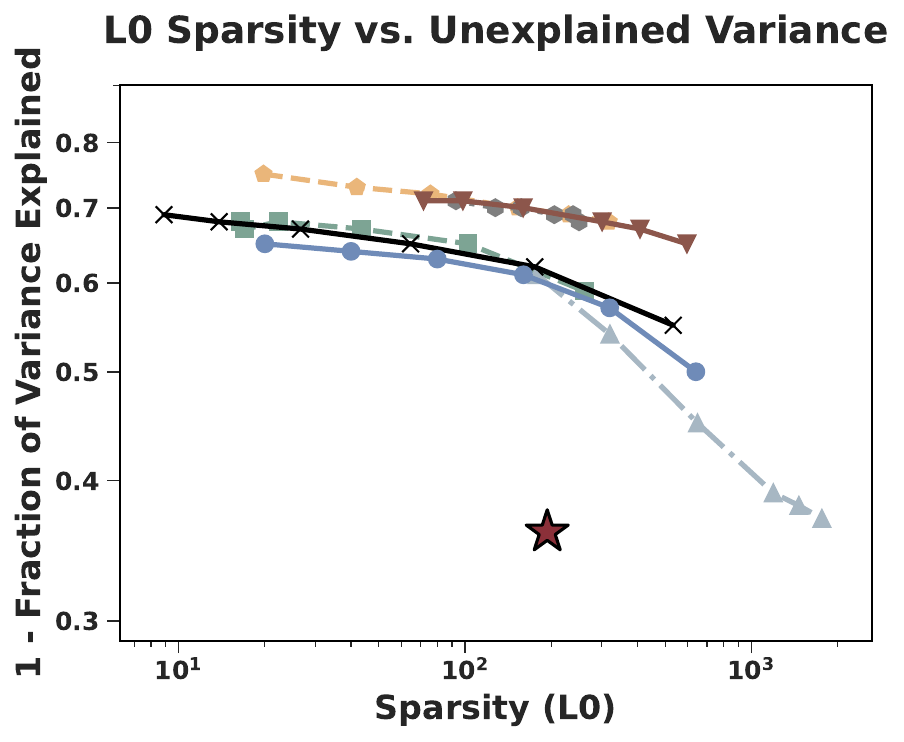}
    \caption{Llama-3.1-8B}
  \end{subfigure}
  \hfill
  \begin{subfigure}[b]{0.24\linewidth}
    \centering
    \includegraphics[width=\linewidth]{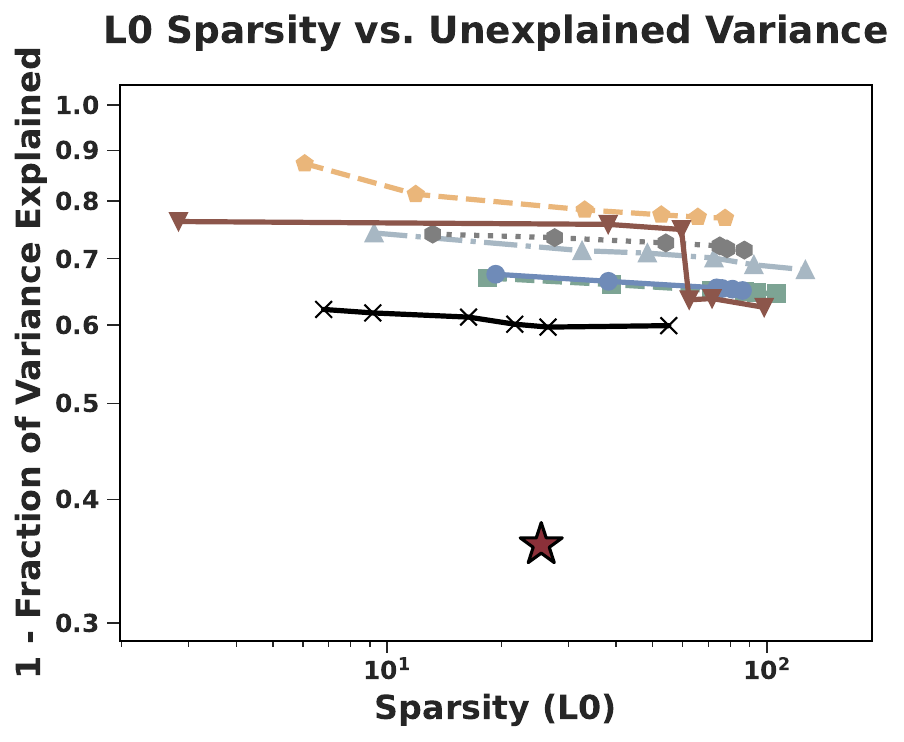}
    \caption{Qwen-3-14B}
  \end{subfigure}
  
  \caption{Supplement to the results of Fig. \ref{fig:pareto_unexplained}}
  \label{fig:more pareto_unexplained}
\end{figure*}

\begin{figure*}[t]
  \centering
  \begin{subfigure}[b]{0.24\linewidth}
    \centering
    \includegraphics[width=\linewidth]{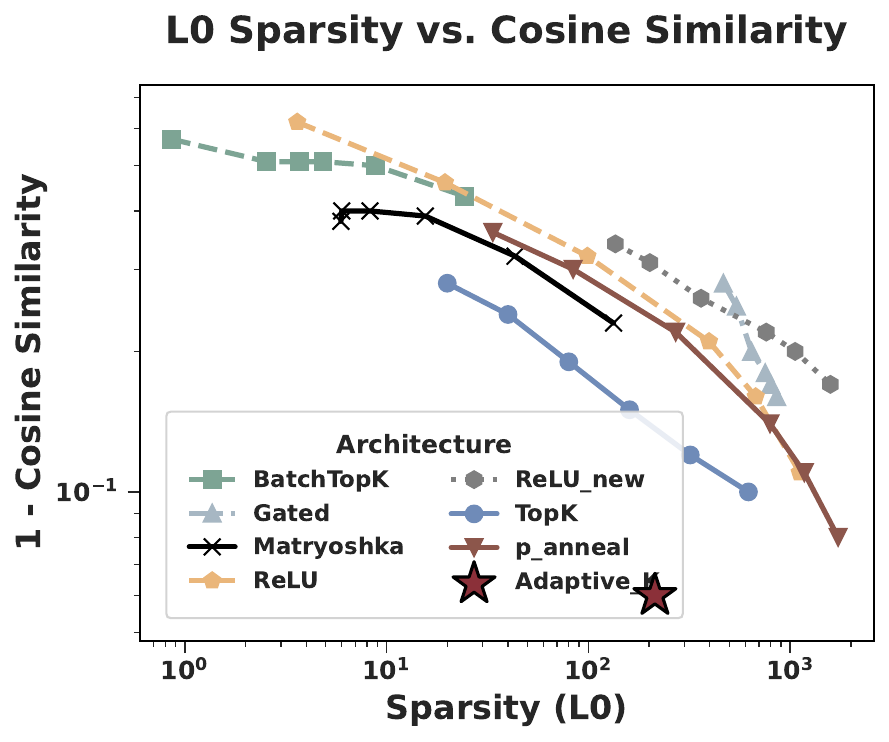}
    \caption{Pythia-160M}
  \end{subfigure}
  \hfill
  \begin{subfigure}[b]{0.24\linewidth}
    \centering
    \includegraphics[width=\linewidth]{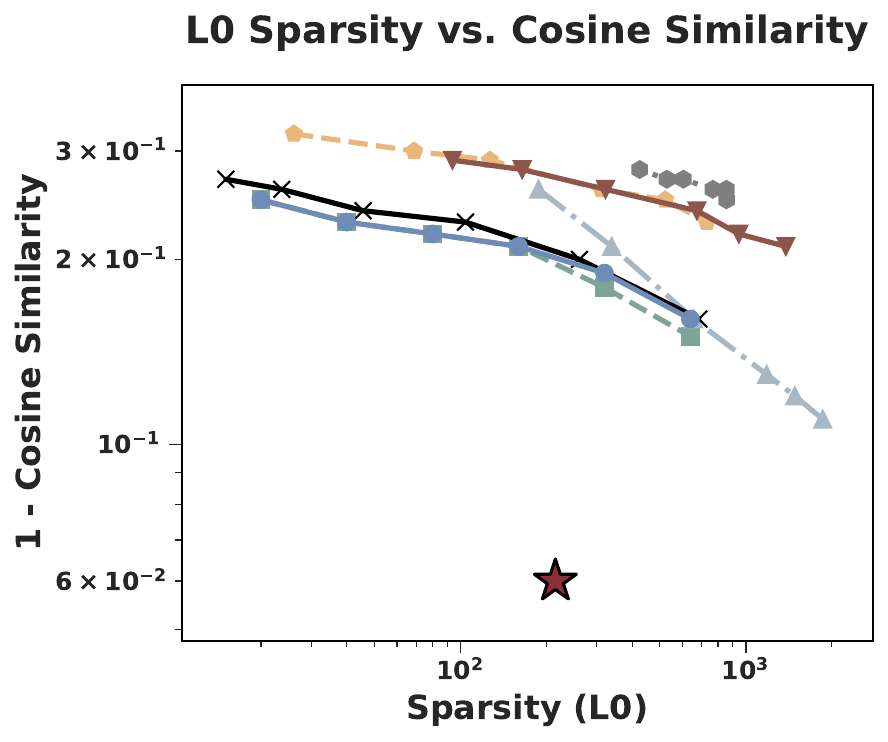}
    \caption{Gemma-2-9B}
  \end{subfigure}
  \hfill
  \begin{subfigure}[b]{0.24\linewidth}
    \centering
    \includegraphics[width=\linewidth]{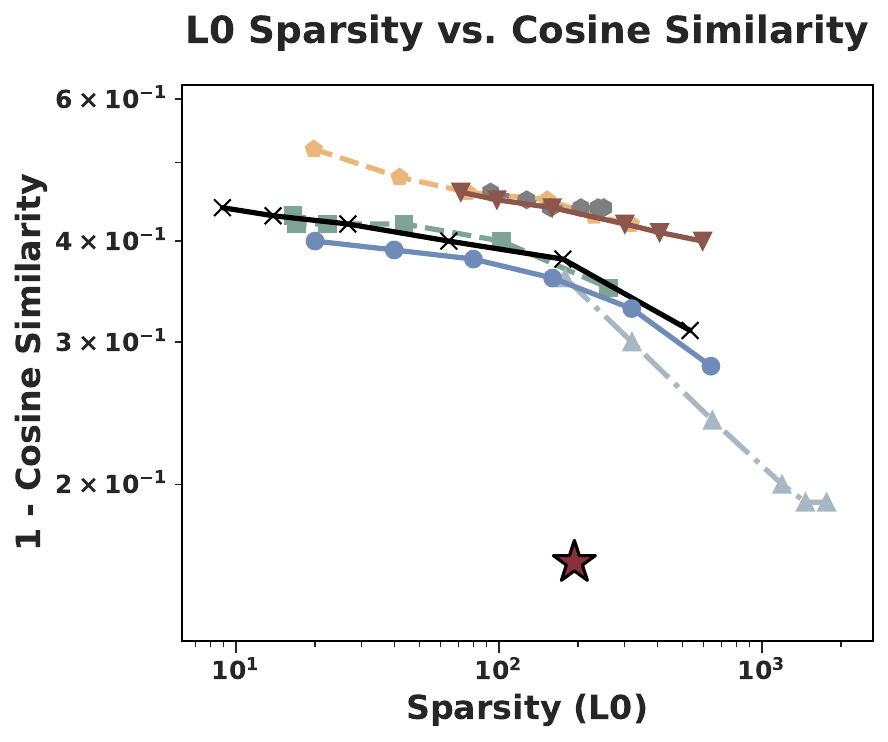}
    \caption{Llama-3.1-8B}
  \end{subfigure}
  \hfill
  \begin{subfigure}[b]{0.24\linewidth}
    \centering
    \includegraphics[width=\linewidth]{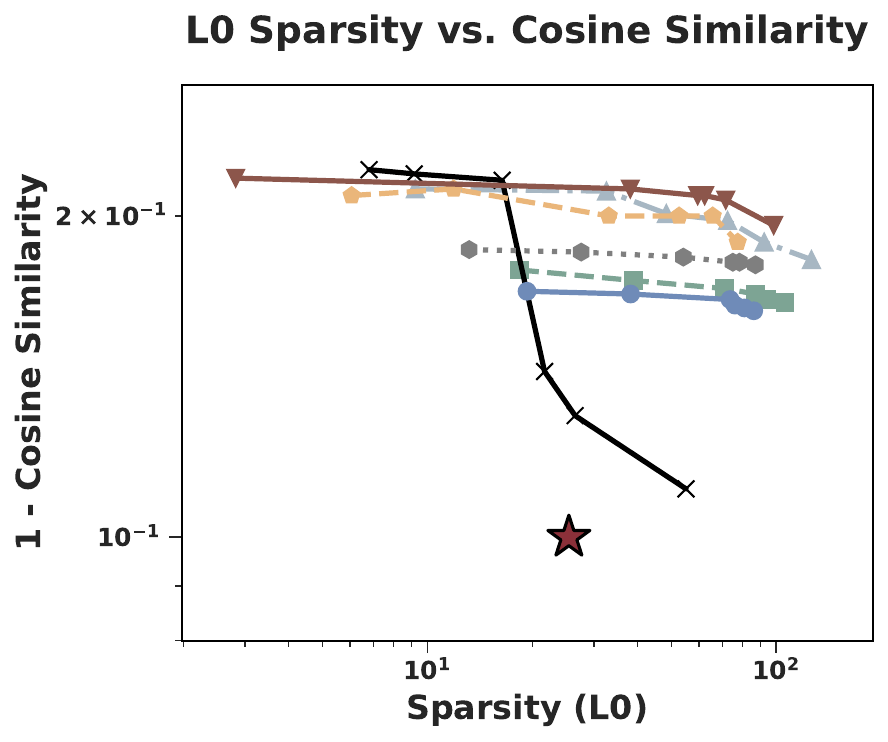}
    \caption{Qwen-3-14B}
  \end{subfigure}
  
  \caption{Supplement to the results of Fig. \ref{fig:pareto_cosine}}
  \label{fig:more pareto_cosine}
\end{figure*}

Fig. \ref{fig:more pareto_l2}, \ref{fig:more pareto_unexplained} and \ref{fig:more pareto_cosine} presents L2 Loss, Unexplained Variance and Cosine Similarity pareto frontier results across a broader set of LLMs.

\begin{table*}[t]
  \centering
  \caption{Performance of encoder-only and encoder-decoder models}
  \label{tab:encoder_models}
  \begin{tabular}{lcccccc}
    \toprule
    Model & Layer & SAE Position & Explained Variance & Cosine Similarity & L2 Ratio \\
    \midrule
    BERT-340M & 8 & encoder & 0.66 & 0.89 & 0.8225 \\
    T5-small  & 3 & encoder & 0.97 & 0.98 & 1.0093 \\
    T5-small  & 3 & decoder & 0.74 & 0.97 & 1.0139 \\
    \bottomrule
  \end{tabular}
\end{table*}

\begin{table*}[t]
  \centering
  \caption{Performance across models with varying $\lambda$ values}
  \label{tab:hyper_lambda}
  \scalebox{0.8}{
  \begin{tabular}{ccccccc}
    \toprule
    $\lambda$ & Gemma-2-2b & Gemma-2-9b & Llama-3.1-8b & Qwen-3-8b & Qwen-3-14b & Phi-4-14b \\
    \midrule
    0.001  & 1.1937 & 1.2663 & 1.2612 & 1.0875 & 1.2425 & 1.1934 \\
    0.01   & 1.1937 & 1.2663 & 1.2612 & 1.0875 & \textbf{1.2424} & 1.1934 \\
    0.1    & 1.1937 & 1.2663 & 1.2612 & 1.0875 & 1.2425 & 1.1934 \\
    1.0    & 1.1937 & \textbf{1.2662} & 1.2612 & 1.0875 & 1.2425 & 1.1934 \\
    10.0   & 1.1937 & 1.2663 & 1.2611 & 1.0874 & 1.2424 & 1.1934 \\
    100.0  & \textbf{1.1935} & 1.2663 & \textbf{1.2605} & \textbf{1.0873} & 1.2424 & \textbf{1.1933} \\
    1000.0 & 1.1937 & 1.2663 & 1.2608 & 1.0875 & 1.2424 & 1.1933 \\
    \bottomrule
  \end{tabular}}
\end{table*}

\section{Additional Linear Probe Evaluation}

\subsection{PCA Dimensionality Reduction Experiments}
\label{PCA}
To provide more rigorous statistical validation, PCA dimensionality reduction experiments were conducted on the activation data. Specifically, since the original probe dimensionality is 2048 (for Gemma-2-2B), it could theoretically "memorize" substantial information. However, if complexity is truly linearly encoded, then only a few dimensions should be needed for accurate prediction. 

\begin{figure*}[t]
  \centering
  \begin{subfigure}[b]{0.32\linewidth}
    \centering
    \includegraphics[width=\linewidth]{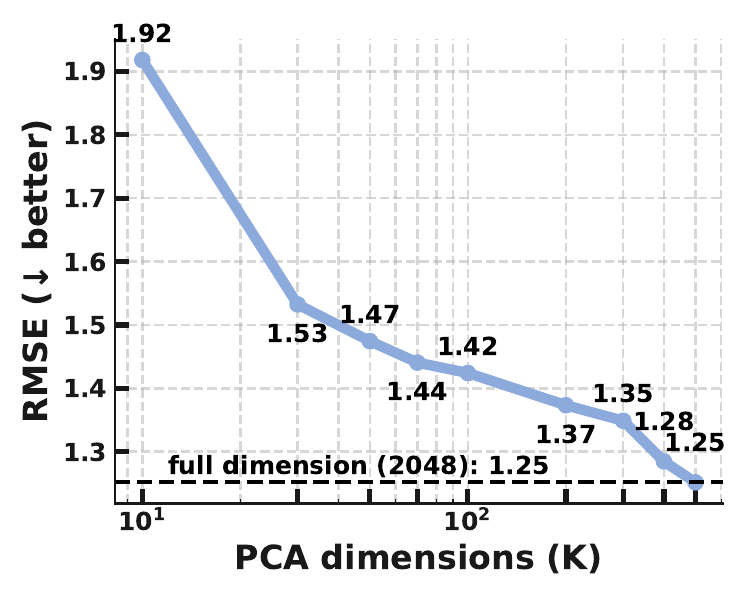}
    \caption{RSME vs. PCA dimensions}
    \label{fig:pca-rmse}
  \end{subfigure}
  \hfill
  \begin{subfigure}[b]{0.32\linewidth}
    \centering
    \includegraphics[width=\linewidth]{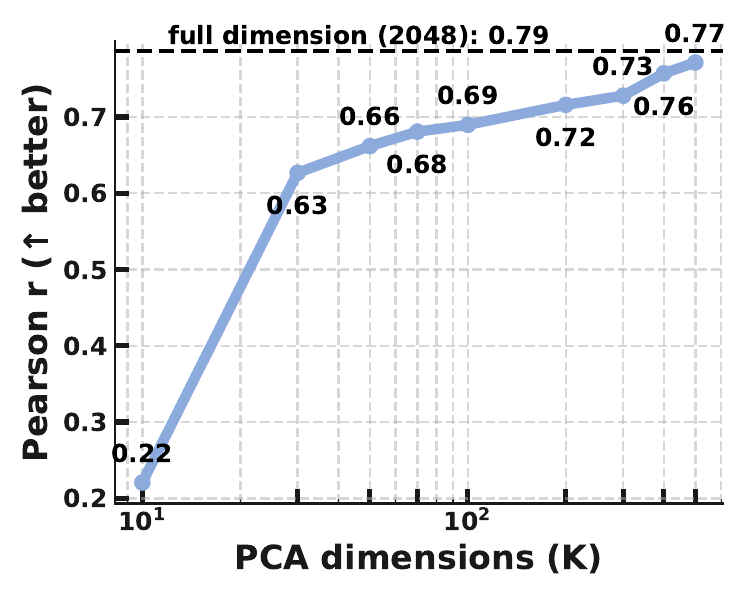}
    \caption{Pearson vs. PCA dimensions}
    \label{fig:pca-pearson}
  \end{subfigure}
  \hfill
  \begin{subfigure}[b]{0.32\linewidth}
    \centering
    \includegraphics[width=\linewidth]{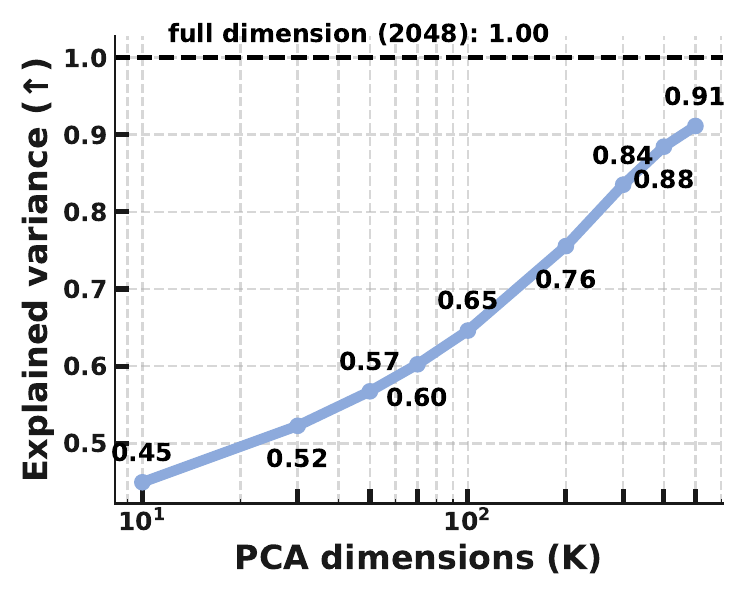}
    \caption{Exp Var vs. PCA dimensions}
    \label{fig:pca-explained}
  \end{subfigure}
  
  \caption{PCA dimensionality reduction experiments on Gemma-2-2B}
  \label{fig:pca}
\end{figure*}

We performed PCA decomposition on all context activation vectors, identifying the top k directions with maximum variance (principal components), and projected the original 2048-dimensional activations onto these k directions (k=10-500). Probes for complexity prediction were then trained on the low-dimensional representations. Fig. \ref{fig:pca} shows the RMSE and Pearson correlation coefficients of probes using different numbers of principal components. Using just 200 principal components (9.8\% of full dimensionality), the probe achieves a Pearson correlation of 0.72 compared to 0.79 for the full-dimensional probe, recovering 91\% of the predictive performance while explaining 75.6\% of the variance. With 400 components (19.5\% of dimensions), performance reaches 96\% of the full probe (Pearson: 0.76 vs 0.79) with 88.4\% explained variance. This demonstrates that text complexity is indeed linearly encoded. If it were merely high-dimensional memorization, many more principal components would be required to achieve good predictive performance.

\subsection{Layer-wise Evaluation}
\label{layer-wise probe}
\begin{figure*}[t]
  \centering
  \hspace*{\fill}
  \begin{subfigure}[b]{0.328\linewidth}
    \centering
    \includegraphics[width=\linewidth]{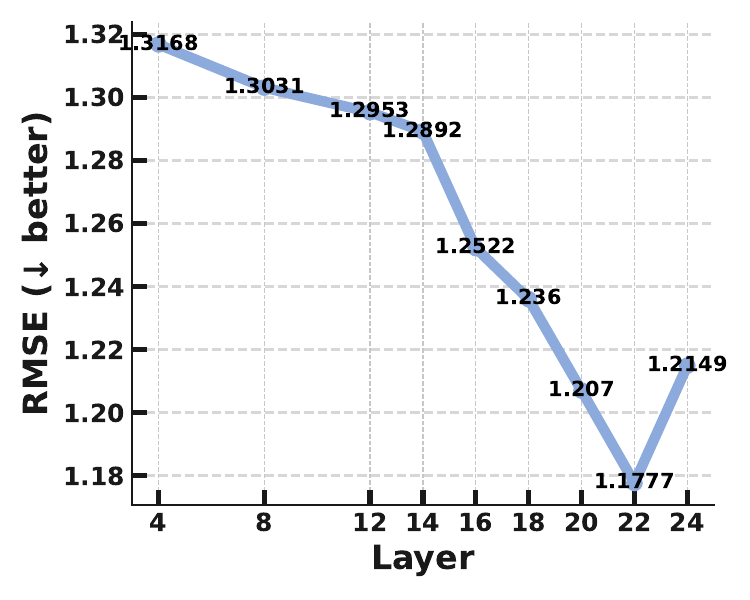}
    \label{fig:probe-layer-rmse}
  \end{subfigure}
  \hspace{0.02\linewidth}
  \begin{subfigure}[b]{0.328\linewidth}
    \centering
    \includegraphics[width=\linewidth]{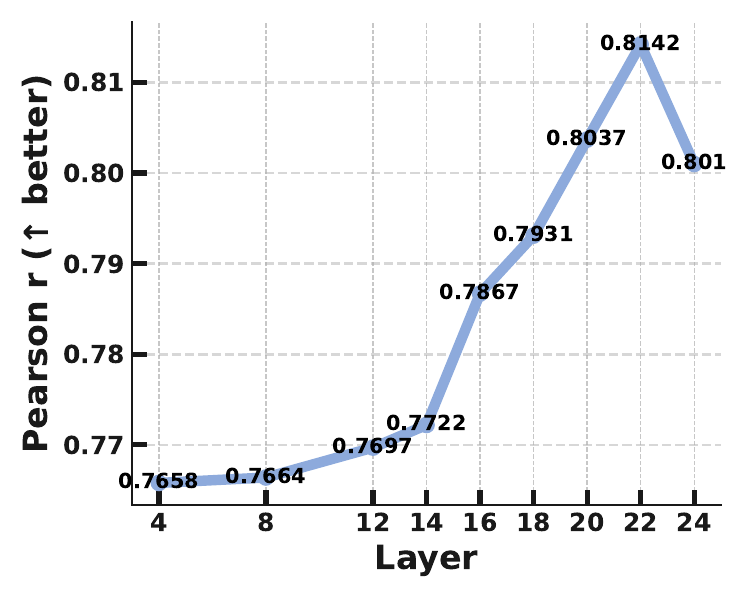}
    \label{fig:probe-layer-pearson}
  \end{subfigure}
  \hspace*{\fill}
  
  \caption{Layer-wise evaluation of linear probe on Gemma-2-2B}
  \label{fig:probe-layer}
\end{figure*}

To further validate the effectiveness of complexity prediction using linear probes, we conducted layer-wise experiments across all 26 layers of Gemma-2-2B to understand how complexity representations develop throughout the model. Following previous work \citep{gurnee2023language}, which found that probing performance exhibits a characteristic pattern of initial growth followed by saturation as layer depth increases, our experimental results in Fig. \ref{fig:probe-layer} demonstrate the same trajectory. Starting from layer 4 with a Pearson correlation of 0.766, performance steadily improves through the middle layers, reaching peak performance at layer 22 with a correlation of 0.814 and RMSE of 1.18. Beyond this point, performance plateaus and even slightly decreases (layer 24: 0.801). This demonstrates that LLM representations contain complexity information in the same way they contain spatial and temporal information, and that deeper layers are progressively better at capturing text complexity.

\section{Additional SAE Evaluation}
\label{sec:additional-experiments-appendix}

\subsection{Layer-wise Evaluation}
\label{analysis of layers}
\begin{figure*}[t]
  \centering
  \begin{subfigure}[b]{0.32\linewidth}
    \centering
    \includegraphics[width=\linewidth]{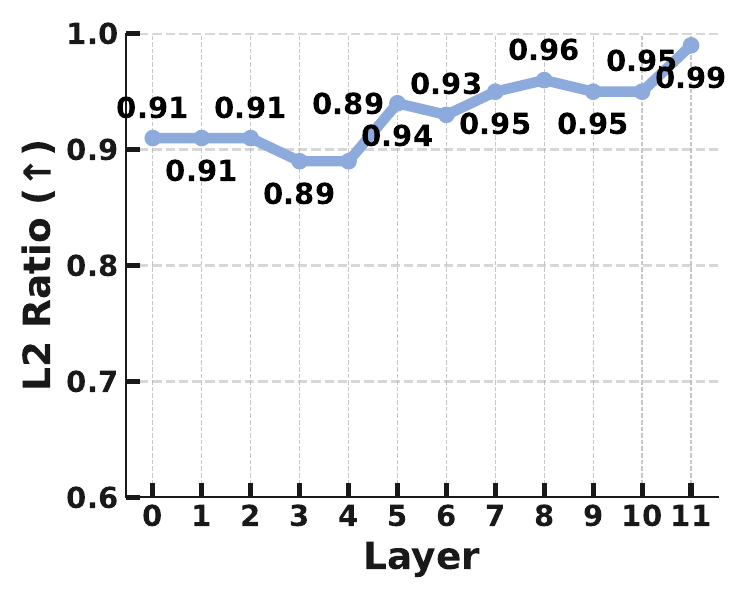}
    \caption{Layer-wise L2 Ratio}
  \end{subfigure}
  \hfill
  \begin{subfigure}[b]{0.32\linewidth}
    \centering
    \includegraphics[width=\linewidth]{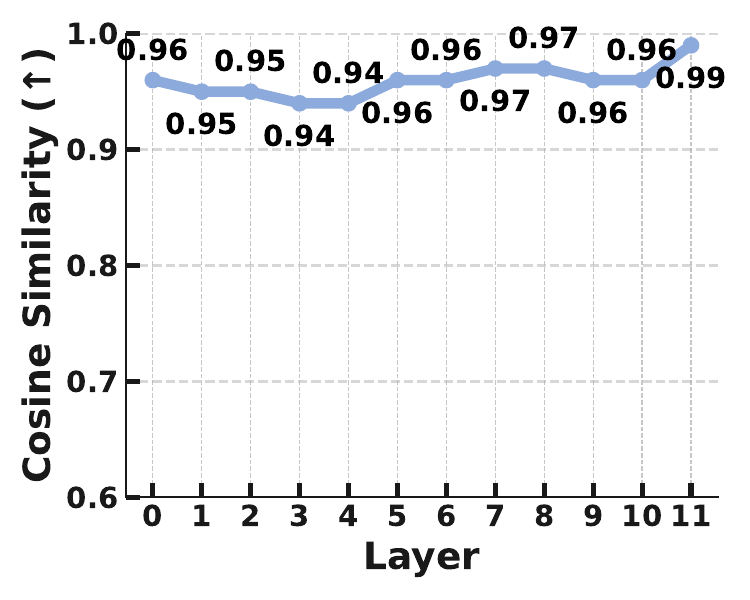}
    \caption{Layer-wise Cosine Similarity}
  \end{subfigure}
  \hfill
  \begin{subfigure}[b]{0.32\linewidth}
    \centering
    \includegraphics[width=\linewidth]{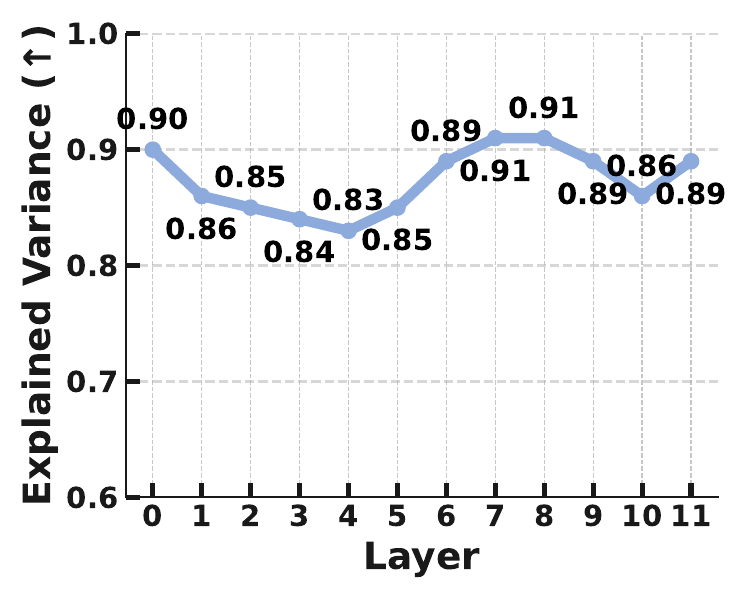}
    \caption{Layer-wise Explained Variance}
  \end{subfigure}
  
  \caption{Layer-wise performance on Pythia-160M}
  \label{fig:pythia160_layer}
\end{figure*}

\begin{figure*}[t]
  \centering
  \begin{subfigure}[b]{0.32\linewidth}
    \centering
    \includegraphics[width=\linewidth]{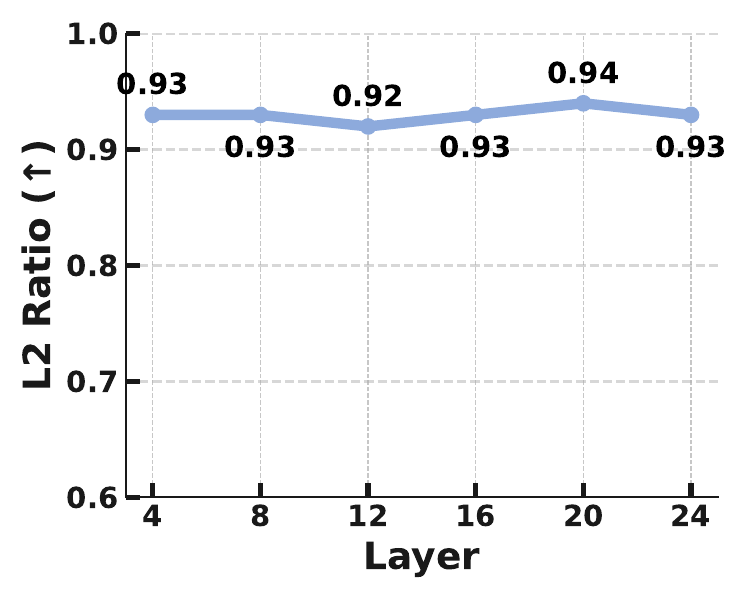}
    \caption{Layer-wise L2 Ratio}
  \end{subfigure}
  \hfill
  \begin{subfigure}[b]{0.32\linewidth}
    \centering
    \includegraphics[width=\linewidth]{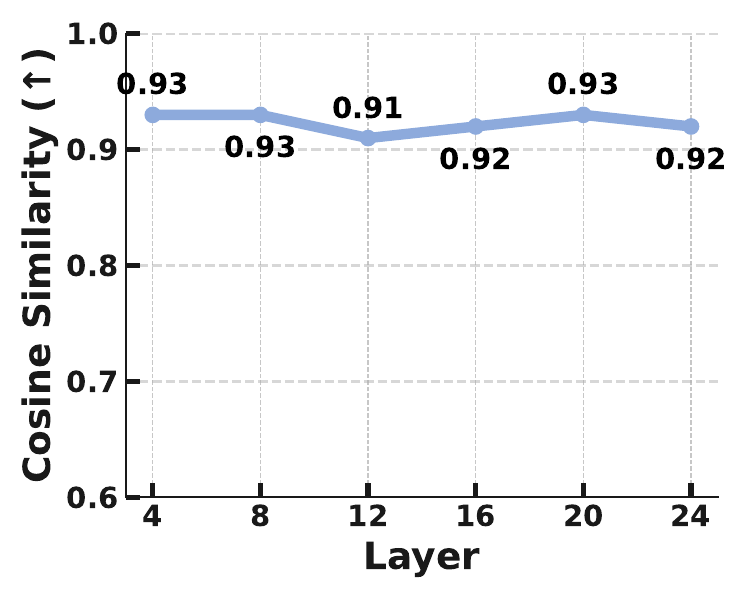}
    \caption{Layer-wise Cosine Similarity}
  \end{subfigure}
  \hfill
  \begin{subfigure}[b]{0.32\linewidth}
    \centering
    \includegraphics[width=\linewidth]{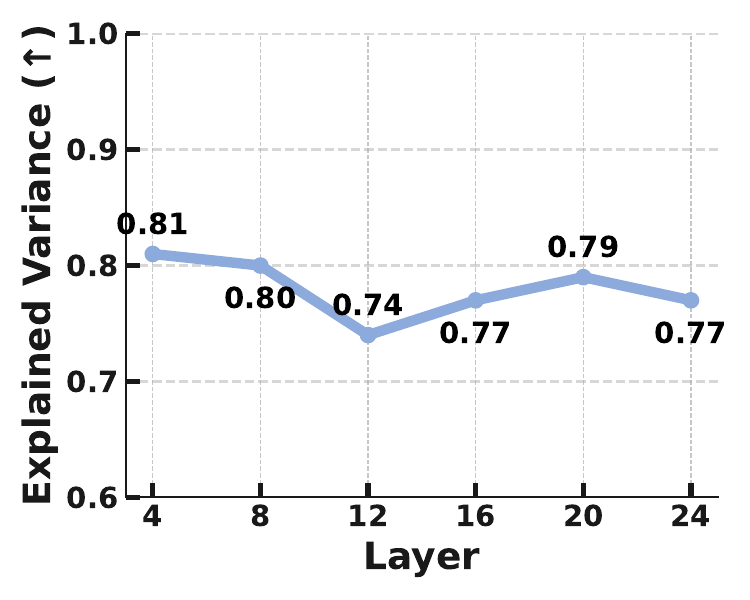}
    \caption{Layer-wise Explained Variance}
  \end{subfigure}
  
  \caption{Layer-wise performance on Gemma-2-2B}
  \label{fig:gemma_layer}
\end{figure*}

We trained AdaptiveK SAE on each layer of Pythia-160M and on layers 4, 8, 12, 16, 20, and 24 of the Gemma-2-2B model. A cross-layer comparison of key performance metrics is presented in Fig. \ref{fig:pythia160_layer} and \ref{fig:gemma_layer}. L2 ratio measures the proportion between the L2 norms of reconstructed and original activations, with values closer to 1 indicating preservation of the original activation magnitude. AdaptiveK exhibits robust performance across all tested layers in both models, with Explained Variance, Cosine Similarity, and L2 Ratio consistently above 0.74, 0.91, and 0.89 respectively. This confirms the algorithm's generalizability throughout the entire LLM hierarchy.

\subsection{Extending to Encoder-only and Encoder-decoder Models}
\label{encoder-decoder}
We extended our model to encoder-only BERT and encoder-decoder T5, with results shown in Tab. \ref{tab:encoder_models}. BERT-340M achieved 0.89 cosine similarity and 0.82 L2 ratio, demonstrating successful adaptation to bidirectional attention mechanisms. T5-small results show the encoder outperforms the decoder by 23.56\% in explained variance, indicating that input understanding tasks are relatively regular while generation tasks are more complex. The encoder also performs better on other metrics, suggesting that encoders focus on understanding input semantic representations while decoders must handle both understanding and generation tasks simultaneously. All results maintain good reconstruction quality (cosine similarity more than 0.89, L2 ratio close to 1.0), proving the cross-architecture universality of ``complexity → more features''.

\subsection{Other Evaluation Metrics}
\label{Other Evaluation Metrics}
In this section, we evaluate the AdaptiveK SAE using five metrics from SAEBench \citep{karvonen2025saebench}. For the Feature Absorption metric, we directly compare the results of AdaptiveK SAE with those of the baseline SAEs reported in SAEBench, \textbf{noting that their training dataset was 2000 times larger than ours}. For the remaining metrics, (Spurious Correlation Removal, Targeted Probe Perturbation, Resolving Attribute-Value Entanglements in Language Models, and Sparse Probing), the AdaptiveK SAE is compared against baseline SAEs trained on an identical amount of training data.

\subsubsection{Feature Absorption}

One of the primary objectives of SAEs is to enhance feature interpretability through sparse activation patterns. However, when concepts exhibit hierarchical relationships, concept A (\emph{e.g.}, red) inherently implies a broader concept B (\emph{e.g.}, color), instead of dedicating separate, clear latents for both A and B, SAEs tend to develop a latent unit representing A and another representing ``B except A''. While this approach optimizes sparsity, it significantly compromises interpretability.

Following \citep{karvonen2025saebench}, Feature Absorption is measured using a first-letter classification task. It first establishes a ``ground truth'' directional vector $p$, for each first-letter concept by training linear probes on the base language model's activations ($a_{model}$). It then identifies a set of ``main'' SAE latents ($S_{main}$) expected to represent each letter. The core of the measurement involves analyzing individual instances (words). For an instance, if the main latents don't fully capture the ground truth signal (i.e., $\sum_{i \in S_{\text {main }}} a_i d_i \cdot p<a_{\text {model }} \cdot p$, where $a_i$ a is latent activation and $d_i$ its decoder vector), and other ``absorbing'' latents ($S_{abs}$) that align with $p$ compensate for this deficit, an absorption fraction is calculated. Let $P_{main} = \sum_{j \in S_{\text {main }}} a_j d_j \cdot p$ be the projection from main features, and $P_{compensated\_by\_absorbers}$ be projection from the top few absorbing latents that cover the signal portion not captured by $P_{main}$. The instance-level absorption fraction $f_{abs}$ is then:
\begin{equation}
f_{\mathrm{abs}}
= 
\frac{P_{\mathrm{compensated\_by\_absorbers}}}
     {P_{\mathrm{compensated\_by\_absorbers}} + P_{\mathrm{main}}}.
\end{equation}
This value closer to 0 means less feature absorption. We calculate two metrics: Mean Absorption Fraction per letter is the average of these $f_{abs}$ values over all relevant instances for that letter. Separately, an instance is marked for full absorption if stricter binary criteria are met: essentially, if main features are inactive and a single, dominant non-main latent (aligned with $p$) overwhelmingly represents the letter's ground truth signal. The Full Absorption Rate per letter is simply the proportion of relevant instances for that letter which meet these conditions for ``full absorption'', indicating how often extreme absorption occurs. 

\begin{figure*}[t]
  \centering
  \begin{subfigure}[b]{0.48\linewidth}
    \centering
    \includegraphics[width=\linewidth]{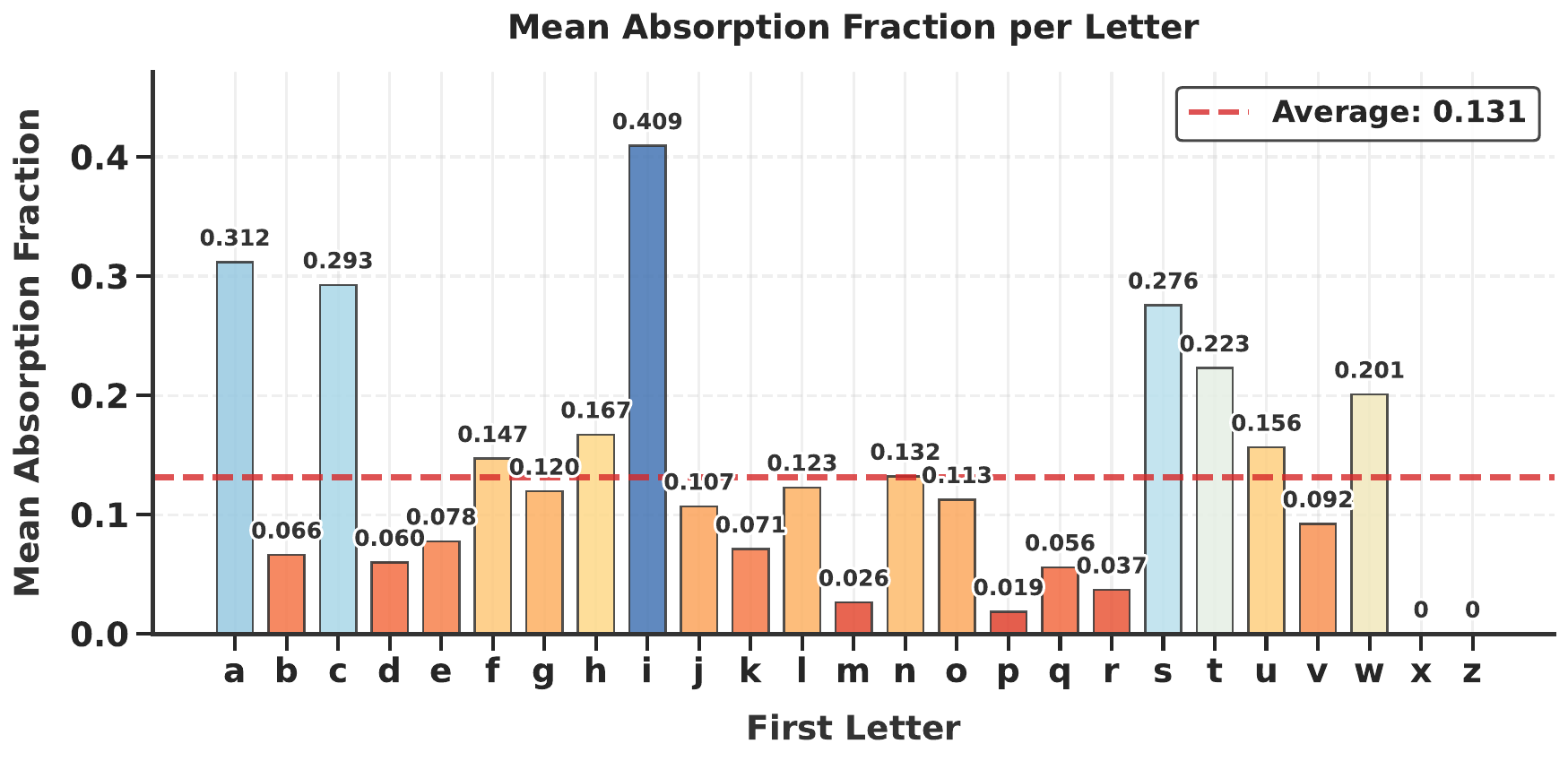}
    \caption{Mean Absorption Fraction per Letter}
    \label{fig:mean_absorption_160}
  \end{subfigure}
  \hfill
  \begin{subfigure}[b]{0.48\linewidth}
    \centering
    \includegraphics[width=\linewidth]{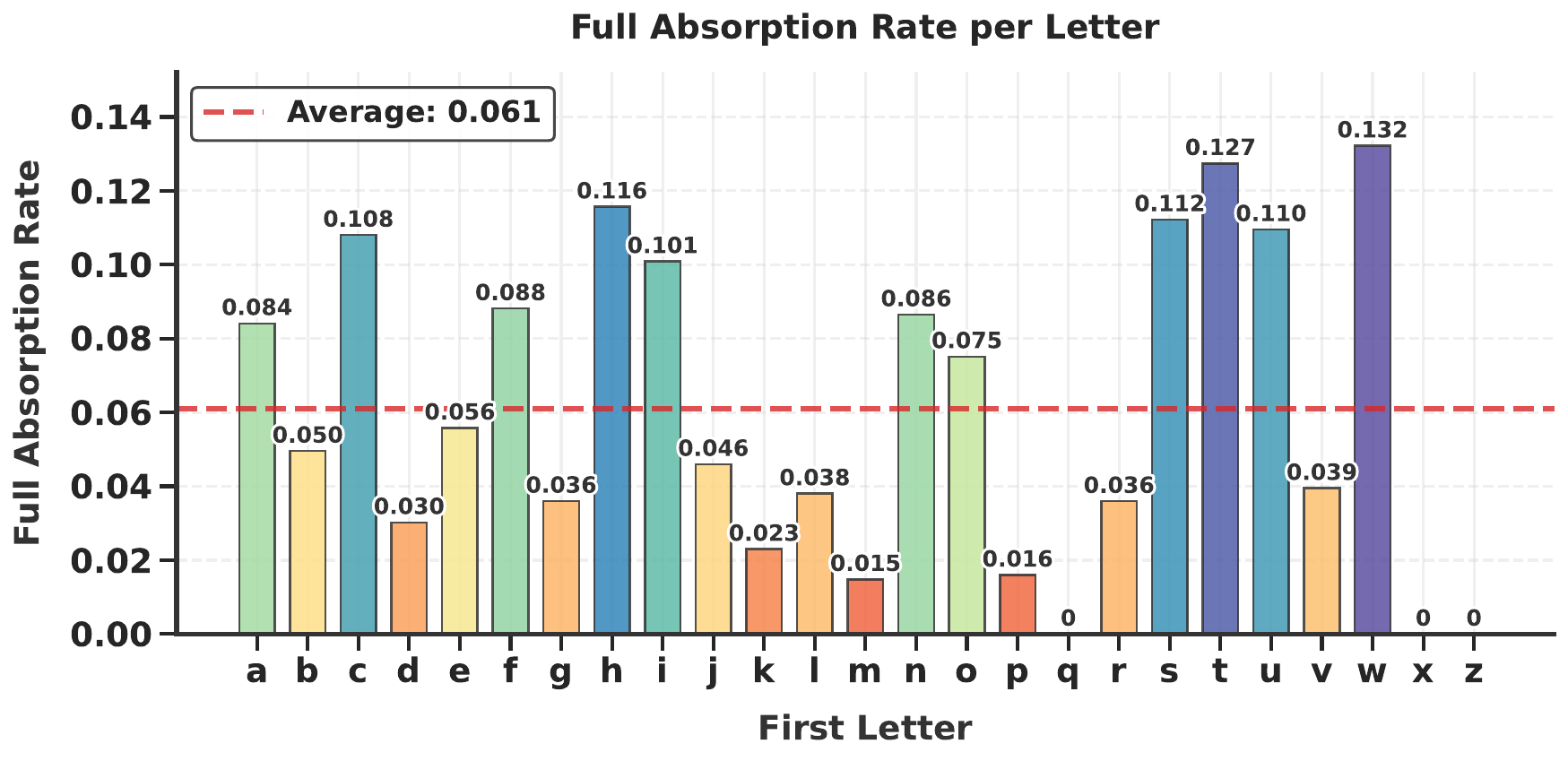}
    \caption{Full Absorption Rate per Letter}
    \label{fig:full_absorption_160}
  \end{subfigure}
  
  \caption{Letter Absorption Results on Pythia-160M}
  \label{fig:absorb_160}
\end{figure*}

\begin{figure*}[t]
  \centering
  \begin{subfigure}[b]{0.48\linewidth}
    \centering
    \includegraphics[width=\linewidth]{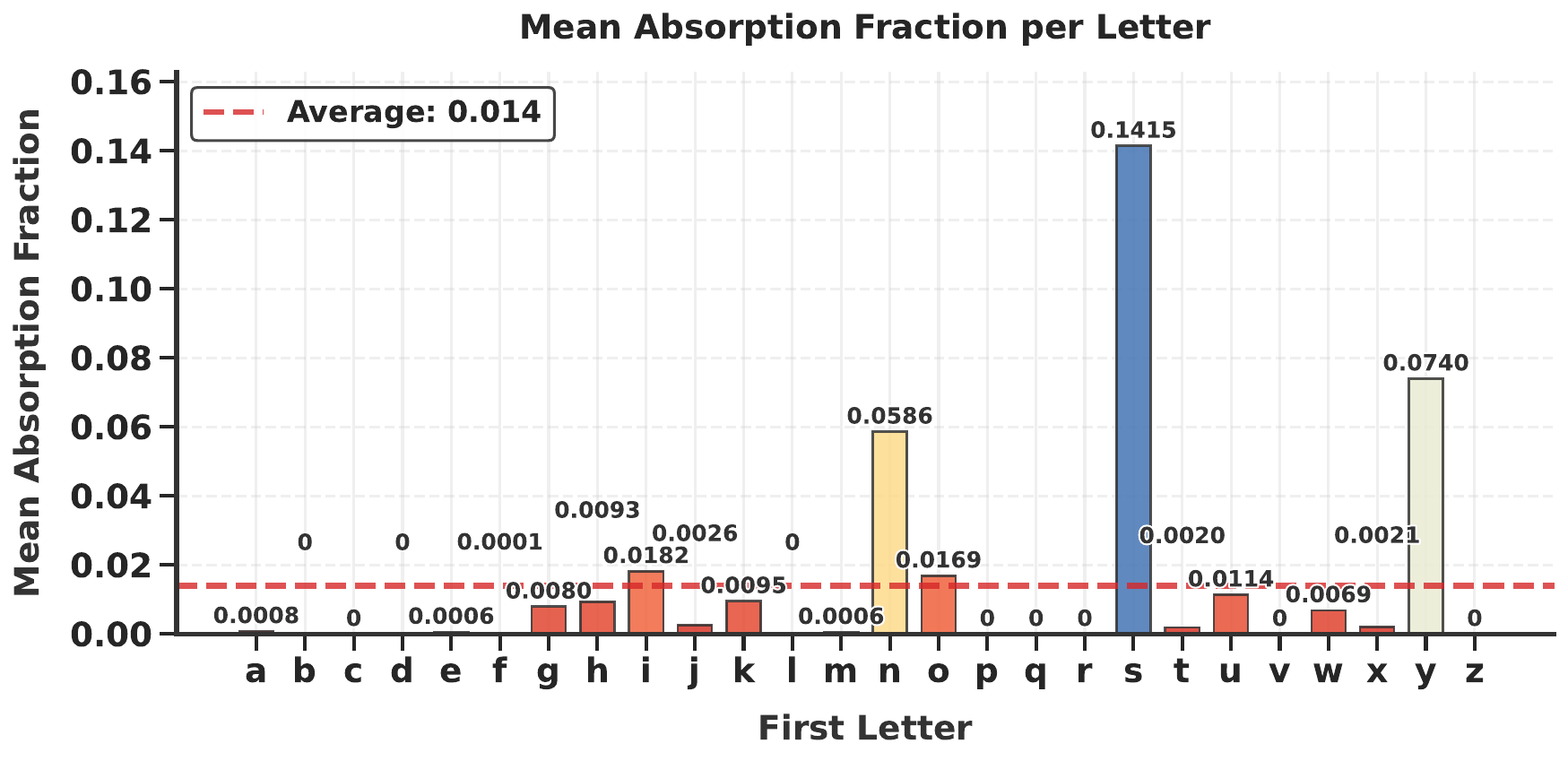}
    \caption{Mean Absorption Fraction per Letter}
    \label{fig:mean_absorption_gemma}
  \end{subfigure}
  \hfill
  \begin{subfigure}[b]{0.48\linewidth}
    \centering
    \includegraphics[width=\linewidth]{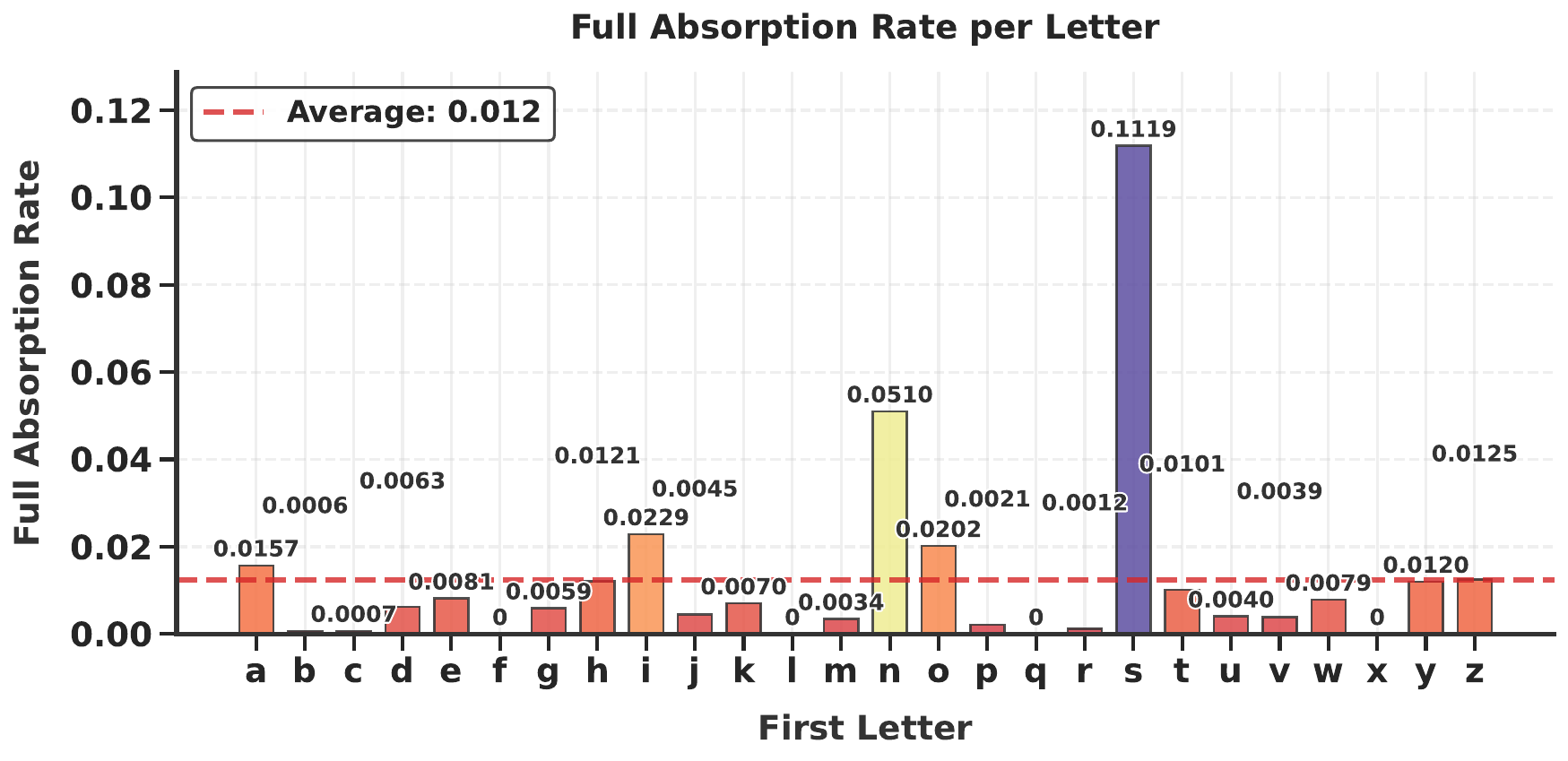}
    \caption{Full Absorption Rate per Letter}
    \label{fig:full_absorption_gemma}
  \end{subfigure}
  
  \caption{Letter Absorption Results on Gemma-2-2B}
  \label{fig:absorb_gemma}
\end{figure*}

\begin{figure*}[t]
  \centering
  \begin{subfigure}[b]{0.32\linewidth}
    \centering
    \includegraphics[width=\linewidth]{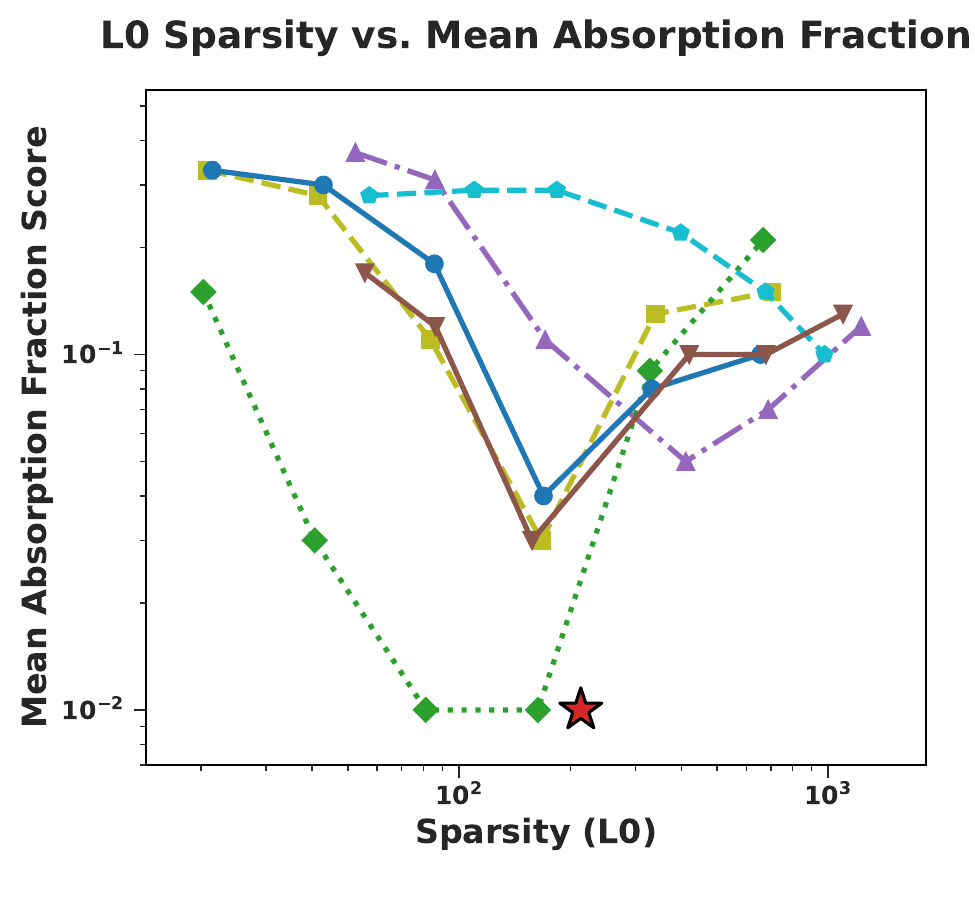}
    \caption{Mean Absorption Fraction}
    \label{fig:mean_absorption}
  \end{subfigure}
  \hfill
  \begin{subfigure}[b]{0.32\linewidth}
    \centering
    \includegraphics[width=\linewidth]{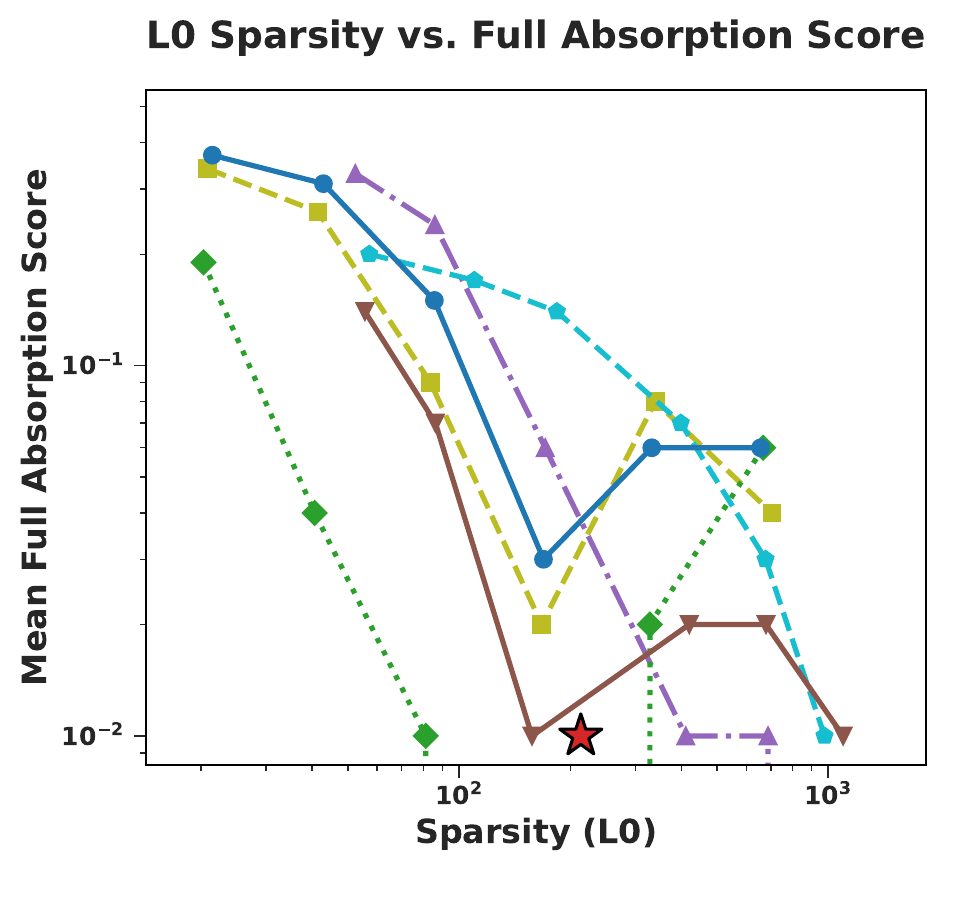}
    \caption{Full Absorption Rate}
    \label{fig:full_absorption}
  \end{subfigure}
  \hfill
  \begin{subfigure}[b]{0.32\linewidth}
    \centering
    \includegraphics[width=\linewidth]{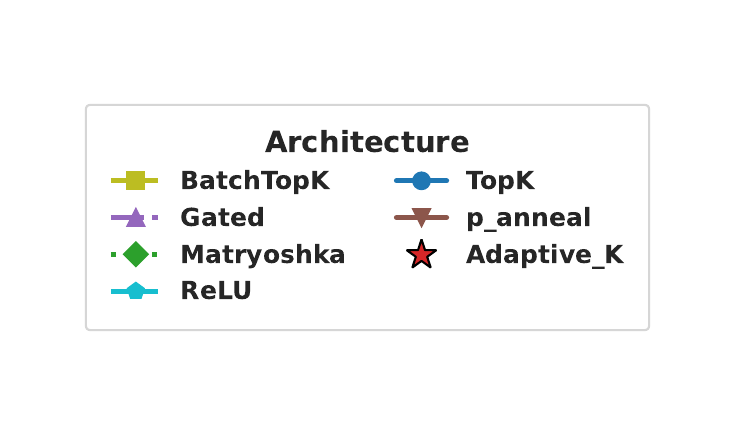}
  \end{subfigure}
  
  \caption{Two Feature Absorption metrics across different SAEs on Gemma-2-2B Layer 12}
  \label{fig:feature absorption}
\end{figure*}

Utilizing this metric, we assessed the performance of our AdaptiveK SAE on layer 3 of Pythia-160M and layer 12 of Gemma-2-2B, as shown in Fig. \ref{fig:absorb_160} and \ref{fig:absorb_gemma}. Notably, when directly benchmarked against SAEs from SAEBench \citep{karvonen2025saebench} for Gemma-2-2B's layer 12, AdaptiveK exhibited superior results on both metrics (Fig. \ref{fig:feature absorption}). This outperformance is particularly significant given that our AdaptiveK was trained on only 250,000 tokens, a dataset 2000 times smaller than the 500,000,000 tokens used for the SAEBench models. In AdaptiveK, concepts are correctly allocated to their intended primary latents rather than being dispersed across unrelated variables. Cases where concepts are entirely misrepresented in non-primary latents are notably rare. This demonstrates AdaptiveK's exceptional effectiveness in maintaining conceptual integrity and preventing feature fragmentation.

\subsubsection{Spurious Correlation Removal (SCR)}

Spurious Correlation Removal evaluates an SAE's ability to disentangle distinct concepts by measuring how effectively it can remove spurious correlations from classifiers. Likewise, utilizing method from SAEBench \citep{karvonen2025saebench}, we first generated biased datasets containing spurious correlations (\emph{e.g.}, professor+male and nurse+female from the Bias in Bios dataset). A linear classifier is then trained on this biased dataset, learning to rely on both the target concept (profession) and the spurious concept (gender). To evaluate an SAE, the method identifies latents most strongly associated with the spurious concept (gender) through probe attribution scores. These identified latents are then zero-ablated, creating a modified classifier. The final SCR score is normalized as:
\begin{equation}
\text{SCR Score}=\frac{A_{\text{abl}} - A_{\text{base}}}{A_{\text{oracle}} - A_{\text{base}}},
\end{equation}
where $A_{\text{abl}}$ is accuracy after ablation, $A_{\text{base}}$ is baseline accuracy, and $A_{\text{oracle}}$ is the accuracy of a classifier trained directly on the desired concept. Higher SCR scores indicate better concept disentanglement, suggesting the SAE effectively isolates distinct concepts into separate latents. 

\begin{figure*}[t]
  \centering
  \begin{subfigure}[b]{0.32\linewidth}
    \centering
    \includegraphics[width=\linewidth]{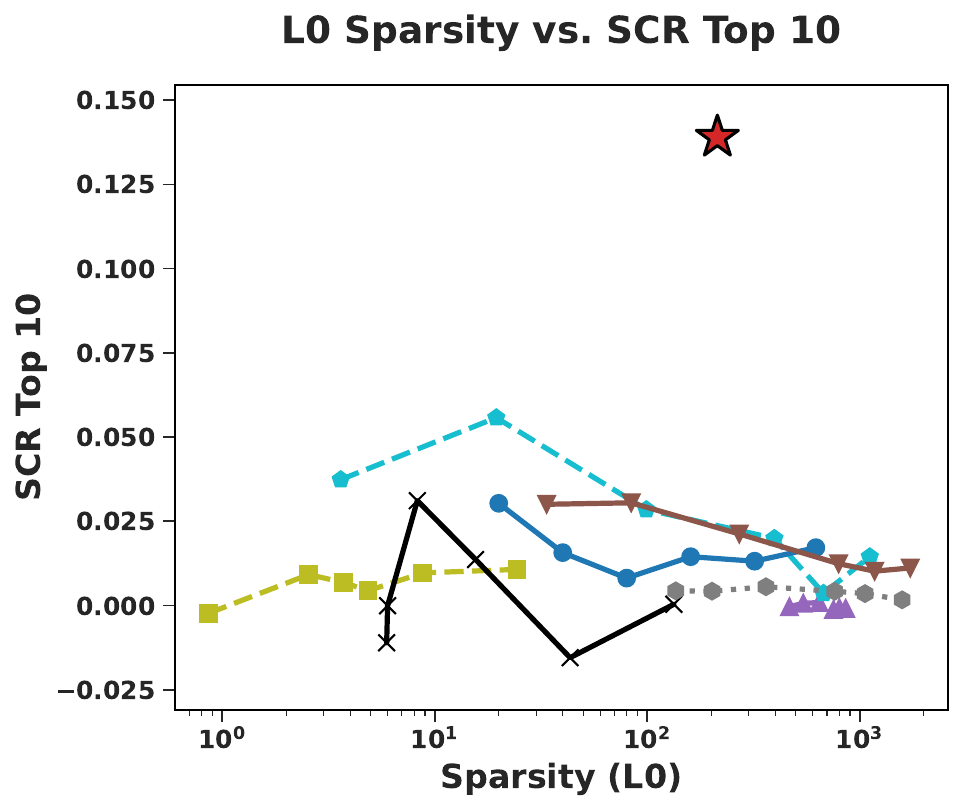}
    \caption{SCR with 10-latent ablation}
    \label{fig:scr_top10}
  \end{subfigure}
  \hfill
  \begin{subfigure}[b]{0.32\linewidth}
    \centering
    \includegraphics[width=\linewidth]{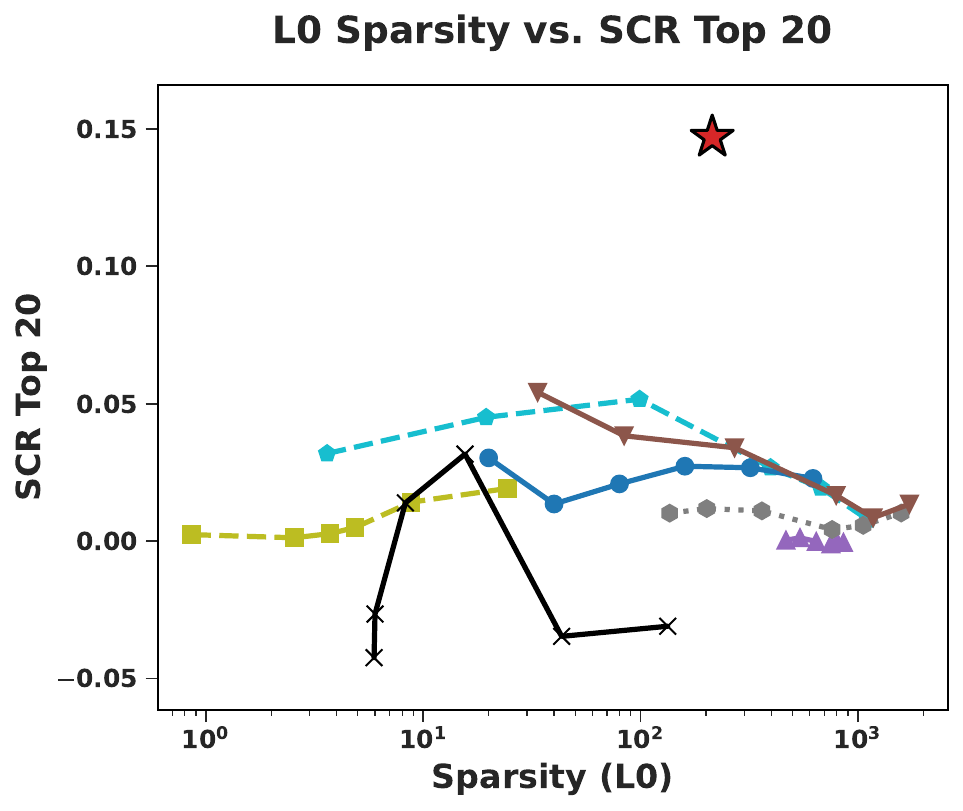}
    \caption{SCR with 20-latent ablation}
    \label{fig:scr_top20}
  \end{subfigure}
  \hfill
  \begin{subfigure}[b]{0.32\linewidth}
    \centering
    \includegraphics[width=\linewidth]{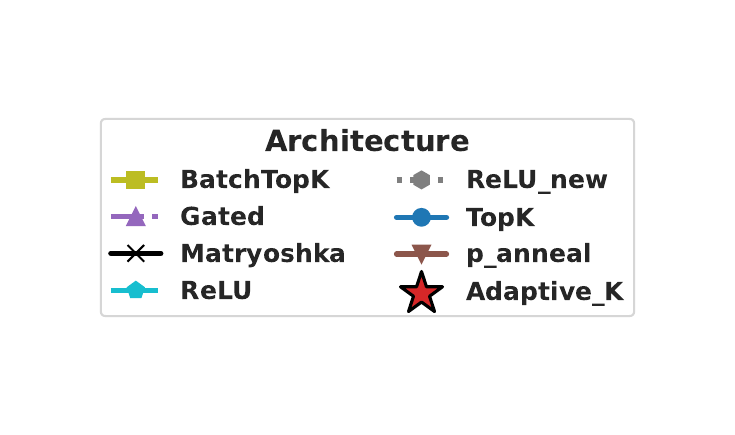}
  \end{subfigure}
  
  \caption{SCR scores across two intervention settings on Pythia-160M Layer 3}
  \label{fig:scr}
\end{figure*}

The results of comparing the AdaptiveK SAE trained on Pythia-160M layer 3 against other baseline SAEs, where all SAEs were trained using 250,000 tokens, are depicted in Fig. \ref{fig:scr}. SCR Top10 and SCR Top20 refer to the SCR scores when ablating the top 10 and top 20 latents respectively that are most associated with the spurious concept. AdaptiveK shows dramatically higher SCR scores in both settings, directly indicating its superior concept disentanglement, with clearer separation between concepts like gender and profession. Additionally, it produces latent representations where concepts are more cleanly isolated in specific latents, enabling more effective debiasing of classifiers.

\subsubsection{Targeted Probe Perturbation (TPP)}
Unlike SCR which works with binary correlated labels, TPP extends SCR to multiclass settings. For each class $i$ in a dataset with $m$ classes, TPP identifies the most relevant latents $L_i$ for that class and trains linear classifiers $C_j$ for each class $j$ with accuracy $A_j$. Then, it creates modified classifiers $C_{i,j}$ by ablating latents $L_i$, with accuracy $A_{i,j}$. We can calculate TPP Score as:
\begin{equation}
\text{TPP Score} =
\text{mean}_{i=j}(A_{i,j} - A_j) - \text{mean}_{i \neq j}(A_{i,j} - A_j).
\end{equation}
This formula captures the difference between within-class effects (when $i=j$) and cross-class effects (when $i \neq j$). A high TPP score indicates good disentanglement, ablating latents for class $i$ primarily affects only class $i$'s accuracy while leaving other class accuracies unchanged. This shows concepts are encoded in separate, non-overlapping latent dimensions. In the same way, TPP Top10 and TPP Top20 refer to the TPP scores when ablating the top 10 and top 20 most relevant latents for each class, respectively. 

\begin{figure*}[t]
  \centering
  \begin{subfigure}[b]{0.32\linewidth}
    \centering
    \includegraphics[width=\linewidth]{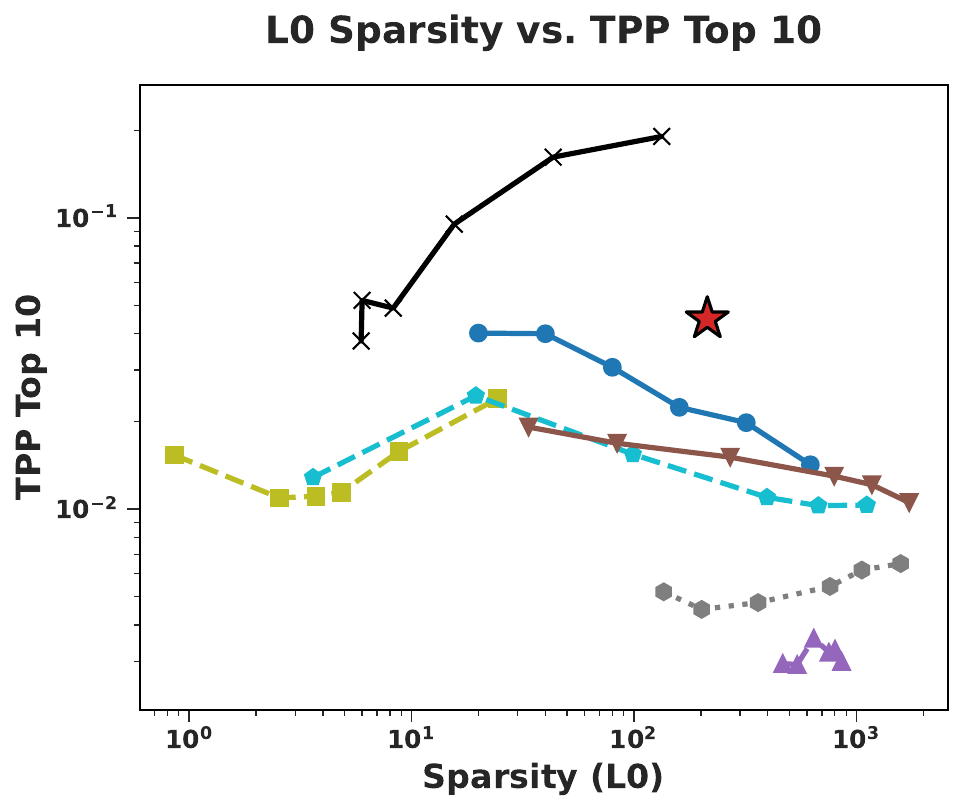}
    \caption{TPP with 10-latent ablation}
    \label{fig:tpp_top10}
  \end{subfigure}
  \hfill
  \begin{subfigure}[b]{0.32\linewidth}
    \centering
    \includegraphics[width=\linewidth]{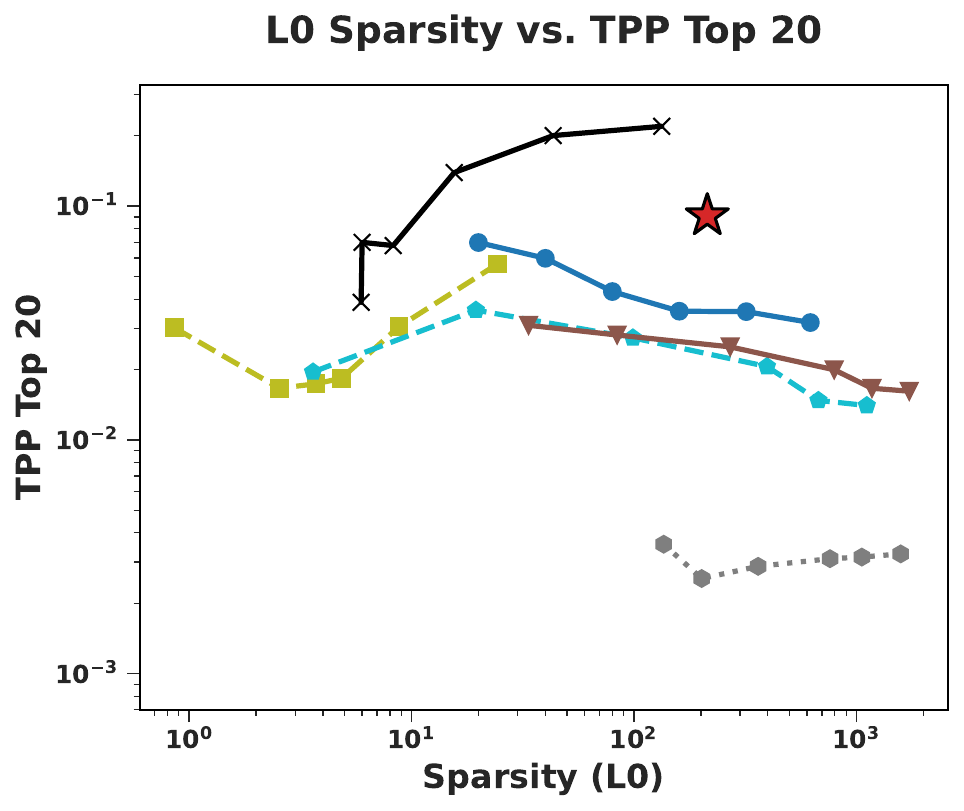}
    \caption{TPP with 20-latent ablation}
    \label{fig:tpp_top20}
  \end{subfigure}
  \hfill
  \begin{subfigure}[b]{0.32\linewidth}
    \centering
    \includegraphics[width=\linewidth]{figs/scrtpp_legend.pdf}
  \end{subfigure}
  
  \caption{TPP scores across two intervention settings on Pythia-160M Layer 3}
  \label{fig:tpp}
\end{figure*}

As shown in Fig. \ref{fig:tpp}, AdaptiveK outperforms most SAEs, showing that ablating latents identified for one class primarily affects only that class's probe accuracy. This precise targeting indicates AdaptiveK organizes its latent space with clearer conceptual boundaries. Combined with its SCR results in Fig. \ref{fig:scr}, AdaptiveK creates a more structurally organized latent space with minimal concept overlap. While Matryoshka Batch TopK performs well on TPP, AdaptiveK's consistent high performance across both TPP and SCR metrics demonstrates its representation efficiently separates both binary concepts and multiclass classes. This dual strength in disentanglement makes it particularly suited for interpretability tasks requiring precise concept isolation and targeted.

\subsubsection{Resolving Attribute-Value Entanglements in Language Models (RAVEL)}
RAVEL directly  measures a key application of interpretability, which is the practical utility of an SAE for targeted knowledge editing. It tests whether interventions on specific latents can modify one attribute of an entity while preserving other attributes. The evaluation focuses on whether an SAE can help a language model make targeted factual modifications, for example, changing Paris's country from France to Japan while correctly maintaining that the language spoken there is still French (rather than incorrectly switching to Japanese). The evaluation begins by collecting high-confidence entity-attribute predictions across five diverse categories (cities, Nobel laureates, physical objects, etc.). For each entity, RAVEL identifies which latents most strongly encode specific attributes using trained probes. It then tests what happens when these latents are manipulated - can the model be made to believe Paris is in Japan while still knowing French is spoken there? This capability is measured through two complementary metrics: the cause score (how effectively the intervention changes the target attribute) and the isolation score (how well other attributes remain unaffected). These are averaged into a final disentanglement score. Higher scores indicate better attribute separation in the SAE's latent space, showing it has successfully disentangled different factual properties into distinct latent dimensions that can be independently manipulated.

\begin{figure*}[t]
  \centering
  \begin{subfigure}[b]{0.32\linewidth}
    \centering
    \includegraphics[width=\linewidth]{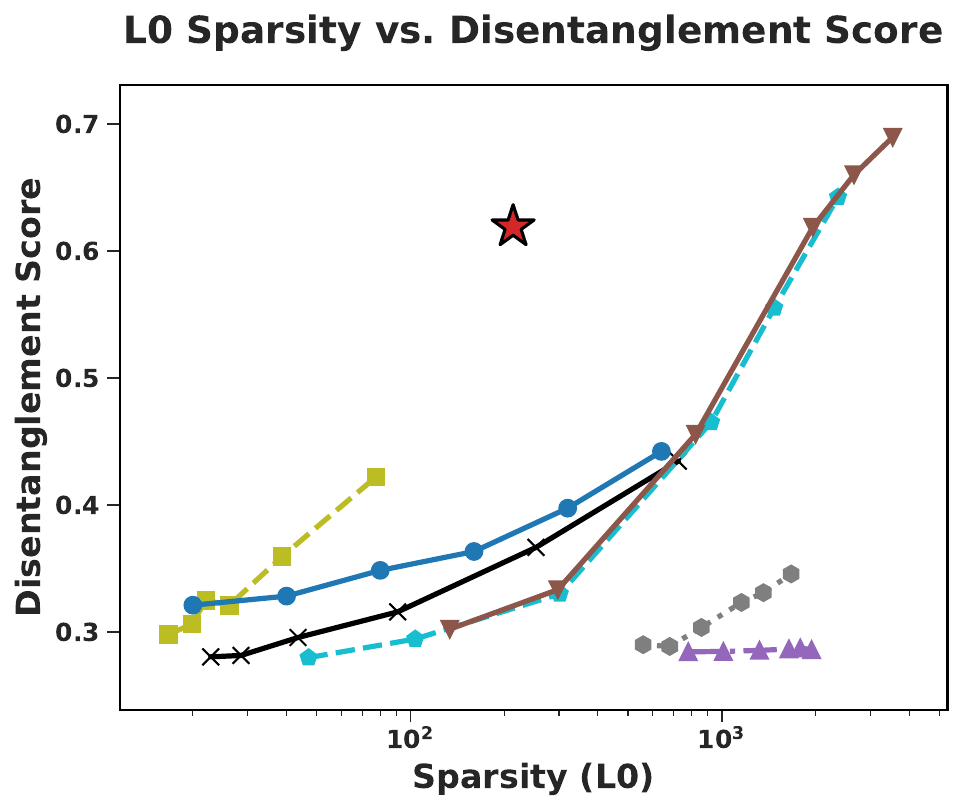}
    \caption{Disentanglement Score}
    \label{fig:disentanglement}
  \end{subfigure}
  \hfill
  \begin{subfigure}[b]{0.32\linewidth}
    \centering
    \includegraphics[width=\linewidth]{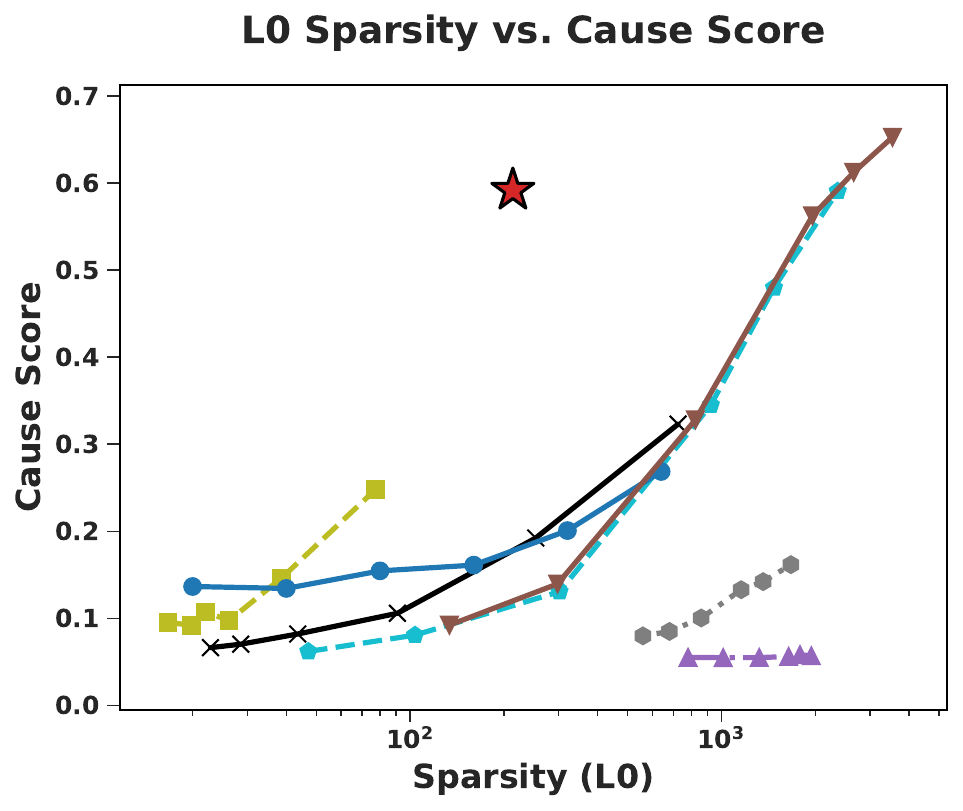}
    \caption{Cause Score}
    \label{fig:cause}
  \end{subfigure}
  \hfill
  \begin{subfigure}[b]{0.32\linewidth}
    \centering
    \includegraphics[width=\linewidth]{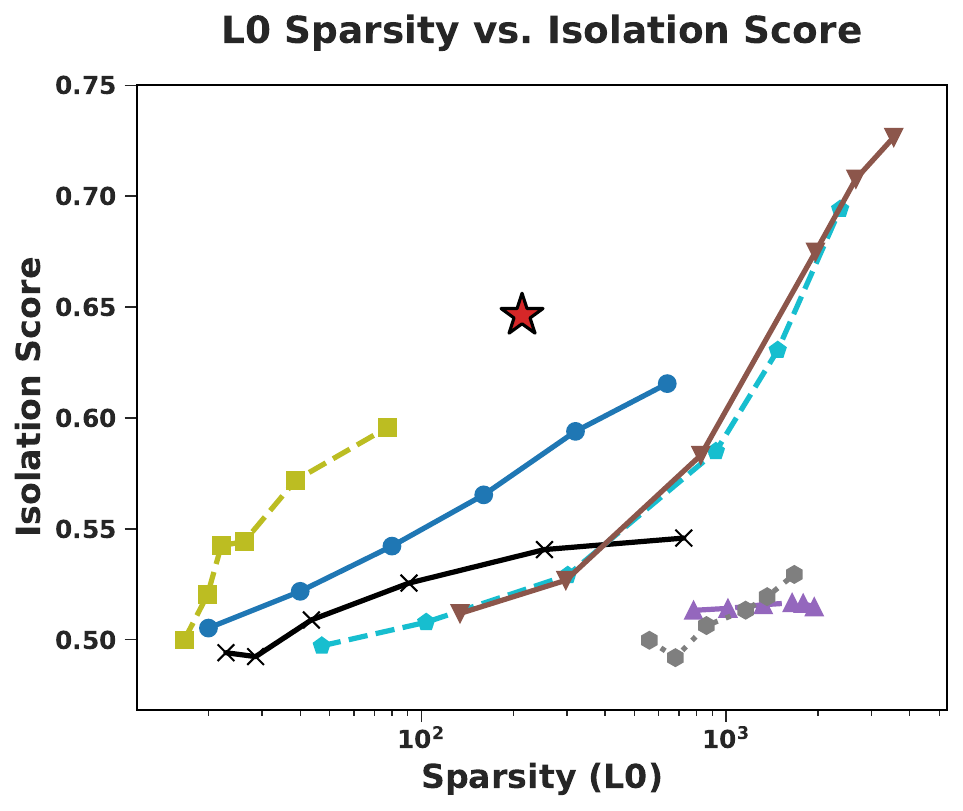}
    \caption{Isolation Score}
    \label{fig:isolation}   
  \end{subfigure}
  
  \includegraphics[width=0.95\textwidth]{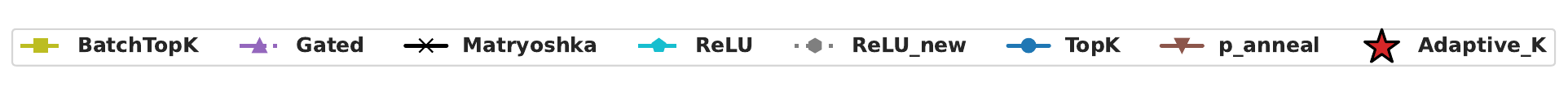}
  
  \caption{Three RAVEL results on Gemma-2-2B Layer 12}
  \label{fig:RAVEL}
\end{figure*}

In the disentanglement score (Fig. \ref{fig:disentanglement}), AdaptiveK achieves 0.62, substantially higher than contemporary architectures like TopK and Matryoshka at comparable sparsity levels. For the cause score (Fig. \ref{fig:cause}), AdaptiveK reaches 0.6, roughly double the effectiveness of other SAEs at similar sparsity. This indicates AdaptiveK is exceptionally good at identifying and modifying the specific latents that control target attributes (like a city's country). The isolation score (Fig. \ref{fig:isolation}) shows AdaptiveK at 0.65, demonstrating it maintains unrelated attributes more effectively than others. While P Anneal and some other SAEs eventually reach similar or higher disentanglement scores, they require much higher sparsity levels (L0 $>$ 1000) to do so, making them less practical for interpretability work that benefits from more compact representations. Combined with the earlier SCR and TPP results, these RAVEL findings confirm that AdaptiveK creates a latent space with exceptionally clean separation between different concepts and attributes, enabling more precise and controlled interventions on language model knowledge.

\subsubsection{Sparse Probing}
Unlike other metrics that focus on concept separation, Sparse Probing evaluates an SAE's ability to organize meaningful semantic features by measuring how effectively it concentrates concept-specific information in individual latents. The method applies the SAE to encode texts from five diverse datasets covering profession classification (Bias in Bios), product categorization and sentiment analysis (Amazon Reviews), language identification (Europarl), programming language detection (GitHub), and news topic categorization (AG News). For each of the 35 binary classification tasks, the evaluation first identifies which latents show the greatest activation difference between positive and negative examples. A logistic regression probe is then trained using only these selected latents (ranging from just the single most relevant latent to the top 5), with performance measured on 1,000 held-out test examples. The key insight is that if the SAE has effectively organized information, even a small number of latents (the top-K) should contain sufficient information to perform specific classification tasks. 

\begin{figure*}[t]
  \centering
  \begin{subfigure}[b]{0.328\linewidth}
    \centering
    \includegraphics[width=\linewidth]{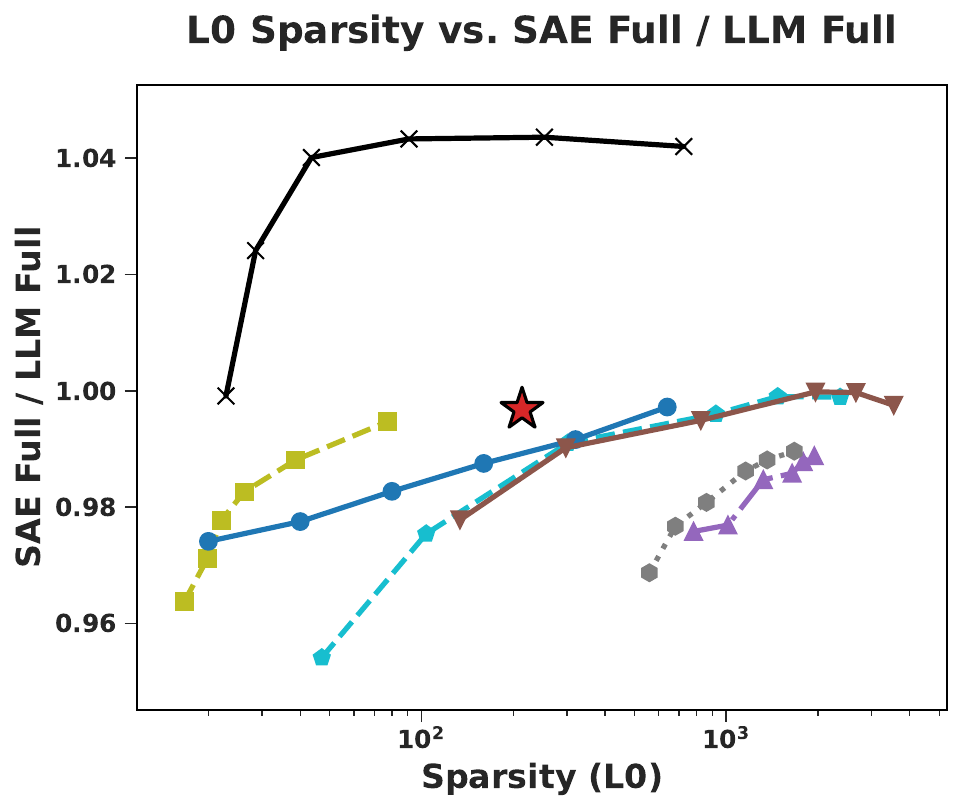}
    \label{fig:sae_full_llm_full}
  \end{subfigure}
  \hspace{0.02\linewidth}
  \begin{subfigure}[b]{0.328\linewidth}
    \centering
    \includegraphics[width=\linewidth]{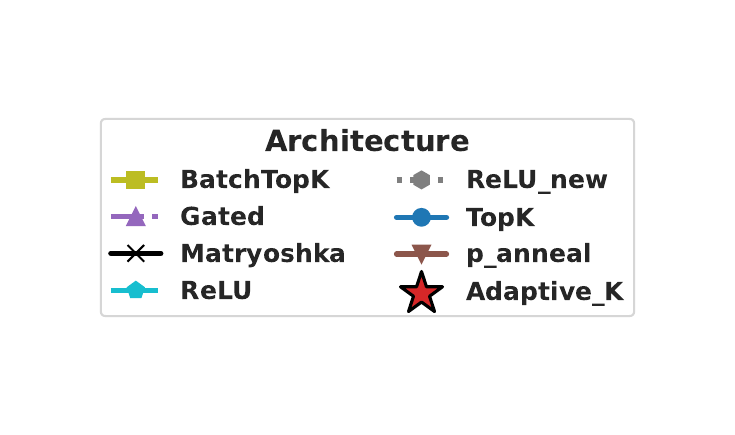}
    \label{fig:sparse_legend}
  \end{subfigure}
  
  \caption{SAE Full Accuracy / LLM Full Accuracy on Gemma-2-2B Layer 12}
  \label{fig:sae_full_llm_full_2}
\end{figure*}

Metric ``Full Activations Accuracy'' represents classification performance when using all SAE activations, establishing an upper bound. Metrics ``Top-K Accuracy'' (where K is 1, 2, or 5) measure performance when restricting the probe to only the K most relevant latents for each task. First, we measure information retention rate (SAE Full Activations Accuracy / LLM Full Activations Accuracy), which quantifies how well each SAE preserves the original model's information when using all reconstructed activations. As shown in Fig. \ref{fig:sae_full_llm_full_2}, AdaptiveK's retention ratio of approximately 0.997 demonstrates near-perfect preservation of the original model's information. While Matryoshka's slightly higher ratio ($>$1) suggests beneficial feature reorganization or denoising during reconstruction, such enhancement represents a supplementary advantage rather than a necessity.

\begin{figure*}[t]
  \centering
  \begin{subfigure}[b]{0.32\linewidth}
    \centering
    \includegraphics[width=\linewidth]{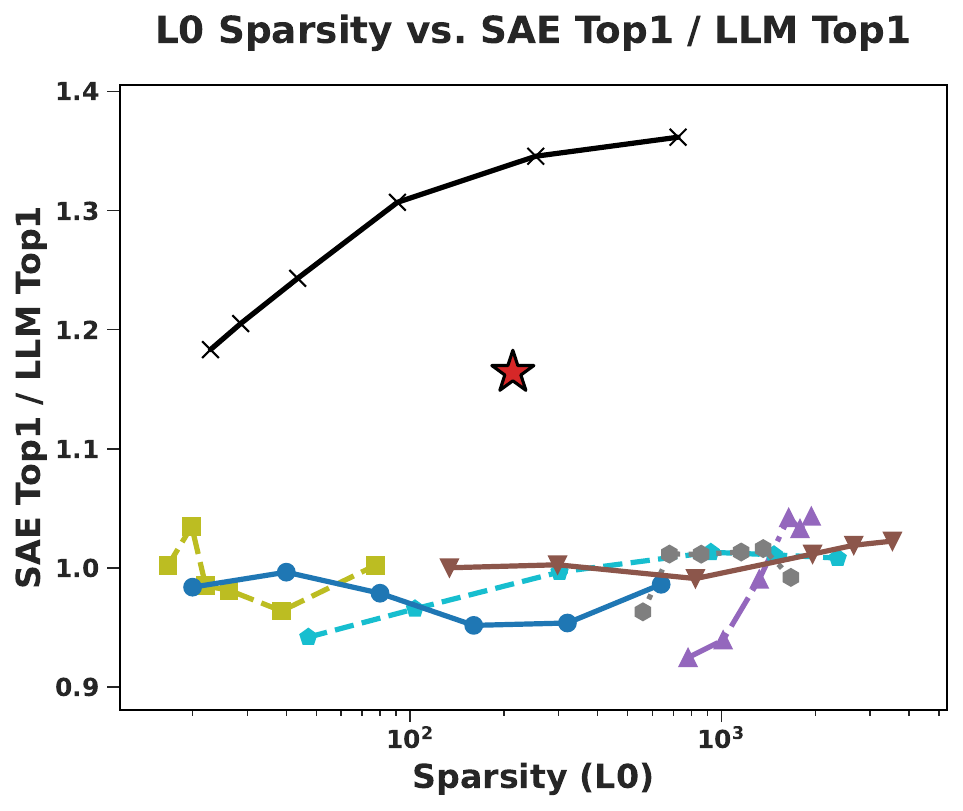}
    \caption{SAE Top-1 / LLM Top-1}
    \label{fig:sae_top1_llm_top1}
  \end{subfigure}
  \hfill
  \begin{subfigure}[b]{0.32\linewidth}
    \centering
    \includegraphics[width=\linewidth]{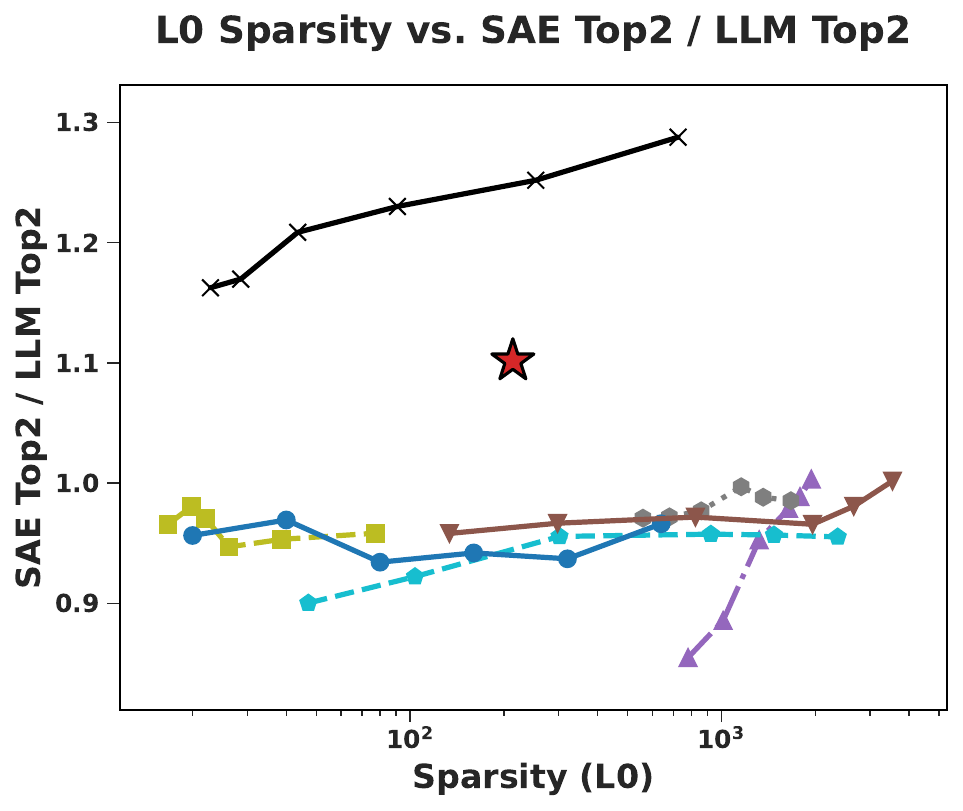}
    \caption{SAE Top-2 / LLM Top-2}
    \label{fig:sae_top2_llm_top2}
  \end{subfigure}
  \hfill
  \begin{subfigure}[b]{0.32\linewidth}
    \centering
    \includegraphics[width=\linewidth]{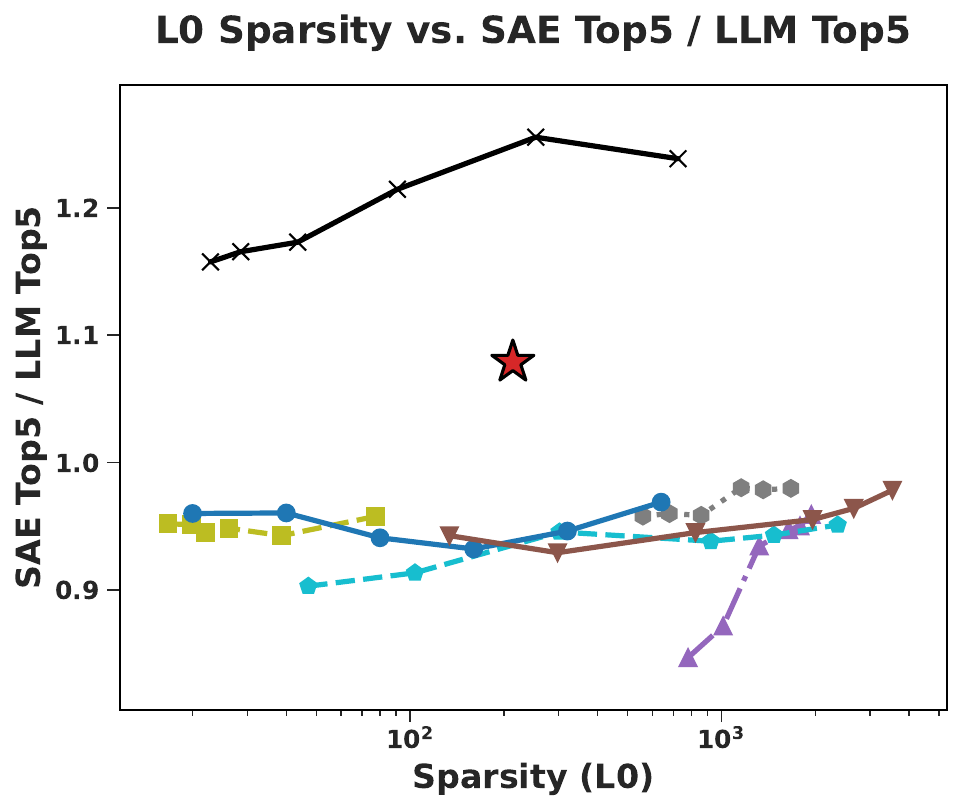}
    \caption{SAE Top-5 / LLM Top-5}
    \label{fig:sae_top5_llm_top5}   
  \end{subfigure}
  
  \caption{SAE Top-K / LLM Top-K on Gemma-2-2B Layer 12}
  \label{fig:sae_topk_llm_topk}
\end{figure*}

Second, we examine relative feature concentration (SAE Top-K / LLM Top-K) across three granularities (K=1,2,5) as illustrated in Fig. \ref{fig:sae_topk_llm_topk}. These metrics reveal how efficiently each architecture concentrates concept-specific information in its most relevant latents compared to the base model. The SAE Top-K/LLM Top-K ratios for AdaptiveK consistently exceed 1. Though marginally below Matryoshka, these values convincingly demonstrate AdaptiveK's superior ability to concentrate concept-relevant information in fewer latent dimensions than the original model requires, indicating more efficient latent space organization.

\begin{figure*}[t]
  \centering
  \begin{subfigure}[b]{0.32\linewidth}
    \centering
    \includegraphics[width=\linewidth]{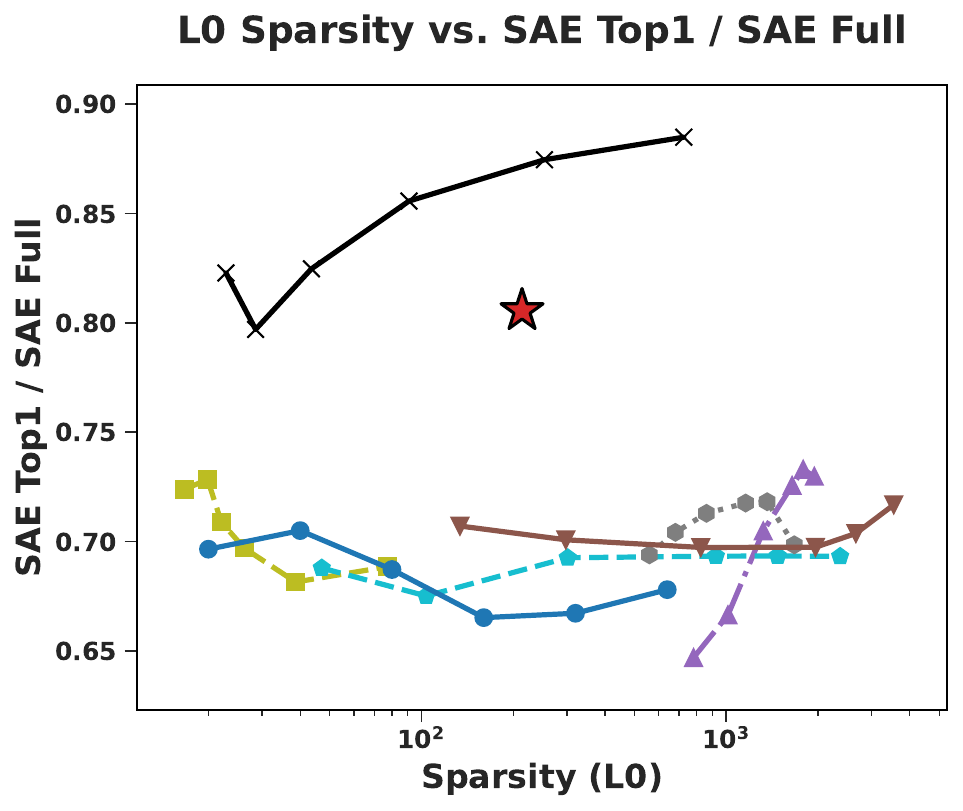}
    \caption{SAE Top-1 / SAE Full}
    \label{fig:sae_top1_sae_full}
  \end{subfigure}
  \hfill
  \begin{subfigure}[b]{0.32\linewidth}
    \centering
    \includegraphics[width=\linewidth]{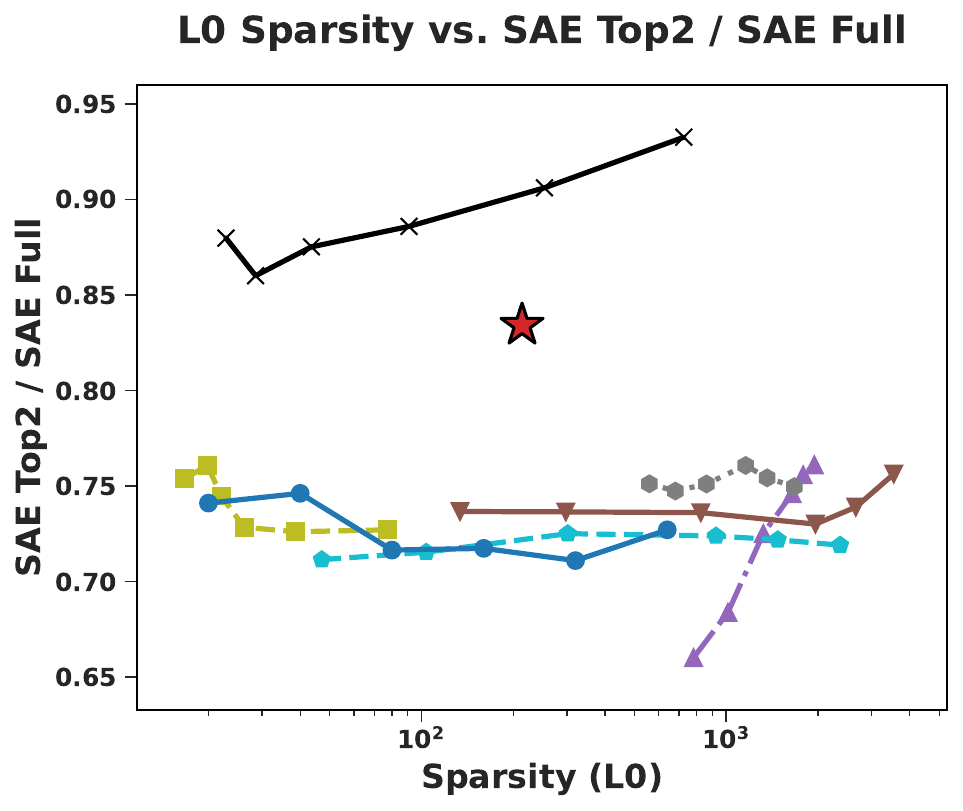}
    \caption{SAE Top-2 / SAE Full}
    \label{fig:sae_top2_sae_full}
  \end{subfigure}
  \hfill
  \begin{subfigure}[b]{0.32\linewidth}
    \centering
    \includegraphics[width=\linewidth]{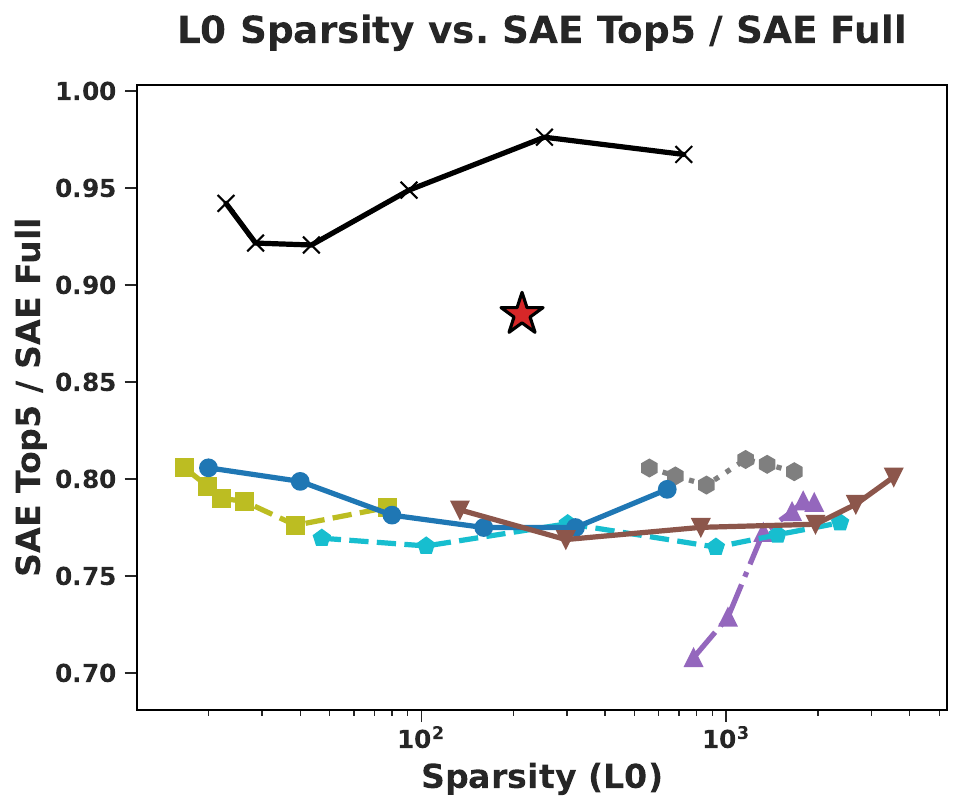}
    \caption{SAE Top-5 / SAE Full}
    \label{fig:sae_top5_sae_full}   
  \end{subfigure}
  
  \caption{SAE Top-K / SAE Full on Gemma-2-2B Layer 12}
  \label{fig:sae_topk_sae_full}
\end{figure*}

Finally, we assess information concentration efficiency (SAE Top-K / SAE Full) in Fig. \ref{fig:sae_topk_sae_full}, which measures how much of an SAE's total information is captured in its K most relevant latents. AdaptiveK captures approximately 82\% of its complete representation using just its most relevant single latent variable, surpassing most other SAE architectures and demonstrating exceptional information compression.

Collectively, these metrics establish AdaptiveK's balanced excellence: it maintains original model information (retention rate), concentrates more concept-relevant information in fewer dimensions than the original model (relative feature concentration), and achieves a highly organized internal representation that localizes most information in a minimal number of latents (information concentration efficiency).

\subsection{Hyperparameters Analysis}
\label{hyperparameters}

\subsubsection{Regularization strength $\lambda$}
As stated in Section \ref{Training Settings}, to determine the optimal regularization strength $\lambda$, we perform 5-fold cross-validation. For each $\lambda$ in the set $\{0.001,\,0.01,\,0.1,\,1.0,\,10.0,\,100.0,\,1000.0\}$, the probe is trained on four folds and evaluated on the remaining fold using root mean squared error (RMSE). Tab. \ref{tab:hyper_lambda} reports the average RMSE across folds for each $\lambda$ on selected models, with the best $\lambda$ highlighted in bold.

\subsubsection{Steepness of sigmoid function $s$}
$s$ determines how complexity scores map to k-values. Our experiments on Gemma-2-2B 
(Tab. \ref{tab:hyper_s}) show that $s$ exhibits remarkable robustness across the tested range of 2.0 to 12.0. As $s$ increases, the k-value distribution becomes more dynamic: lower values ($s$=2.0) produce conservative allocation with a narrow range (min k=143, max k=225), while higher values ($s$=12.0) enable more aggressive feature allocation with expanded ranges (min k=58, max k=315). Reconstruction quality remains stable across all tested values. These results indicate that our method is highly robust to $s$ parameter selection, with performance variations of less than 1\% across the entire range, validating our choice of $s$=6.0 as a balanced default configuration.

\begin{table*}[t]
  \centering
  \caption{Effect of $s$ on $k$ statistics and performance metrics}
  \label{tab:hyper_s}
  \begin{tabular}{ccccccc}
    \toprule
    $k$ on test set & min $k$ & max $k$ & avg $k$ & Explained Variance & Cosine Similarity & L2 Ratio \\
    \midrule
    $s=2.0$  & 143 & 225 & 188 & 0.738 & 0.908 & 0.911 \\
    $s=4.0$  & 118 & 267 & 203 & 0.741 & 0.909 & 0.914 \\
    \textbf{$s=6.0$}  & \textbf{96} & \textbf{291} & \textbf{214} & \textbf{0.743} & \textbf{0.909} & \textbf{0.921} \\
    $s=8.0$  & 80  & 304 & 222 & 0.743 & 0.909 & 0.916 \\
    $s=12.0$ & 58  & 315 & 232 & 0.742 & 0.909 & 0.917 \\
    \bottomrule
  \end{tabular}
\end{table*}

\subsubsection{Probe weight $\gamma$}
$\gamma$ balances the SAE reconstruction loss and probe loss during joint fine-tuning, where smaller values prioritize reconstruction quality and produce more conservative k-value allocation, while larger $\gamma$ values enhance probe dominance and expand the k-value range to better reflect complexity differences. Our experiments (Tab. \ref{tab:hyper_gamma}) confirm this expected behavior, with k-value ranges expanding as $\gamma$ increases (from min k=98, max k=285 at $\gamma$=0.5 to min k=92, max k=293 at $\gamma$=1.0), but across $\gamma \in[0.5,1.0]$ show minimal performance variation. This stability demonstrates that our method is robust to $\gamma$ selection, with performance variations under 1\%, validating our default choice of $\gamma$=0.9.

\begin{table*}[t]
  \centering
  \caption{Effect of $\gamma$ on $k$ statistics and performance metrics}
  \label{tab:hyper_gamma}
  \begin{tabular}{ccccccc}
    \toprule
    $k$ on test set & min $k$ & max $k$ & avg $k$ & Explained Variance & Cosine Similarity & L2 Ratio \\
    \midrule
    $\gamma=0.5$ & 98 & 285 & 208 & 0.742 & 0.909 & 0.913 \\
    $\gamma=0.7$ & 97 & 289 & 210 & 0.743 & 0.909 & 0.916 \\
    \textbf{$\gamma=0.9$} & \textbf{96} & \textbf{291} & \textbf{214} & \textbf{0.743} & \textbf{0.909} & \textbf{0.921} \\
    $\gamma=1.0$ & 92 & 293 & 215 & 0.743 & 0.909 & 0.919 \\
    \bottomrule
  \end{tabular}
\end{table*}

\subsubsection{Deviation penalty $\delta$}
$\delta$ prevents probe parameters from deviating too far from their pre-trained values. Starting with $\delta$=0.2, we dynamically adjust this weight during training by monitoring probe loss changes over the recent 3 steps. We calculate the loss change rate as $\frac{\text{earliest loss} - \text{latest loss}}{\text{earliest loss}}$ and adjust $\delta$ accordingly: when loss decreases rapidly (change rate $>$ 0.05), indicating good probe learning progress in complexity prediction, we reduce the deviation constraint by setting $\delta = \text{current}_{\delta} \times 0.8 \quad (\text{minimum } 0.01)$ to allow more parameter flexibility; when loss stagnates or increases, suggesting learning difficulties or overfitting, we strengthen the constraint by setting $\delta = \text{current}_{\delta} \times 1.2 \quad (\text{maximum } 0.5)$ to prevent excessive parameter drift. The upper bound of 0.5 is critical because at this value, the deviation loss becomes comparable in magnitude to the probe loss, and higher values would make the deviation penalty too dominant, completely preventing the probe from adapting to new patterns when learning becomes difficult. Additionally, since deviation loss measures distance from initial values, excessive weights would create large gradients that harm training stability.

\subsubsection{Sigmoid-based Transformation}
The sigmoid function provides a smooth non-linear mapping that aligns well with empirical observations about complexity–feature relationships. Our implementation uses only two intuitive parameters: mid\_point, which specifies the complexity level corresponding to the base $k$, and steepness, which controls the transition smoothness. In practice, we found these parameters to be stable across models. Setting mid\_point=0.5 centers the allocation around normalized complexity, while steepness=6.0 ensures appropriate transition smoothness. To test robustness, we also experimented with Linear Mapping and Exponential Mapping, with results on Gemma-2-2B reported in Tab. \ref{tab:mapping_methods}. All three mapping methods achieved similar performance across metrics, demonstrating that our approach is not sensitive to the specific choice of mapping.

\begin{table*}[t]
  \centering
  \caption{Comparison of mapping methods}
  \label{tab:mapping_methods}
  \begin{tabular}{lccccccc}
    \toprule
    Method & avg $k$ & min $k$ & max $k$ & Explained Var & Cosine Sim & L2 Ratio & Rel Recon Bias \\
    \midrule
    Sigmoid & 214 & 95  & 291 & 0.80 & 0.93 & 0.93 & 0.9995 \\
    Linear        & 206 & 98  & 296 & 0.79 & 0.93 & 0.92 & 0.9988 \\
    Exponential   & 144 & 53  & 271 & 0.78 & 0.92 & 0.92 & 0.9949 \\
    \bottomrule
  \end{tabular}
\end{table*}

\subsection{SAE Performance with Larger k Value}
Additional experiments have been conducted with different $k_{min}$ and $k_{max}$. As shown in Tab. \ref{tab:k_range}, when scaling $k_{max}$ from 320 to 640, we observe steady improvements: Explained Variance increases from 0.743 to 0.789, Cosine Similarity from 0.909 to 0.926, and L2 Ratio from 0.921 to 0.935. These results indicate that AdaptiveK benefits from increased capacity similarly to standard SAEs.

\begin{table*}[t]
  \centering
  \caption{Performance with different $k_{\min}$ and $k_{\max}$ settings}
  \label{tab:k_range}
  \begin{tabular}{lcccccc}
    \toprule
    $k$ on test set & min $k$ & max $k$ & avg $k$ & Explained Var & Cosine Sim & L2 Ratio \\
    \midrule
    $k_{\min}=20, \; k_{\max}=320$ & 96  & 291 & 214 & 0.743 & 0.909 & 0.921 \\
    $k_{\min}=20, \; k_{\max}=480$ & 132 & 435 & 313 & 0.768 & 0.919 & 0.926 \\
    $k_{\min}=20, \; k_{\max}=640$ & 170 & 579 & 415 & 0.789 & 0.926 & 0.935 \\
    \bottomrule
  \end{tabular}
\end{table*}

\section{Training Efficiency Analysis}
\label{training efficiency}
In large-scale or token-level training scenarios, complexity annotation is often regarded as a scalability challenge, since exhaustive annotation may become costly. AdaptiveK addresses this issue through several design considerations. 

Firstly, annotation is a separate process. Complexity annotation occurs during the data preprocessing stage as targets for complexity prediction, independent of SAE's three-stage training. Once completed, the annotations can be used to train linear probes for any model architecture. Secondly, for large-scale training data, only a subset is annotated for linear probe training while using the full dataset for SAE training.

Training efficiency is further assessed through comparisons across multiple SAE configurations. While traditional SAEs avoid complexity annotation, they require training multiple SAEs with different sparsity settings (different k values or sparsity penalties). Tab. \ref{tab:total_time} shows complete training times for other SAEs with single sparsity configurations on Gemma-2-2B, all exceeding AdaptiveK's training time. For six different sparsity levels (\emph{e.g.}, k=20,40,80,160,320,640), the total training time would exceed AdaptiveK by more than 6-fold.

\begin{table*}[t]
  \centering
  \caption{Total training time (minutes) for different SAEs}
  \label{tab:total_time}
  \small
  \begin{tabular}{lcccccccc}
    \toprule
    & AdaptiveK & Batch TopK & Gated & Matryoshka & P Anneal & Relu & Relu New & TopK \\
    \midrule
    Total (min) & 11084 & 13853 & 13908 & 13773 & 13902 & 13835 & 13814 & 13955 \\
    \bottomrule
  \end{tabular}
\end{table*}

\section{Adaptability to Token-Level Evaluation}
\label{token-level evaluation}
Although AdaptiveK operates at the context level by default, its design also enables reliable token-level evaluation. The training process uses contexts of 1024 tokens each (as mentioned in Sec. \ref{Training Settings}). These contexts contain multiple sentences of varying lengths, including truncated incomplete sentence. The complexity of the last token in each context is used to represent the entire context, ensuring generalization capability. This deliberate approach of not using complete sentences enables training with the last token regardless of sentence length (whether 32 tokens, 256 tokens, etc.).

Due to training process, during evaluation, even when the input context is not 1024 tokens (for example, the first 256 or 500 tokens), our method can still effectively compute complexity based on the representation of the last token in the first n tokens. This is the reason and mechanism for why and how it can effectively predict the complexity of tokens at arbitrary positions.

Empirical results on Gemma-2-9B further confirm this adaptability. As shown in Tab. \ref{tab:token_positions} reconstruction performance is consistent across token positions 10, 100, 200, 500, 800, and 1000. The cosine similarity varies by only 0.3\% (0.954-0.957), indicating highly consistent reconstruction directions. L2 ratios and reconstruction bias remain close to 1.0 (0.9583-0.9701, 1.0052-1.0208), demonstrating accurate reconstruction magnitudes. All metrics vary minimally ($<$ 5\%) across positions, proving that AdaptiveK SAE maintains stable reconstruction quality at any context position. Our context-level approach reflects efficiency considerations rather than methodological limitations.

\begin{table*}[t]
  \centering
  \caption{Evaluation across different token positions}
  \label{tab:token_positions}
  \begin{tabular}{ccccccc}
    \toprule
    Token Position & Cosine Similarity & Explained Variance & L2 Ratio & Recon Bias & K Value \\
    \midrule
    10    & 0.9544 & 0.7865 & 0.9701 & 1.0208 & 267 \\
    100   & 0.9570 & 0.8255 & 0.9688 & 1.0130 & 282 \\
    200   & 0.9557 & 0.8268 & 0.9622 & 1.0104 & 285 \\
    500   & 0.9544 & 0.8203 & 0.9674 & 1.0130 & 285 \\
    800   & 0.9557 & 0.8333 & 0.9622 & 1.0104 & 286 \\
    1000  & 0.9557 & 0.8151 & 0.9583 & 1.0052 & 290 \\
    \bottomrule
  \end{tabular}
\end{table*}

\section{Broader Impacts}

Our AdaptiveK Sparse Autoencoder offers significant broader impacts across multiple domains. By dynamically allocating representational capacity based on input complexity, it enhances computational efficiency through optimized resource utilization, potentially reducing energy consumption in large-scale AI systems. This adaptive approach simultaneously improves model interpretability by establishing clear correlations between complexity metrics and feature activation patterns, providing researchers with new insights into representation learning mechanisms.

\clearpage

\begin{figure*}[t]
\begin{scriptsize}
\begin{tcolorbox}[title=Prompt for Scoring Context Complexity]

\textbf{Detailed Evaluation Dimensions}

\begin{enumerate}[leftmargin=*, itemsep=1pt, parsep=1pt, topsep=2pt]

\item Lexical Complexity (\emph{Weight: 20\%}): Evaluate the vocabulary sophistication level using the following criteria
    \begin{itemize}
      \item Word Frequency: Proportion of uncommon words (not in the 5000 most frequent words)
      \item Word Length: Average syllable count and character length of words
      \item Lexical Diversity: Type–token ratio (unique words divided by total words)
      \item Technical Terminology: Presence of specialized or domain‑specific vocabulary
      \item Lexical Density: Ratio of content words (nouns, verbs, adjectives, adverbs) to function words (pronouns, prepositions, articles, etc.)
    \end{itemize}
    
\item Syntactic Complexity (\emph{Weight: 20\%}): Analyze sentence‑structure complexity using these metrics
    \begin{itemize}
      \item Sentence Length: Average number of words per sentence
      \item Clause Density: Number of clauses per sentence
      \item Subordination: Frequency and depth of subordinate clauses
      \item Passive Voice: Proportion of sentences in passive voice
      \item Syntactic Variety: Diversity of sentence structures
      \item Embedding Depth: How deeply clauses are nested within one another
    \end{itemize}

\item Conceptual Density (\emph{Weight: 25\%}): Assess the density and abstraction level of ideas presented
    \begin{itemize}
      \item Concept Count: Number of distinct concepts, ideas, or arguments introduced
      \item Concept Abstraction: Level of concreteness vs.\ abstraction of concepts
      \item Conceptual Networks: Complexity of relationships between concepts
      \item Information Density: Amount of information conveyed per paragraph
      \item Theoretical Complexity: Depth of theoretical constructs presented
    \end{itemize}

\item Domain Specificity (\emph{Weight: 15\%}): Evaluate how much specialized domain knowledge is required
    \begin{itemize}
      \item Background Knowledge: Prerequisite knowledge assumed by the text
      \item Domain Vocabulary: Concentration of field‑specific terminology
      \item Conceptual Familiarity: How familiar concepts would be to general readers
      \item Specialized References: References to domain‑specific methods, theories, or figures
      \item Audience Specificity: How targeted the text is to specialists vs.\ general readers
    \end{itemize}

\item Logical Structure (\emph{Weight: 10\%}): Analyze the complexity of reasoning patterns
    \begin{itemize}
      \item Argument Structure: Complexity of argumentative or explanatory structure
      \item Logical Operations: Presence of conditional, causal, comparative reasoning
      \item Inference Requirements: Extent to which the reader must infer rather than being told explicitly
      \item Logical Connections: Clarity and complexity of connections between ideas
      \item Reasoning Chains: Length and complexity of logical chains
    \end{itemize}

\item Contextual Dependencies (\emph{Weight: 10\%}): Assess how much the text relies on external context
    \begin{itemize}
      \item Intertextual References: References to other texts or knowledge sources
      \item Cultural Knowledge: Required cultural or historical background
      \item Implicit Information: Amount of information that remains unstated yet necessary
      \item Presuppositions: Assumptions the text makes about reader knowledge
      \item Discourse Context: Degree to which meaning depends on broader discourse context
    \end{itemize}

\end{enumerate}

\textbf{Text to Evaluate}

\verb|{text}|

\textbf{Required Output Format}

Only return a JSON object with the following structure:

\begin{verbatim}
{
  "lexical_complexity": {
    "score": <0-10 number>
  },
  "syntactic_complexity": {
    "score": <0-10 number>
  },
  "conceptual_density": {
    "score": <0-10 number>
  },
  "domain_specificity": {
    "score": <0-10 number>
  },
  "logical_structure": {
    "score": <0-10 number>
  },
  "contextual_dependencies": {
    "score": <0-10 number>
  },
  "final_weighted_score": 
  <calculated final score as decimal>,
  "normalized_complexity_score": 
  <rounded to one decimal place, e.g. 4.5>
}
\end{verbatim}

\end{tcolorbox}
\end{scriptsize}
\end{figure*}

\end{document}